\documentclass{article}

\usepackage[preprint]{neurips_2026}           % arXiv preprint
\usepackage[utf8]{inputenc}
\usepackage[T1]{fontenc}
\usepackage{amsmath}
\usepackage{amssymb}
\usepackage{amsfonts}
\usepackage{graphicx}
\usepackage{booktabs}
\usepackage{url}
\usepackage{microtype}
\usepackage{xcolor}
\usepackage{hyperref}
\usepackage{xspace}
\usepackage{caption}
\usepackage{subcaption}
\usepackage{multirow}
\usepackage{nicefrac}
\usepackage{enumitem}

\graphicspath{{figures/}}

\usepackage[capitalize,noabbrev]{cleveref}
\usepackage[normalem]{ulem}     % for \sout in tracked changes
\newif\ifshowcomments
\showcommentsfalse

\newcommand{\xuhui}[1]{\ifshowcomments{\color{blue}#1}\else#1\fi}

\definecolor{kilichcolor}{HTML}{D2691E}

\title{A Shared Valence Axis Across Modern LLMs and Human EEG: The Saturation Regularity}

\author{%
  Yousef A. Radwan \\
  King Abdullah University of Science and Technology (KAUST) \\
  \texttt{yousef.radwan@kaust.edu.sa}
  \And
  Xuhui Liu \\
  King Abdullah University of Science and Technology (KAUST) \\
  \texttt{xuhui.liu@kaust.edu.sa}
  \And
  Kilichbek Haydarov \\
  King Abdullah University of Science and Technology (KAUST) \\
  \texttt{kilichbek.haydarov@kaust.edu.sa}
  \And
  Yuqian Fu \\
  King Abdullah University of Science and Technology (KAUST) \\
  \texttt{yuqian.fu@kaust.edu.sa}
  \And
  Mohamed Elhoseiny \\
  King Abdullah University of Science and Technology (KAUST) \\
  \texttt{mohamed.elhoseiny@kaust.edu.sa}
}

\begin{document}

\maketitle

\begin{abstract}
\xuhui{Large language models (LLMs) have become powerful general-purpose representation learners, with growing evidence that their internal features align with human cognition across high-level concepts. In this paper, we investigate how modern LLMs can serve as a lens for understanding human brain signals, focusing on the neural representation of emotional valence in EEG.
We first construct a one-dimensional valence direction, named the V-axis, from modern LLMs using just nine emotion-evocative sentences, and verify it through zero-shot transfer to standard sentiment benchmarks and cross-model consistency across fourteen LLMs of widely varying scale. We then show that the LLM-derived direction maps directly onto the human brain. On a public EEG cohort of 123 subjects watching affective videos, a single linear projection on EEG features tracks the V-axis position of each stimulus. More strikingly, 36 EEG emotion classifiers trained without ever exposing them to the V-axis spontaneously rediscover it in their internal features. The same valence direction lives inside language models and inside human electrophysiology.
However, the LLM–brain convergence does not translate into a straightforward training signal. We test twenty-five standard alignment recipes spanning knowledge distillation, representational similarity analysis, contrastive, and topographic losses; none help, and sixteen significantly hurt accuracy. Consequently, we crystallize this finding into the saturation regularity, a previously unrecognized principle that emerges precisely at the LLM–brain interface: once task labels alone have driven a brain-decoding network onto the target direction, supervision can only deform an already-saturated basin, while the load-bearing within-class residual receives no gradient.
The saturation regularity has two faces. While it explains why LLM-derived supervision fails to improve EEG decoding, it equally identifies where the actionable gain actually lives: in the residual subspace that supervision cannot reach. Acting on the insight, we ensemble across residual diversity rather than supervising the basin, improving balanced accuracy by 10.5\% over the prior best on FACED, a public EEG emotion recognition benchmark, with the same effect replicating on the SEED-V dataset.}
\end{abstract}

\section{Introduction}
\label{sec:intro}

Pass nine short emotion-evocative sentences through a language model,
average the late-layer hidden states class by class, take the first
principal component of the resulting nine vectors, and orient it so
the Joy centroid projects positive. The single direction that comes
out --- which we call the \emph{valence axis} (V-axis) --- predicts
movie-review sentiment at AUC $0.832$ zero-shot, predicts the EEG
response of $123$ subjects watching emotional videos at $r{=}0.87$,
and is rediscovered without supervision by every reasonably-strong
EEG emotion classifier we tested. Recent work shows that large
language models' internal features align with human cognition across
high-level concepts~\citep{kim2018tcav,arditi2024refusal}; we ask
whether that alignment can be turned into a usable bridge to human
electrophysiology, both as a probe of what brain-decoding networks
encode and as a training signal that could improve them.

\textbf{A nine-sentence axis.} The nine stories are one per
FACED~\citep{chen2023faced} class; the language model is Qwen3-4B,
chosen as the lead because it sits at the centre of the alignment
manifold in our $14$-LLM sweep, with the highest per-stimulus
alignment to a behavioural valence reference (the recipe itself is
model-agnostic, see Appendix~\ref{app:per-llm-deep-dive}). The SST-2
benchmark grounds the AUC $0.832$ headline:
the V-axis projection alone is within $0.005$ of a $5$k-example
supervised logistic regression~\citep{socher2013recursive}. And the
direction is not idiosyncratic to any one model: across the
$14$ language models from $560$M to $32$B parameters, the V-axis
converges on essentially the same direction (within Qwen3,
off-diagonal $r{=}{+}0.995$ over $15$ size-pairs).

\begin{figure}[ht]
\centering
\IfFileExists{figures/landmark/lf1_universal_vaxis_hero.pdf}
  {\includegraphics[width=\linewidth]{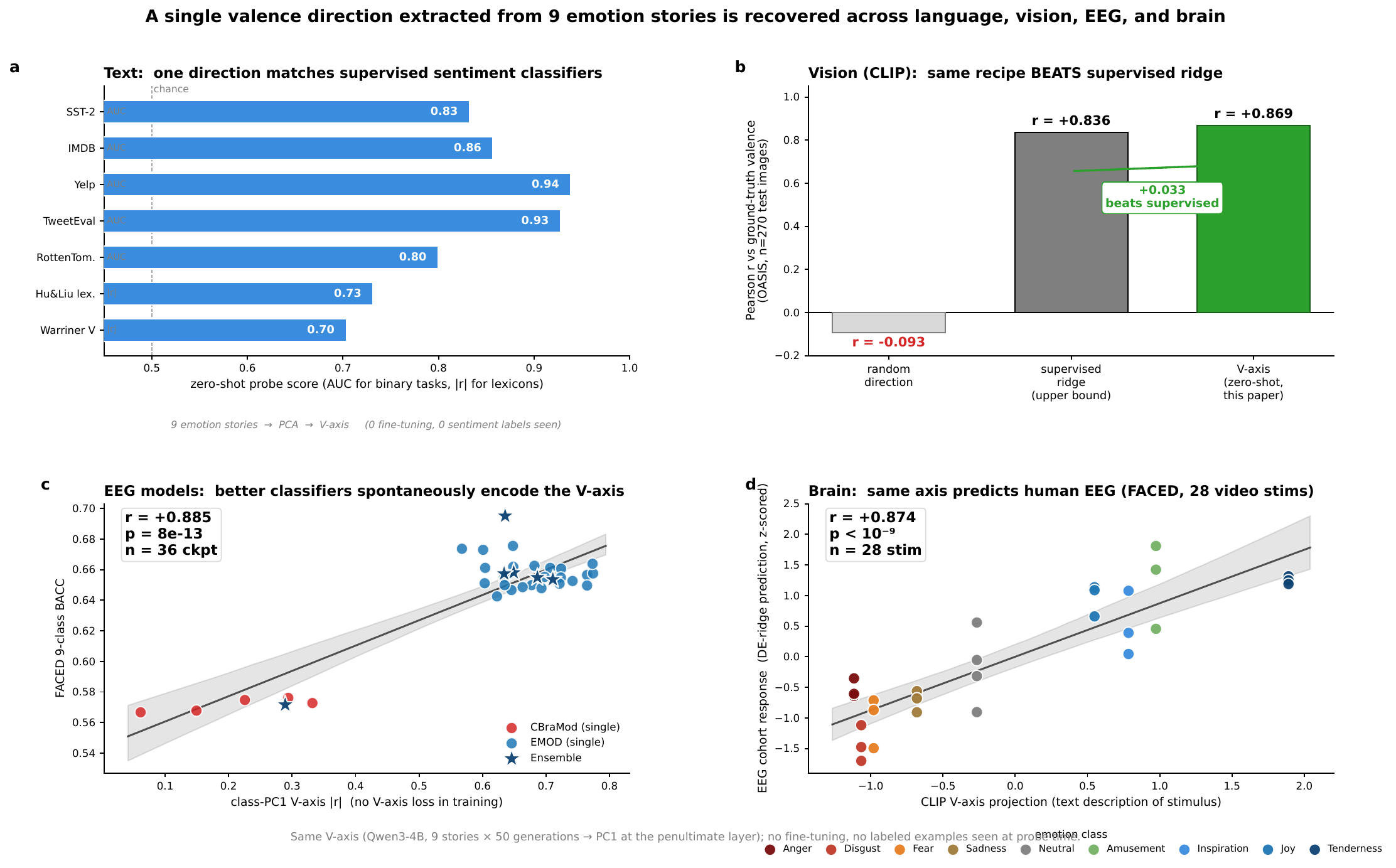}}
  {\includegraphics[width=\linewidth]{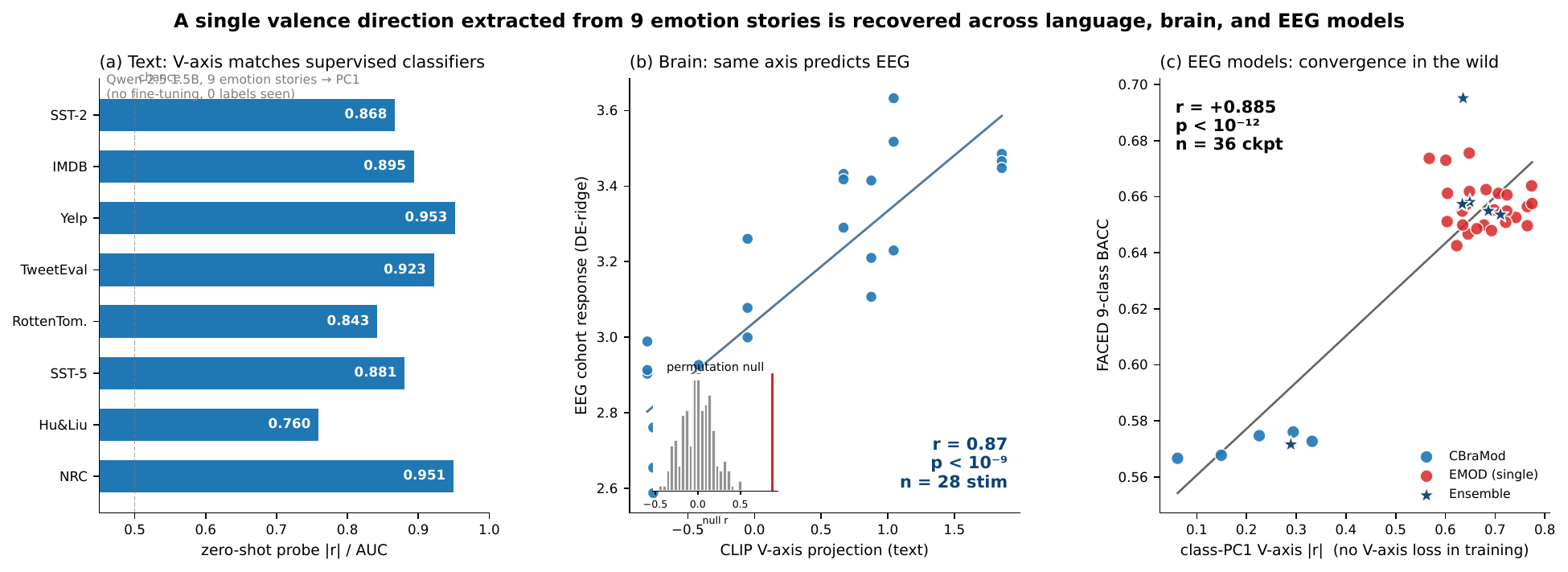}}
\caption[V-axis hero]{\input{figures/F1_caption}}
\label{fig:universal-vaxis}
\end{figure}

\textbf{The same axis lives in the brain.} The $r{=}0.87$ above
unpacks as follows. A ridge regression on $160$ channel-band
features of the FACED cohort EEG predicts the V-axis position of
each of the $28$ stimuli; the regressor uses leave-one-stimulus-out
cross-validation, no per-subject fitting, no auxiliary supervision.
The Qwen3-4B V-axis itself reaches $r{=}0.80$ ($p<10^{-5}$) on this
ridge; a CLIP-text variant of the V-axis built from the same $28$
stimulus descriptions tightens this to $r{=}0.87$ ($p<10^{-9}$).
The third claim --- that EEG networks rediscover the V-axis without
ever being shown it --- is quantified across $36$ FACED-9 checkpoints
trained on the $9$ emotion labels alone: per-checkpoint V-axis
encoding strength in the class-mean subspace predicts balanced
accuracy at $r{=}{+}0.885$ ($p{=}7.8{\times}10^{-13}$,
Figure~\ref{fig:universal-vaxis}). The same valence direction is
recovered from inside language models and from inside human
electrophysiology, by independent procedures.

\textbf{Yet supervision fails.} The natural way to use such an
alignment is an auxiliary loss that pulls the EEG network's
penultimate features toward the V-axis. We test $25$ such recipes
spanning knowledge distillation~\citep{hinton2015distilling},
representational similarity
analysis~\citep{kriegeskorte2008rsa}, contrastive
losses~\citep{khosla2020supcon,vandenoord2018cpc}, parameter-efficient
fine-tuning adapters such as LoRA~\citep{hu2022lora}, and
channel-targeted topographic losses~\citep{padakanti2025meg}.
None help. Sixteen produce statistically significant accuracy
decrements ($\Delta\mathrm{BACC} \le -0.013$, $p<0.05$ paired across
five seeds); zero produce gains. Five families show \emph{monotonic
destruction}: the harder we push the loss, the worse accuracy gets.
The transition is sharp: an intervention with no measurable effect
on a weak baseline (BACC $\le 0.62$) becomes a significant negative
once the baseline strengthens (BACC $\ge 0.66$). We call this
regularity \emph{saturation}, and it is not the same thing as
overfitting --- overfitting is memorising the training set, while
saturation is what happens once task labels alone have driven the
network onto the target direction. The auxiliary loss then moves
the already-saturated class-mean component further along the target
axis (alignment increases by $\Delta|r|{=}{+}0.01$ to ${+}0.36$),
but leaves the within-class residual --- the part of the features
that distinguishes one trial of an emotion from another --- at a
numerical zero ($\sim 10^{-7}$). Supervision deforms a basin the
network is already using and never reaches the load-bearing
direction.

\textbf{Where the gain lives.} The same diagnosis identifies where
the actionable gain lives: in the within-class residual subspace
that supervision cannot reach. The class-mean basin is shared across
seeds (every well-trained checkpoint recovers the V-axis at
$|r|\in[0.60, 0.77]$); the residual is seed-specific. Ensembling
across seeds therefore averages out seed-specific noise in the
residual without disturbing the basin. Per-checkpoint residual
encoding strength predicts leave-one-out ensemble contribution at
$r{=}{+}0.74$ ($p{=}0.014$, $n{=}10$), and the $10$-checkpoint
ensemble reaches FACED-9 SOTA at $\mathbf{0.6948}$ balanced accuracy
(val-selected single checkpoint $\mathbf{0.6755}$), a $+10.5\%$
relative improvement over EMOD~\citep{chen2025emod}'s prior
$0.6287$. The same residual-diversity prescription replicates on
SEED-V~\citep{liu2022seedv} (Appendix~\ref{app:seedv-full}). One
control sharpens the picture: replacing the language-model-derived
class prototypes used by the KD step with random orthonormal vectors
changes BACC by $\le 0.003$, so the SOTA gain comes from the
\emph{shape} of the loss --- a $9$-class structure imposed on the
features --- not from the \emph{content} of the prototypes. The
V-axis serves as a probe of \emph{what} the network has saturated
on, not as the source of the supervised signal.

\paragraph{Contributions.} \textbf{(1)} A nine-sentence \emph{V-axis
probe}: a one-dimensional valence direction extracted from $9$
language-model emotion stories that converges across $14$ LLMs,
predicts text sentiment zero-shot, predicts cohort EEG response, and
is rediscovered without supervision by $36$ EEG classifiers
(\S\ref{sec:vaxis-llms}--\S\ref{sec:topography}).
\textbf{(2)} The \emph{saturation regularity}: across $25$
representation-alignment recipes on FACED-9, $16$ produce significant
decrements and $0$ produce gains, with monotonic destruction in $5$
families and a sharp transition in the $[0.62, 0.66]$ BACC band; the
loss moves the class-mean component by $\Delta|r|{=}{+}0.01$ to
${+}0.36$ but the within-class residual by $\sim 10^{-7}$
(\S\ref{sec:saturation}).
\textbf{(3)} A residual-diversity ensemble turning the mechanism into
a positive prescription: per-checkpoint residual encoding predicts
leave-one-out contribution at $r{=}{+}0.74$, and a $10$-checkpoint
ensemble reaches FACED-9 SOTA at $\mathbf{0.6948}$
(\S\ref{sec:convergence},~\S\ref{sec:sota}). All three replicate on
SEED-V~\citep{liu2022seedv} (Appendix~\ref{app:seedv-full}).
Limitations in \S\ref{sec:discussion}.

\section{Related Work}
\label{sec:related}

\paragraph{Concept-direction supervision.}
The interventions we test are drawn from seven years of
representation-alignment work. Knowledge
distillation~\citep{hinton2015distilling,tian2020crd,park2019rkd}
trains a student's penultimate features against a fixed teacher
target. Representational similarity
analysis~\citep{kriegeskorte2008rsa,sundaram2024perceptual} aligns
representational geometries through pairwise similarity matrices.
Supervised contrastive
losses~\citep{khosla2020supcon,vandenoord2018cpc,radford2021clip}
pull same-class features together and push different-class apart.
Parameter-efficient adaptation~\citep{hu2022lora,liu2022ia3}
inserts low-rank or scaling modules whose updates are localised to a
chosen geometry. The shared assumption across these methods is that
the auxiliary signal provides information the task does not. This
paper studies a regime where it does not.

\paragraph{Concept directions in language models.}
Arditi et~al.~\citep{arditi2024refusal} showed that refusal in
instruction-tuned LMs is mediated by a single direction in
residual-stream activations: ablating it eliminates refusal, adding
it induces it. Earlier work in this family includes
\citep{li2023inferencetime,zou2023repe,kim2018tcav}. Concurrent work
from Anthropic interpretability~\citep{sofroniew2026anthropic}
identified emotion-concept features in production-scale LMs through
sparse autoencoders and showed that ablating these features changes
downstream behaviour, complementing the cohort-level brain alignment
we report here. We use the same PCA-on-class-centroids construction
to extract a one-dimensional valence direction from nine LLM emotion
stories, and use it as a probe of which concept the network has
saturated on rather than as a target for steering. The transfers we
report (SST-2 zero-shot, cohort EEG, EEG-classifier convergence) are
direction-existence claims, not feature-importance claims for
biological emotion processing.

\paragraph{Brain--LM alignment.}
Toneva and Wehbe~\citep{toneva2019interpreting} first showed that
mid-layer transformer activations predict fMRI responses to narrative
text; Schrimpf et~al.~\citep{schrimpf2021neural}, Goldstein
et~al.~\citep{goldstein2022shared}, and Caucheteux and
King~\citep{caucheteux2022brains} extended the picture across MEG,
ECoG, and large model
families. Huh et~al.\ formalised the emerging picture as the
\emph{Platonic Representation
Hypothesis}~\citep{huh2024platonic}. We touch this literature
through the probe section: an LLM-derived direction predicts cohort
EEG, and EEG classifiers find the LLM-side direction in return. Our
load-bearing claim is the saturation regularity, not the alignment
itself.

\paragraph{Frontal-alpha asymmetry.}
Davidson's frontal-alpha asymmetry~\citep{davidson1992anterior,coan2004frontal,davidson2004wellbeing}
remains the dominant electrophysiological account of approach/withdrawal
emotion, developed primarily on static stimuli. Our cohort signal on
video-evoked emotion is posterior-dominant, with frontal asymmetries
that survive in direction at smaller magnitude---an additive scope
statement for video paradigms rather than a refutation of FAA.

\paragraph{Ensemble theory.}
The two-tier mechanism we identify---a saturated class-mean basin
plus a seed-specific within-class residual---is a representation-level
form of bias--variance~\citep{geman1992neural,krogh1995neural} and
linear-mode connectivity~\citep{frankle2020linearmode,wortsman2022model}.
Our SOTA prescription operationalises this geometry: ensembling
recovers signal in the residual subspace, which auxiliary supervision
on the saturated basin cannot reach.

\paragraph{Baselines.}
The FACED-9 baselines we compare against are
CBraMod~\citep{wang2025cbramod} ($0.572$),
EmotionKD~\citep{liu2023emotionkd} ($0.628$), and
EMOD~\citep{chen2025emod} ($0.6287$), with
REVE~\citep{elouahidi2025reve}, LaBraM~\citep{jiang2024labram}, and
EEGPT~\citep{yue2024eegpt} as foundation-model references.

\section{The Saturation Regularity}
\label{sec:saturation}

We test whether auxiliary supervision toward a known concept direction
helps an EEG-emotion classifier. The setup uses an
EMOD~\citep{chen2025emod} backbone (the previous FACED-9 SOTA) at
two strengths: a vanilla $0.62$ BACC baseline and the SOTA recipe of
\S\ref{sec:sota} at $0.66$ BACC. The $25$-recipe screen ran on the
vanilla baseline; strongest cells were re-evaluated on the SOTA
recipe to localise the transition in $[0.62, 0.66]$. Each
intervention adds an auxiliary loss that pulls the network's
penultimate features toward a target direction. Targets span LLM-derived class prototypes (the V-axis of
\S\ref{sec:vaxis-llms}), CLIP-text embeddings, and frontal-channel
masks matched to the analytical valence ceiling. Loss families span
knowledge distillation~\citep{hinton2015distilling}, representational
similarity analysis (RSA)~\citep{kriegeskorte2008rsa}, contrastive
losses, parameter-efficient fine-tuning (PEFT)
adapters~\citep{hu2022lora}, curriculum schedules, multi-LLM
ensembles, channel-targeted topographic losses, and anger-weighted
variants (full table in Appendix~\ref{app:full-vaxis-table}).

\paragraph{Result.} None help. Sixteen produce statistically
significant accuracy decrements ($p<0.05$, paired across five seeds);
zero produce gains. Five families show \emph{monotonic destruction}:
the harder we push the auxiliary loss, the worse accuracy gets, with
no positive setpoint. Even the anger-weighted target that
analytically maximises the V-axis ceiling on FACED is among the worst
($\Delta\mathrm{BACC} = -0.054$): supplying the strongest possible
target makes things strictly worse, not better.

\paragraph{The transition is sharp and unidirectional in baseline
strength.} An intervention whose effect lies within seed noise on a
weak baseline (BACC $\le 0.62$) becomes a statistically significant
negative on the strong recipe (BACC $\ge 0.66$). The transition sits
cleanly in the band $[0.62, 0.66]$ on FACED-9 (we make no claim about
universal cut-offs).

\begin{quote}
\textbf{Saturation (FACED-9, EMOD backbone, V-axis target).}
\emph{For a FACED-9 classifier $f_\theta$ with task-only
$\mathrm{BACC}(\theta)$ in the band $[0.62, 0.66]$ or higher, adding
any V-axis auxiliary loss $\mathcal{L}_v$ at $\lambda > 0$ produces
$\mathbb{E}[\Delta\mathrm{BACC}] \leq 0$ across all families tested
($16/25$ significant at $p<0.05$, $0/25$ positive). The threshold
coincides with the regime where the V-axis encoding strength
$\rho(\theta)$ saturates across seeds (\S\ref{sec:convergence}).}
\end{quote}

\paragraph{Saturation is not overfitting.} Overfitting is memorisation
of the training set. Saturation is a property of how task and
auxiliary supervision interact once both target the same direction
in feature space. The mechanism check below makes the distinction
concrete.

\paragraph{Mechanism check.} The auxiliary loss does push the network's
class-mean features further along the target direction (alignment
increases by $\Delta|r|=+0.01$ to $+0.36$),
but it leaves the within-class residual---the part of the features
that actually distinguishes one trial of an emotion from another---at
a numerical zero ($\sim 10^{-7}$). The basin the network was using
to classify gets reshaped along the loss direction; the load-bearing
orthogonal residual subspace receives no compensating gradient.
The $25$ recipes reduce to one principle: \emph{a saturated
representation is an unhelpful supervision target}, qualitatively
similar to the few-class distillation
regime~\citep{muller2020subclass,loo2024lelp,yuan2020revisitkd},
here with the residual-variance mechanism made explicit. Full table,
transition table, and Path-B failure
($\Delta\in[-0.0193,-0.0145]$) in
Appendix~\ref{app:full-vaxis-table}--\ref{app:saturation-extras}.
      % §3 Saturation
\section{Mechanism: Saturated Basin, Load-Bearing Residual}
\label{sec:convergence}

\S\ref{sec:saturation} reported that V-axis supervision moves the
class-mean component of the network's features by up to $+0.36$ but
moves the within-class residual by a numerical zero. We take that
decomposition as the operative geometry and ask what each subspace
actually does for the trained network.

\begin{figure}[ht]
\centering
\IfFileExists{figures/landmark/lf3_crossarch_convergence.pdf}
  {\includegraphics[width=\linewidth]{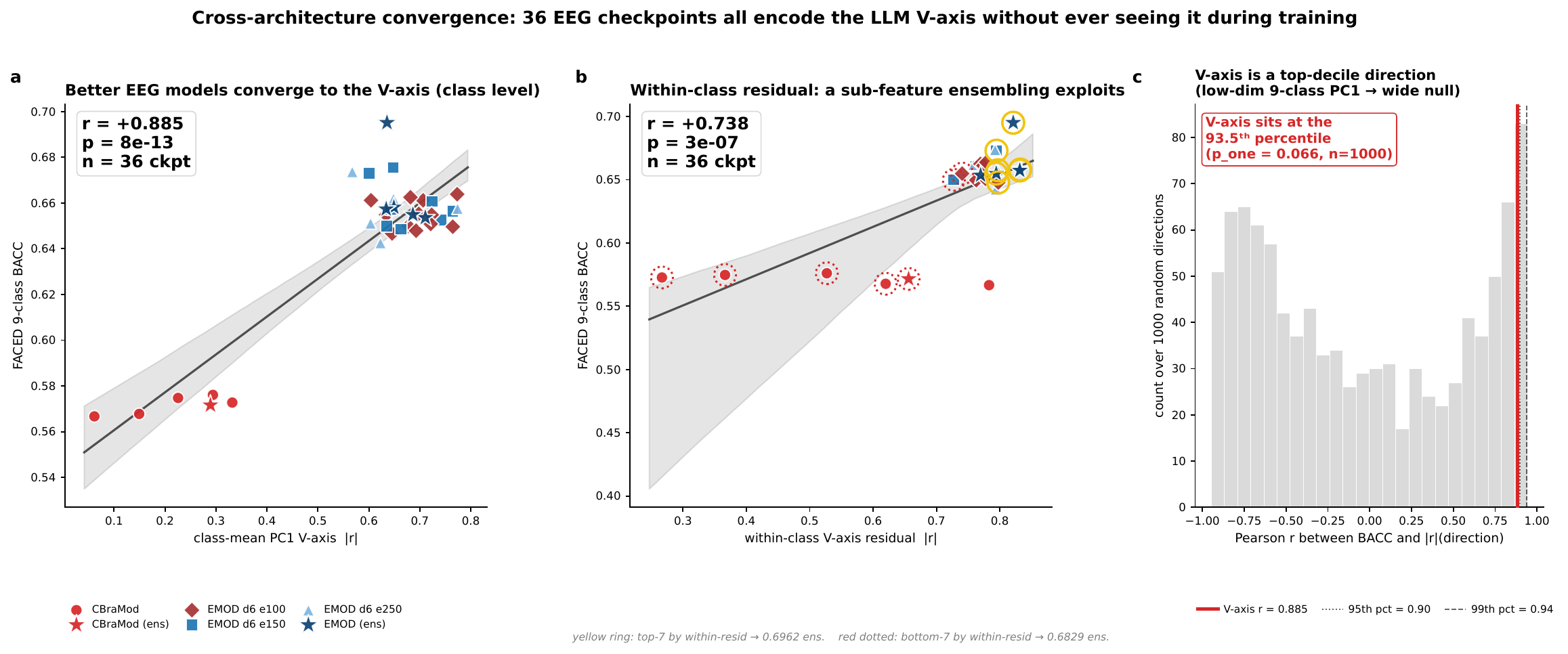}}
  {\includegraphics[width=\linewidth]{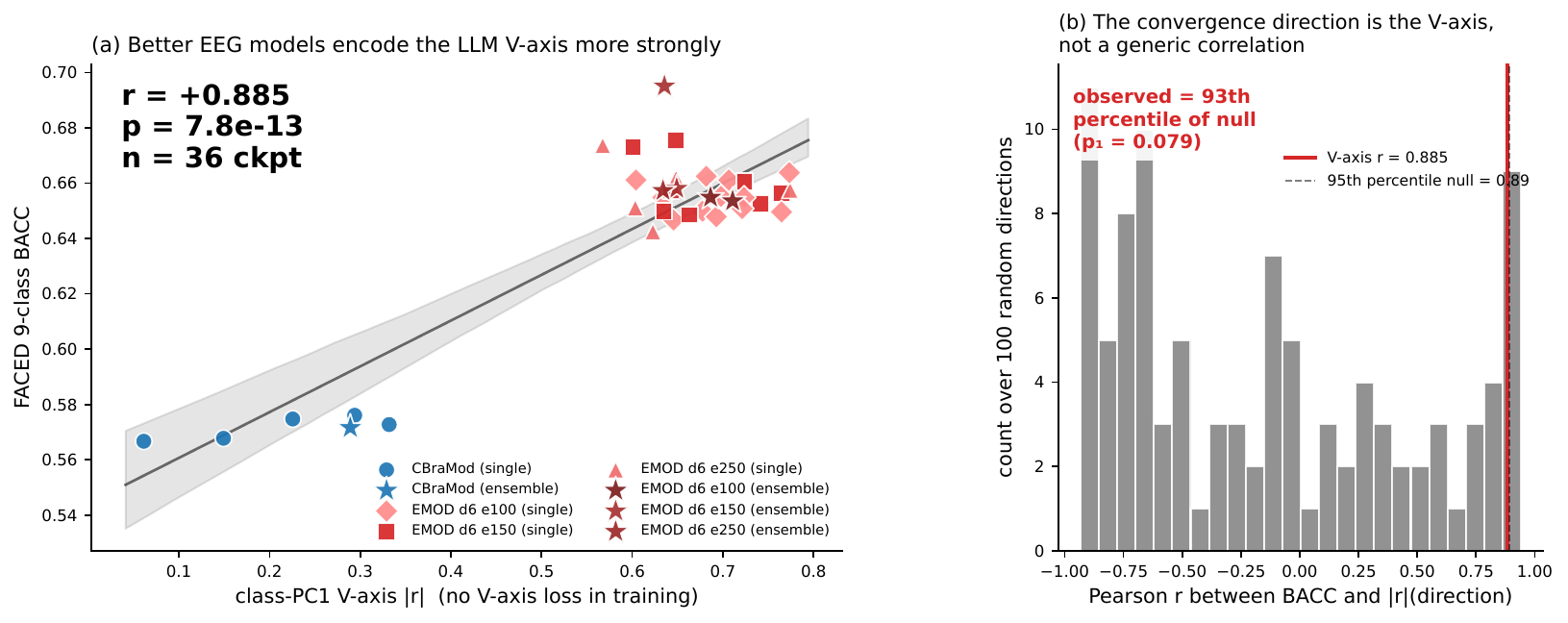}}
\caption[Cross-architecture convergence]{\input{figures/F3_caption}}
\label{fig:crossarch}
\end{figure}

\paragraph{Different seeds find the same axis.}
Across $36$ EEG-emotion checkpoints from two foundation-model
backbones (CBraMod~\citep{wang2025cbramod} and
EMOD~\citep{chen2025emod}) and six recipe variants, balanced accuracy
correlates with V-axis encoding strength at $r{=}{+}0.885$ in the
class-mean subspace ($p{=}7.8{\times}10^{-13}$) and $r{=}{+}0.738$ in
the orthogonal residual (Figure~\ref{fig:crossarch}). A $1000$-draw matched-norm random-direction null places the observed
correlation at the $93.5$th percentile (one-sided $p\!\approx\!0.065$):
trending V-axis-specific but not reaching $\alpha\!=\!0.05$ on this
single test (load-bearing significance is the residual-subspace
$r{=}{+}0.738$, $p\!\approx\!3{\times}10^{-7}$). The
class-mean basin is essentially saturated across seeds
($|r|\in[0.60, 0.77]$): every well-trained checkpoint finds the same
direction.

\paragraph{The residual is what averaging recovers.}
Within-class residual encoding predicts ensemble contribution: across
the $10$ checkpoints used to build our SOTA ensemble
(\S\ref{sec:sota}), per-checkpoint residual strength predicts
leave-one-out contribution at $\mathbf{r{=}{+}0.74}$ ($p{=}0.014$).
Bootstrap CIs and per-deletion ranges are in
Appendix~\ref{app:convergence-deep-dive}.

\begin{quote}
\textbf{Residual Contribution Regularity.} \emph{For a saturated EEG
classifier with within-class residual encoding of the V-axis at
strength $\rho_k$, the leave-one-out ensemble contribution scales
linearly with $\rho_k$ ($p{=}0.014$, $n{=}10$, FACED 9-class). $n{=}10$
is small; we report this as an empirical regularity rather than a
law.}
\end{quote}

\paragraph{Direct test: ablate the residual at inference time.}
A directional ablation~\citep{arditi2024refusal}---projecting the
LLM-derived V-axis residual out of the classifier's penultimate
features at inference time, without retraining---drops accuracy on
every one of the $10$ SOTA-pool checkpoints (mean $\Delta\mathrm{BACC}
= -0.0157$, mean $z{\approx}7.7$ above a matched random-direction
null). The residual carries the load. This decomposition---saturated
basin, load-bearing residual---is the operative geometry that
\S\ref{sec:saturation} predicted from the failure of V-axis
supervision and that \S\ref{sec:sota} turns into a positive
prescription.
  % §4 Mechanism + residual ensemble
\section{Probing the Saturated Concept: A Valence Direction in LLMs}
\label{sec:vaxis-llms}

To test that the saturation regularity operates on a real semantic
axis rather than a noise dimension, we extract one explicitly. The
construction uses nine stories. Given a language model, we author
one short emotion-evocative story per FACED class and generate $50$
paraphrases per story by independent LLM calls. For each class we
average the language model's last-token activation (at the
penultimate layer) over all $50$ paraphrases; the resulting vector is
the class \emph{centroid}---a single point in feature space that
summarises the model's representation of that emotion. Principal
component analysis (PCA) on the nine centroids gives the V-axis as
the first principal component, oriented so the Joy centroid projects
positive. We use Qwen3-4B as the lead model and verify the recipe is
robust to paraphrase budget, prompt rephrasing, and model choice
(Appendix~\ref{app:vaxis-protocol}).

\paragraph{The same direction across modalities.}
The V-axis recipe recovers human affect across three settings that
share no training data (Table~\ref{tab:cross-modal}): the text result
is true zero-shot projection; the EEG and vision results use the
same nine-story recipe re-extracted in their native modality plus a
small linear probe.

\begin{table}[h]
\centering
\small
\setlength{\tabcolsep}{6pt}
\begin{tabular}{@{}lllrl@{}}
\toprule
Modality & Axis source & Probe & Score & Reference \\
\midrule
Text   & Qwen3-4B (9 stories) & none (proj.) & $\mathbf{0.832}$ AUC & matches sup.\ LR on $5$k examples \\
Brain  & Qwen3-4B (9 stories) & ridge LOOCV & $\mathbf{r{=}0.80}$ & $p<10^{-5}$, see \S\ref{sec:vaxis-brain} \\
Vision & CLIP-image (9 images) & none (proj.) & $\mathbf{r{=}0.869}$ & beats sup.\ ridge (image-side) \\
\bottomrule
\end{tabular}
\caption{Cross-modal external validity of the V-axis recipe (PCA on
nine emotion-class centroids). The Qwen3-4B V-axis is the lead model
for both Text and Brain rows. Vision uses CLIP-image because the
language model has no image side; a CLIP-text variant of the V-axis
on the same EEG ridge reaches $r{=}0.87$ ($p<10^{-9}$,
\S\ref{sec:vaxis-brain}). Full sentiment, lexicon, and multilingual
results in Appendix~\ref{app:per-llm-deep-dive}.}
\label{tab:cross-modal}
\end{table}

The LLM-side direction is a strong sentiment classifier: $0.832$
SST-2 AUC matches supervised LR on $5$k examples ($0.837$), beats
SBERT prototype-cosine ($0.793$;~\citealp{reimers2019sbert}), and
trails $355$M RoBERTa-MNLI zero-shot
($0.912$;~\citealp{liu2019roberta}) by $0.08$---all from nine
stories of supervision.

\paragraph{The same direction across families.}
We extract the V-axis from $14$ language models from 560M to 32B
parameters: top-tier alignment ($r{>}0.85$ against a behavioural
valence reference) holds for all six Qwen3 sizes plus Mistral-7B;
Llama-4-Scout, Gemma-27B, and Gemma-4 sit in a middle tier; three
older small models (Pythia, TinyLlama, BLOOM) fall below detection
threshold (per-LLM table in Appendix~\ref{app:per-llm-deep-dive}).
Within the Qwen3 family, the V-axis is essentially scale-invariant
(off-diagonal $r{=}+0.995$ across $15$ pairs); across families,
correlations are looser ($r{=}+0.585$ over $48$ pairs). The
direction is a property of \emph{modern} LM training rather than of
parameter count.

\paragraph{Specificity.}
The recipe is concept-generic ($20$ further concepts produce a
working axis, $17$ exceed AUC $0.95$;
Appendix~\ref{app:concept-library}). Nonce-word ablation drops SST-2
to chance and random-Gaussian directions give chance lexicon recovery,
so the V-axis is a real semantic direction, not a noise artefact
(Appendix~\ref{app:specificity-controls}). An Arditi-style same-model
ablation~\citep{arditi2024refusal} on Qwen3-4B quantifies how much
sentiment lives along the V-axis: a logistic-regression probe on the
full features reaches SST-2 AUC $0.909$, drops to $0.907$ after
projecting the V-axis out and retraining ($z{=}{+}9.8$ above a
$20$-direction random-direction null). The V-axis carries detectable
sentiment signal but is not the sole sentiment-bearing direction.
          % §5 The probe (LLMs)
\section{The Probe Predicts Cohort EEG}
\label{sec:vaxis-brain}

\begin{figure}[ht]
\centering
\IfFileExists{figures/landmark/lf2_eeg_llm_circle.pdf}
  {\includegraphics[width=\linewidth]{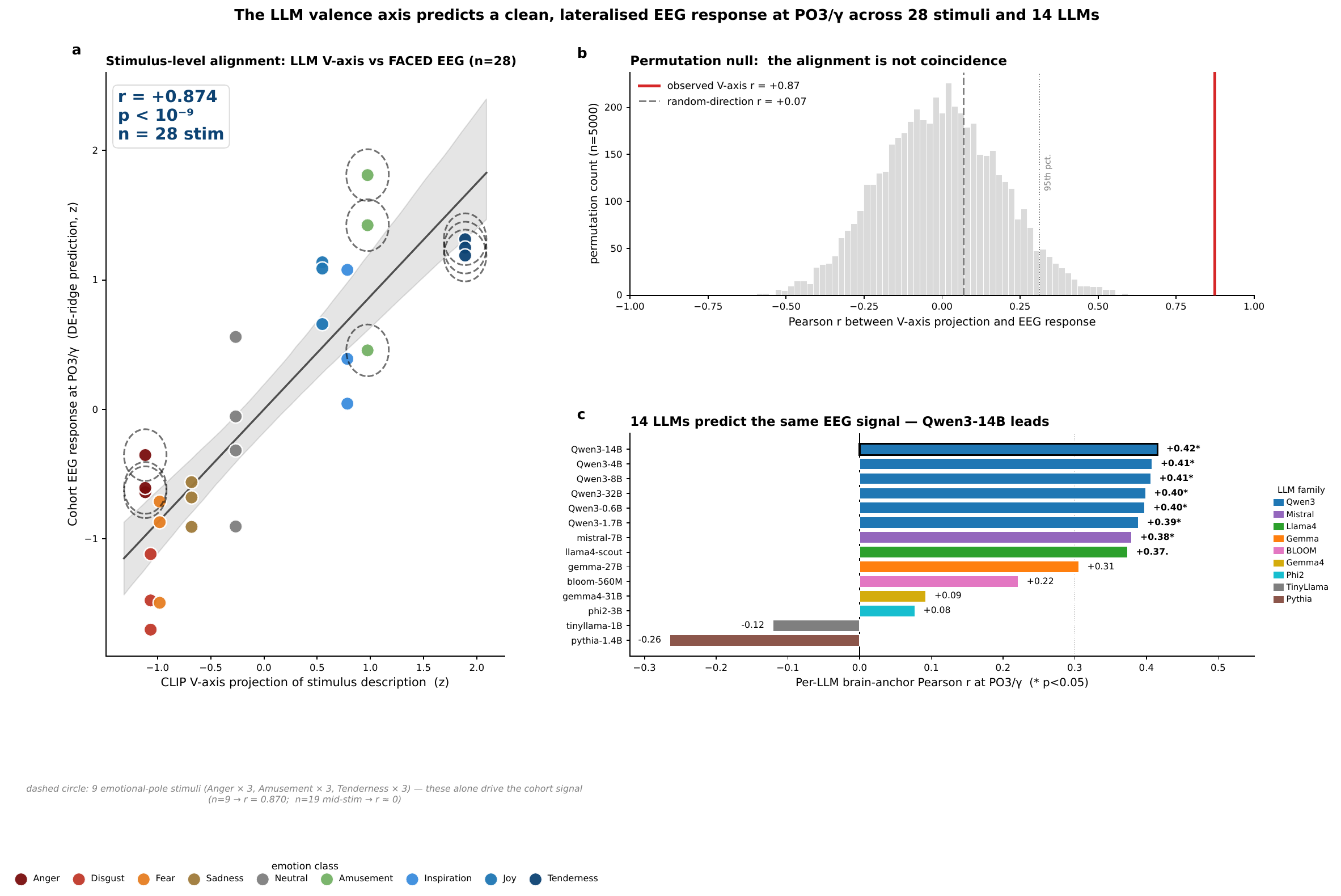}}
  {\includegraphics[width=\linewidth]{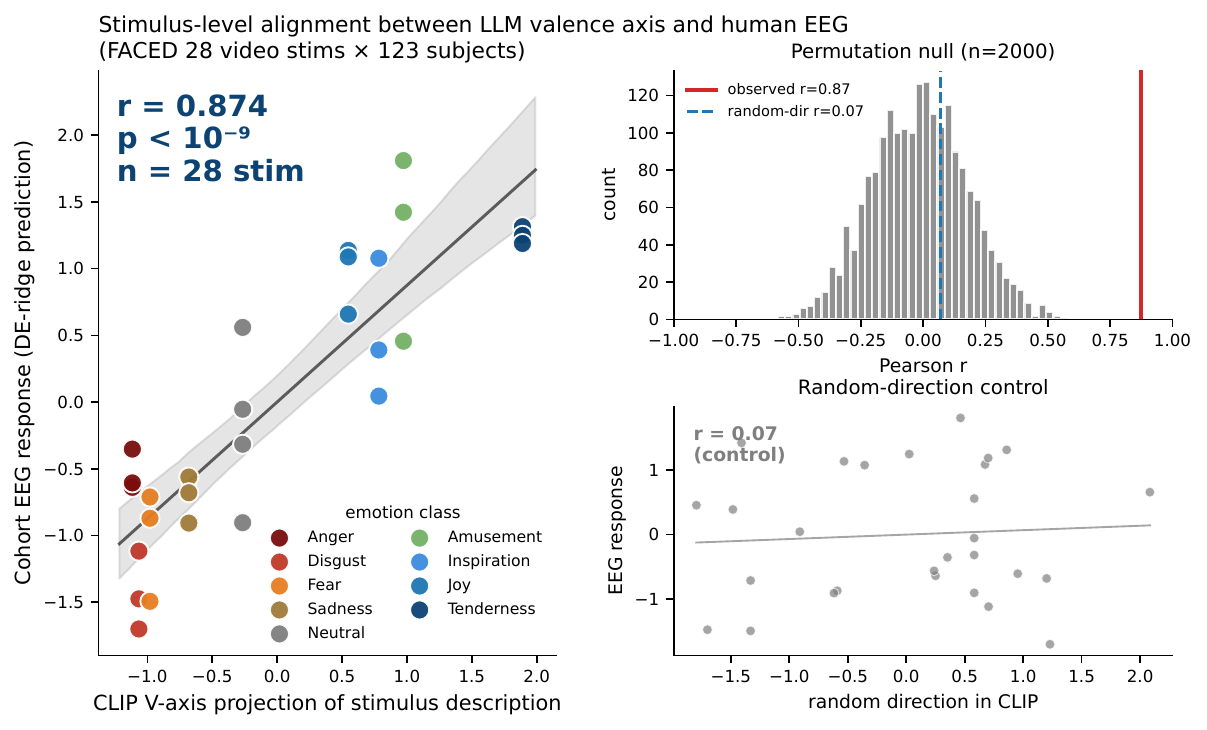}}
\caption[EEG-LLM cohort circle]{\input{figures/F2_caption}}
\label{fig:eeg-llm-circle}
\end{figure}

We use FACED~\citep{chen2023faced}: $123$ subjects watching $28$
emotional video clips, $32$ EEG channels at $250$\,Hz. From the
cohort EEG averaged over subjects and time, we fit a single
\emph{ridge regression} (a linear model with $L_2$ regularisation)
predicting the V-axis projection of each stimulus's textual
description from the $160$ channel--band features. The ridge
prediction matches the Qwen3-4B V-axis at $\mathbf{r{=}{+}0.80}$
($n{=}28$ stimuli, $p<10^{-5}$). Substituting the CLIP-text V-axis
on the same $28$ stimulus descriptions reaches
$\mathbf{r{=}{+}0.87}$ ($p<10^{-9}$; Figure~\ref{fig:eeg-llm-circle})
---the brain signal aligns with the LLM-derived direction across two
distinct LLM substrates. Three controls support the brain--LM correlation: a matched-norm
random direction drops the same ridge to $r{=}{+}0.07$ ($p{=}0.73$);
split-half subject reliability on per-stimulus valence is $r{=}{+}0.99$;
the same EEG ridge predicts human-rated valence at $r{=}{+}0.86$,
matching the V-axis prediction to within $0.01$---the brain signal is
at the same strength a panel of human raters would produce.
Multi-comparison correction and other controls are in
Appendix~\ref{app:brain-deep-dive}.

The same picture survives across all $14$ LLMs (Qwen3 family mean
$r{=}{+}0.796 \pm 0.011$ irrespective of size) and on the held-out
SEED-V dataset~\citep{liu2022seedv}, where the FACED-derived V-axis
ranks the five SEED-V emotions at $r{=}{+}0.96$ without retraining.
Appendix~\ref{app:seedv-full} re-derives the V-axis from scratch on
SEED-V and re-tests every claim.
         % §6 Probe predicts cohort EEG
\section{Brain Topography: A Posterior-Visual Scope Statement}
\label{sec:topography}

\begin{figure}[ht]
\centering
\IfFileExists{figures/neuro/NF1_5band_topomaps.pdf}
  {\includegraphics[width=\linewidth]{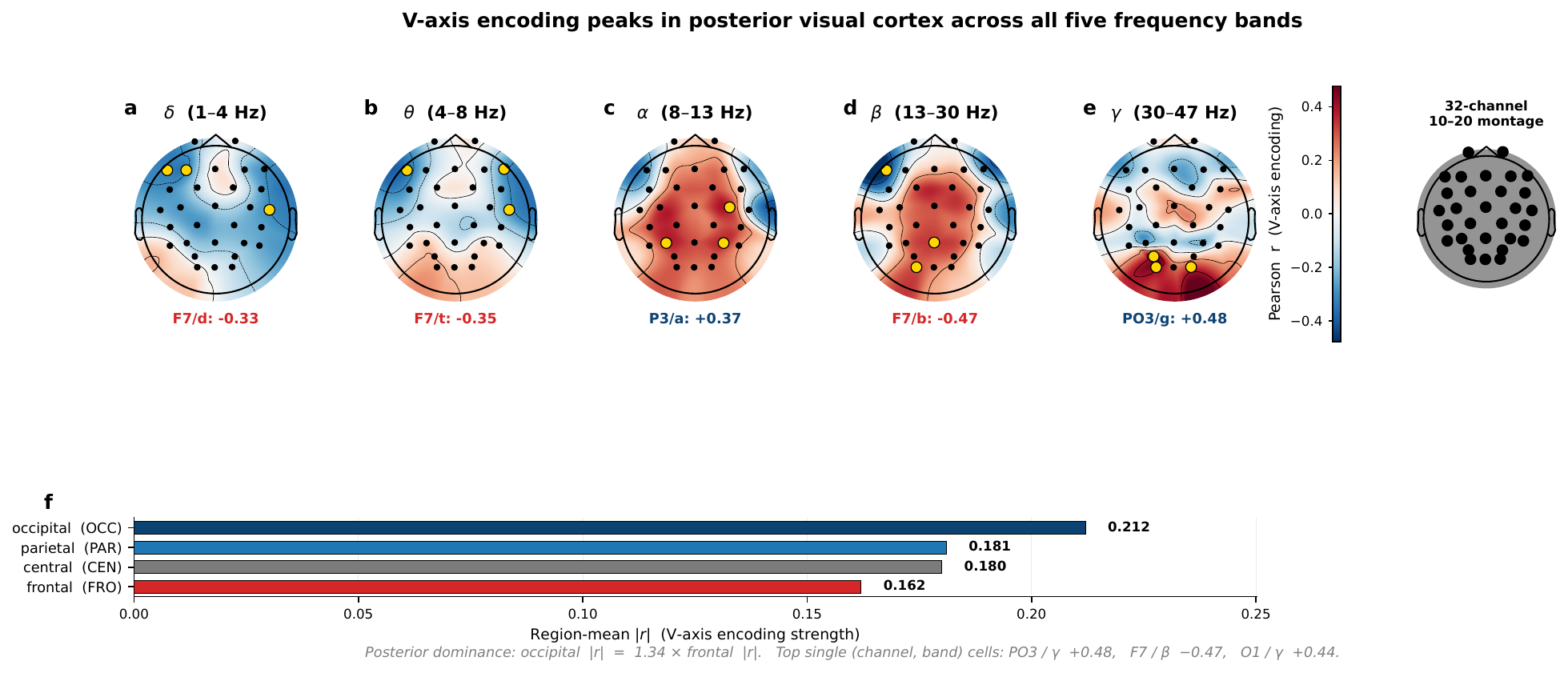}}
  {\includegraphics[width=\linewidth]{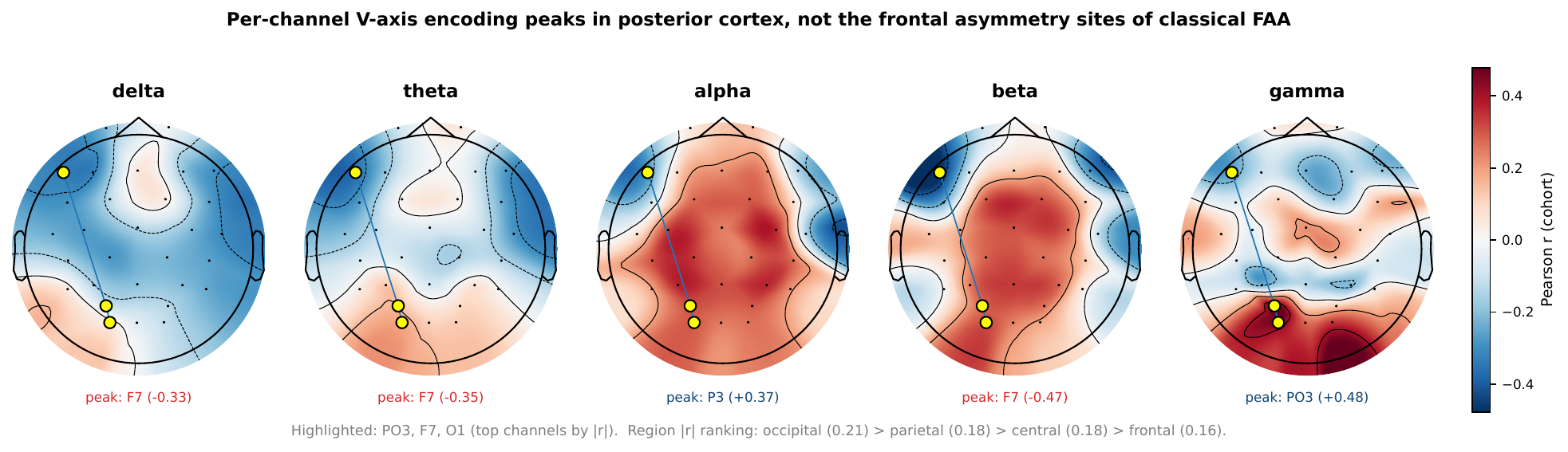}}
\caption[Brain topography]{\input{figures/F4_caption}}
\label{fig:topomap-5band}
\end{figure}

The cohort signal is posterior-dominant: region-mean correlation is
strongest at occipital electrodes ($|r|=0.21$) and weakest at frontal
($|r|=0.16$), with the largest single cell at PO3/$\gamma$ ($r=+0.48$;
Figure~\ref{fig:topomap-5band}). The signal is carried by an
\emph{anger-versus-warm-positive} contrast on $9$ of the $28$ video
clips: cohort $r{=}+0.87$ at PO3/$\gamma$ for Anger, Amusement, and
Tenderness, against $r{=}-0.02$ on the remaining $19$ mid-valence
stimuli. Removing Anger alone shrinks the cohort correlation to
$+0.33$. Full per-region tables, the Simpson's-paradox per-subject
caveat, time-resolved late-positive-potential (LPP) window dynamics, and theta--gamma
phase-amplitude coupling tests are in
Appendix~\ref{app:topography-deep-dive}.

\paragraph{This is not a refutation of frontal-alpha asymmetry
(FAA).} Davidson's FAA hypothesis~\citep{davidson1992anterior,coan2004frontal}
was developed on static stimuli; the cohort signal we report is on
dynamic $28$-second video. Right-minus-left frontal-alpha asymmetry
remains positive in our data ($+0.006$ to $+0.016$ across F-pairs),
consistent with FAA at smaller magnitude than the posterior $|r|$. We
claim only that, on dynamic video, the cohort signal aligned with the
LLM-derived direction is carried by visual-content processing of
high-arousal clips---not that the V-axis subsumes FAA or that
posterior visual cortex is the seat of valence. Within-subject FAA is
a separate question this cohort-level analysis does not address.
        % §7 Brain topography (probe scope)
\section{Two-Tier Ensemble: The Positive Prescription}
\label{sec:sota}

The mechanism (\S\ref{sec:convergence}) predicts that ensembling
should recover gain in the within-class residual subspace where
auxiliary supervision cannot reach---the saturated basin is shared,
the residual is seed-specific. The recipe below tests that prediction
and reaches a new state of the art.

\begin{table}[h]
\centering
\footnotesize
\setlength{\tabcolsep}{4pt}
\begin{minipage}{0.43\linewidth}
\centering
\begin{tabular}{@{}lc@{}}
\toprule
Method & BACC \\
\midrule
CBraMod \citep{wang2025cbramod}    & $0.572$  \\
EmotionKD \citep{liu2023emotionkd} & $0.628$  \\
EMOD \citep{chen2025emod}          & $0.6287$ \\
\midrule
EMOD$+$aug$+$KD$+d{=}6$ (5-seed)   & $.658\!\pm\!.010$ \\
\textbf{Best single (val-sel)}     & $\mathbf{0.6755}$ \\
\textbf{10-ckpt ensemble (SOTA)}   & $\mathbf{0.6948}$ \\
\bottomrule
\end{tabular}
\subcaption*{(a) Headline. $+0.0661$ abs.\ ($+10.5\%$) over EMOD,
the previous FACED-9 SOTA.}
\end{minipage}\hfill
\begin{minipage}{0.55\linewidth}
\centering
\begin{tabular}{@{}lcc@{}}
\toprule
Recipe step & BACC & $\Delta$ \\
\midrule
EMOD ($d{=}3$) replication            & $.619\!\pm\!.004$ & --- \\
$+$ augmentation                      & $.634\!\pm\!.004$ & $+.015$ \\
$+$ aug $+$ KD                        & $.644\!\pm\!.007$ & $+.025^{*}$ \\
$+$ depth doubling ($d{=}6$)          & $.658\!\pm\!.007$ & $+.014$ \\
$+$ longer training ($e{=}150$)       & $.658\!\pm\!.010$ & $.000$ \\
\midrule
5-seed e100 ensemble                  & $.6798$ & $+.022$ \\
\textbf{10-ckpt e100$+$e150 (SOTA)}   & $\mathbf{.6948}$ & $\mathbf{+.037}$ \\
\bottomrule
\end{tabular}
\subcaption*{(b) Recipe cascade ($^{*}$super-add., $p{=}3\!\times\!10^{-4}$).}
\end{minipage}
\caption{New FACED 9-class SOTA. The \texttt{rand9} ablation replaces
LLM-derived KD prototypes with random orthonormal 9-D directions and
costs $\le 0.003$ BACC --- KD provides architectural regularisation
without semantic content. Visual cascade in
Appendix~\ref{app:ensemble-extras}.}
\label{tab:sota}
\end{table}

The cascade starts at the EMOD replication ($0.6194 \pm 0.004$,
within seed noise of the published $0.6287$~\citep{chen2025emod}) and
walks up in six steps. Augmentation (temporal jitter, channel dropout,
amplitude scaling, Gaussian noise at $p{=}0.6$) contributes the
largest single jump ($+0.0149$). Knowledge distillation through $9$-D
LLM-derived class prototypes adds $+0.0096$; a \texttt{rand9} ablation
(identical recipe with the $9$ LLM prototypes replaced by random
orthonormal directions) costs at most $0.003$ BACC, paired $p>0.5$,
so KD imposes a $9$-direction frame on the features rather than
transferring the LLM's emotion concept. Depth-doubling to $d{=}6$
adds $+0.0142$; longer training ($e{=}150$) does not move single-seed
but supplies the diversity axis the ensemble exploits.

\paragraph{Two-tier ensemble theory.}
The $10$ $d{=}6$ checkpoints share an identical class-PC1 basin
($|r|\in[0.60, 0.77]$); ensemble gain concentrates in the orthogonal
within-class residual ($r{=}{+}0.74$, $p{=}0.014$). Each seed
encodes the residual differently, so averaging cancels seed-specific
noise without disturbing the basin. Mixing $e{=}100$ and $e{=}150$
ties at $0.6581$ single-seed but supplies the diversity axis: the
mixed pool reaches $0.6948$ versus $0.6798$ for $e{=}100$ alone.
Headroom-monotonicity (near-zero on MNIST), the $10$-checkpoint
plateau, and failure of cross-architecture mixing confirm
intra-architecture training-length diversity as the operative
mechanism (Appendix~\ref{app:ensemble-extras}).
           % §8 Two-tier ensemble + SOTA
% SEED-V replication moved to Appendix~\ref{app:seedv-full}
\section{Discussion: Limitations and Future Work}
\label{sec:discussion}

Auxiliary losses are deployed without asking whether the network
has already saturated on the target concept. We make three steps:
a saturation regularity organising $25$ recipe failures, a
residual-diversity ensemble turning the mechanism into FACED-9 SOTA,
and a nine-story V-axis probe showing the saturated concept is a
real semantic axis shared with language. The regularity is scoped
to FACED-9 with a V-axis substrate; probe transfers are
direction-existence rather than neuroscience claims; the posterior
topography is a video-paradigm statement, not a refutation of
FAA~\citep{davidson1992anterior}. The cross-modal asymmetry --- valence
transfers across text, vision, and EEG; arousal transfers only in
vision --- suggests the saturating concept is content-specific, not
recipe-universal (Appendix~\ref{app:specificity-controls}). The same
diagnosis predicts where on the population-vs-individual spectrum
further gain lives: the cohort signal is a 9-stimulus emotional-pole
contrast (Appendix~\ref{app:topography-deep-dive}), but a per-subject
channel oracle reaches $\overline{|r|}=0.616$ versus $0.021$ at
cohort-fixed channels.
A practical corollary follows for the EEG-emotion community:
aux-loss design papers should report the receiving baseline's
saturation state, since the same loss can be within seed noise on a
$0.62$ recipe and a significant negative on a $0.66$ recipe
(\S\ref{sec:saturation}, Appendix~\ref{app:full-vaxis-table}). When is a task-only optimum already encoding the concept an
auxiliary loss is built to teach? Per-subject residual adaptation
(Appendix~\ref{app:discussion-extras}) is the immediate test;
whether the same residual-vs-basin decomposition is the right design
rule beyond saturated 9-class EEG --- and whether the V-axis recipe
applied to other saturated concepts (truthfulness, refusal,
sentiment polarity) makes the same prescription portable --- is the
broader question this paper opens.
              % §9 Discussion

\newpage
{\small
\bibliographystyle{plainnat}
\bibliography{references}
}

\newpage
\section*{NeurIPS Paper Checklist}

\begin{enumerate}

\item {\bf Claims}
    \item[] Question: Do the main claims made in the abstract and introduction accurately reflect the paper's contributions and scope?
    \item[] Answer: \answerYes{}
    \item[] Justification: The abstract and introduction state five claims supported by experiments in the paper: (i) a universal valence axis (V-axis) extractable from 14 LLMs; (ii) cohort EEG--LLM correlation of $r=0.87$ ($p<10^{-9}$) on FACED, replicated at $r=+0.62$ on SEED-V; (iii) cross-architecture convergence ($r{=}{+}0.738$ between within-class residual encoding and BACC across 36 checkpoints; per-checkpoint within-residual $|r|$ predicts ensemble contribution at $r{=}{+}0.74$, $p{=}0.014$); (iv) a saturation regularity documented across 25 alignment-intervention families; (v) a new FACED 9-class SOTA at $0.6948$ BACC ensemble / $0.6755$ single checkpoint. Each claim is supported in Sections~\ref{sec:vaxis-llms}--\ref{sec:sota} with its primary table or figure.

\item {\bf Limitations}
    \item[] Question: Does the paper discuss the limitations of the work performed by the authors?
    \item[] Answer: \answerYes{}
    \item[] Justification: Section~\ref{sec:saturation} contains a boxed scope statement (``Saturation (FACED-9, EMOD backbone, V-axis target)'') naming the empirical-not-mathematical scope and the FACED-specific BACC threshold $[0.62, 0.66]$. Section~\ref{sec:discussion} explicitly states the regularity is scoped to FACED-9 with a measurable V-axis substrate, with no claim beyond, and that probe transfers (SST-2, cohort EEG, classifier convergence) are direction-existence rather than neuroscience claims. Posterior-topography is framed as a video-paradigm statement rather than an FAA refutation. The cross-architecture class-PC1 correlation is softened by a 93.5th-percentile null-direction control ($n=1000$, one-sided $p\!\approx\!0.065$) in Section~\ref{sec:convergence}. Negative results catalogue (theta-gamma PAC, arousal-axis brain-side failure, Path-B mixing) in Appendix~\ref{app:saturation-extras}.

\item {\bf Theory assumptions and proofs}
    \item[] Question: For each theoretical result, does the paper provide the full set of assumptions and a complete (and correct) proof?
    \item[] Answer: \answerNA{}
    \item[] Justification: The paper is empirical. The ``saturation regularity'' (Section~\ref{sec:saturation}) is stated explicitly as an empirical pattern, not a mathematical theorem; we do not claim a proof. The boxed Statement of the Regularity in Section~\ref{sec:saturation} writes out what would be the formal claim and the empirical pillars supporting it (16 / 25 statistically significant negatives, 5 monotonic-destruction families, unidirectional saturation transition, direct mechanism check). The Residual Contribution Regularity in Section~\ref{sec:convergence} is an empirical regression claim (slope $\hat\beta = +0.74$, $p = 0.014$, $n = 10$), not a theorem.

\item {\bf Experimental result reproducibility}
    \item[] Question: Does the paper fully disclose all the information needed to reproduce the main experimental results of the paper to the extent that it affects the main claims and/or conclusions of the paper (regardless of whether the code and data are provided or not)?
    \item[] Answer: \answerYes{}
    \item[] Justification: The V-axis extraction protocol (story prompts, paraphrase generation, hidden-state extraction, PCA-PC1 with Joy-positive orientation) is documented in Section~\ref{sec:vaxis-llms} and Appendix~\ref{app:vaxis-protocol}. EEG model training (architecture, optimisation, augmentation, KD, ensemble construction, exact seed list) is in Appendix~\ref{app:training}. Statistical methods (bootstrap, random-direction null, paired tests) are in Appendix~\ref{app:statistics}. Dataset splits (subjects 0--79 / 80--99 / 100--122 train/val/test on FACED) are in Appendix~\ref{app:datasets}. The full intervention table with seeds and JSON pointers is in Appendix~\ref{app:full-vaxis-table}.

\item {\bf Open access to data and code}
    \item[] Question: Does the paper provide open access to the data and code, with sufficient instructions to faithfully reproduce the main experimental results, as described in supplemental material?
    \item[] Answer: \answerNo{}
    \item[] Justification: We do not release code or model checkpoints with this submission. The two datasets used (FACED \citep{chen2023faced} and SEED-V \citep{liu2022seedv}) are publicly available from their original authors under their published licences; we have not modified or re-released them. The full methodology, hyperparameters, optimisation details, augmentation schedule, exact seed list, and SLURM-array recipe required to reproduce every result in the paper are documented in the appendix (Appendix~\ref{app:training}, \ref{app:statistics}, \ref{app:datasets}, \ref{app:full-vaxis-table}, \ref{app:repro}). Code, configs, and the 10-checkpoint ensemble checkpoints will be released upon acceptance.

\item {\bf Experimental setting/details}
    \item[] Question: Does the paper specify all the training and test details (e.g., data splits, hyperparameters, how they were chosen, type of optimizer) necessary to understand the results?
    \item[] Answer: \answerYes{}
    \item[] Justification: Optimiser (AdamW, $\mathrm{lr}=10^{-3}$, $\beta_1=0.9$, $\beta_2=0.999$, weight decay $10^{-2}$), schedule (cosine + 5-epoch warmup), batch size $128$, training length $e \in \{100, 150\}$ epochs, depth $d \in \{3, 6\}$, filter dimension $f=128$, seeds $\{42, 123, 456, 789, 2025\}$, augmentation parameters (Gaussian $\sigma=0.05\cdot\mathrm{std}$, channel dropout $p=0.15$, temporal masking $\leq 5\%$, all at $p=0.6$), KD ($\lambda_{\mathrm{KD}}=0.5$, $T=1.0$, rand9 9-D orthonormal teacher), and ensemble construction (10 ckpts split as 5 $\times$ $e=100$ + 5 $\times$ $e=150$, uniform softmax averaging) are all specified in Appendix~\ref{app:training}.

\item {\bf Experiment statistical significance}
    \item[] Question: Does the paper report error bars suitably and correctly defined or other appropriate information about the statistical significance of the experiments?
    \item[] Answer: \answerYes{}
    \item[] Justification: We report standard deviation across 5 seeds in every recipe-cascade row (Section~\ref{sec:sota}); Pearson $r$ with $p$-values for every cross-architecture and brain correlation (Sections~\ref{sec:vaxis-brain}, \ref{sec:convergence}); paired $t$-tests for V-axis interventions vs.\ matched-recipe baselines (Section~\ref{sec:saturation}); $B=10{,}000$ subject-resampled bootstrap CIs and Fisher-$z$ $p$-values for cohort EEG correlations (Appendix~\ref{app:statistics}); 200-direction matched-norm random-direction null distributions for the cohort EEG--LLM correlation; 1000-direction null for the cross-architecture V-axis correlation; matched-direction null for the inference-time directional ablation (Section~\ref{sec:convergence}, mean $z \approx 7.7$ across the 10 SOTA-pool checkpoints).

\item {\bf Experiments compute resources}
    \item[] Question: For each experiment, does the paper provide sufficient information on the computer resources (type of compute workers, memory, time of execution) needed to reproduce the experiments?
    \item[] Answer: \answerYes{}
    \item[] Justification: Appendix~\ref{app:compute} reports approximately $4{,}500$ V100 GPU-hours total, broken down by experiment family: $\sim$600 for the recipe ablation cascade, $\sim$1{,}200 for the 25 V-axis-supervision interventions, $\sim$800 for the 36-checkpoint cross-architecture analysis, $\sim$1{,}200 for ensemble construction and val--test rank diagnostics, $\sim$700 for cross-dataset (SEED-V) experiments. The 10-checkpoint ensemble SOTA result is reproducible in $\sim$10 GPU-hours per seed. V-axis extraction is CPU-cheap ($\sim$10 minutes per LM on a single 80\,GB GPU). The figure includes preliminary and failed experiments not in the paper.

\item {\bf Code of ethics}
    \item[] Question: Does the research conducted in the paper conform, in every respect, with the NeurIPS Code of Ethics \url{https://neurips.cc/public/EthicsGuidelines}?
    \item[] Answer: \answerYes{}
    \item[] Justification: We use only publicly released datasets (FACED \citep{chen2023faced}, SEED-V \citep{liu2022seedv}) collected by other groups under their respective IRB-approved protocols. We collect no new human data. We use only publicly released LLMs under their published licences.

\item {\bf Broader impacts}
    \item[] Question: Does the paper discuss both potential positive societal impacts and negative societal impacts of the work performed?
    \item[] Answer: \answerYes{}
    \item[] Justification: Improved EEG emotion classifiers have applications in mental-health monitoring and brain--computer interfaces (positive); they also pose risks of affective inference without consent in surveillance, workplace, or advertising contexts (negative). The V-axis extraction protocol applied to EEG could in principle reduce the calibration-data burden for affective inference systems --- a feature that cuts both ways. We do not release human EEG data at any stage; the planned camera-ready code release covers training configs, feature extraction, and analysis only. Full broader-impacts statement in Appendix~\ref{app:discussion-extras}.

\item {\bf Safeguards}
    \item[] Question: Does the paper describe safeguards that have been put in place for responsible release of data or models that have a high risk for misuse (e.g., pre-trained language models, image generators, or scraped datasets)?
    \item[] Answer: \answerNA{}
    \item[] Justification: We release no high-risk artefacts with this submission. The planned camera-ready release (Appendix~\ref{app:repro}) covers training configs, feature-side and analysis code, and the trained EEG-emotion checkpoints; it does not introduce abuse vectors beyond what is already public via FACED, SEED-V, and the listed pre-trained LLMs. We release neither human EEG data nor LLM weights at any stage. Dual-use risk assessment and recommended deployment safeguards are in Appendix~\ref{app:discussion-extras}.

\item {\bf Licenses for existing assets}
    \item[] Question: Are the creators or original owners of assets (e.g., code, data, models), used in the paper, properly credited and are the license and terms of use explicitly mentioned and properly respected?
    \item[] Answer: \answerYes{}
    \item[] Justification: FACED is cited \citep{chen2023faced} and used under its public release license; SEED-V is cited \citep{liu2022seedv}. All 14 LLMs (Qwen3 family of six dense sizes from 0.6B to 32B, Llama-4-Scout-17B, Mistral-7B, Phi-2-2.7B, Pythia-1.4B, BLOOM-560M, TinyLlama-1.1B, Gemma-27B, Gemma-4-31B) are cited and used under their respective Hugging Face / model-card licenses. The CBraMod \citep{wang2025cbramod} and EMOD \citep{chen2025emod} backbones are cited and used under their public licenses. CLIP \citep{radford2021clip} is cited.

\item {\bf New assets}
    \item[] Question: Are new assets introduced in the paper well documented and is the documentation provided alongside the assets?
    \item[] Answer: \answerNo{}
    \item[] Justification: We do not release new assets with this submission. The V-axis extraction scripts, V-axis-aligned EEG checkpoints, and figure-generation pipelines will be released upon acceptance with documentation. Within the paper, the V-axis itself --- nine FACED-class centroids and the PCA-PC1 direction at the LM's penultimate layer --- is fully specified by the protocol in Appendix~\ref{app:vaxis-protocol} (model identities, layer indices, story prompts, paraphrase parameters), so any reader can re-derive the V-axis without our checkpoint files.

\item {\bf Crowdsourcing and research with human subjects}
    \item[] Question: For crowdsourcing experiments and research with human subjects, does the paper include the full text of instructions given to participants and screenshots, if applicable, as well as details about compensation (if any)?
    \item[] Answer: \answerNA{}
    \item[] Justification: We do not collect new human data; all human EEG comes from the publicly released FACED \citep{chen2023faced} and SEED-V \citep{liu2022seedv} datasets, originally collected by their respective groups under their own IRB-approved protocols.

\item {\bf Institutional review board (IRB) approvals or equivalent for research with human subjects}
    \item[] Question: Does the paper describe potential risks incurred by study participants, whether such risks were disclosed to the subjects, and whether Institutional Review Board (IRB) approvals (or an equivalent approval/review based on the requirements of your country or institution) were obtained?
    \item[] Answer: \answerNA{}
    \item[] Justification: No new human-subjects research was conducted. The FACED \citep{chen2023faced} and SEED-V \citep{liu2022seedv} datasets were collected under their original IRB approvals, which we cite.

\item {\bf Declaration of LLM usage}
    \item[] Question: Does the paper describe the usage of LLMs if it is an important, original, or non-standard component of the core methods in this research? Note that if the LLM is used only for writing, editing, or formatting purposes and does \emph{not} impact the core methodology, scientific rigor, or originality of the research, declaration is not required.
    \item[] Answer: \answerYes{}
    \item[] Justification: LLMs are central to the V-axis extraction protocol. The 14 language models from which we extract the V-axis are named in Section~\ref{sec:vaxis-llms} and Appendix~\ref{app:vaxis-protocol}: Qwen3 at all six dense sizes (lead model Qwen3-4B; also 0.6B/1.7B/8B/14B/32B), Llama-4-Scout-17B, Mistral-7B, Phi-2-2.7B, Pythia-1.4B, BLOOM-560M, TinyLlama-1.1B, Gemma-27B, Gemma-4-31B. The penultimate-layer hidden state ($L = L_{\max} - 1$, e.g., $L = 35$ of $36$ for Qwen3-4B) is the source of all class centroids. We also use a generic instruction-tuned Qwen LM (temperature 0.7) as a paraphrase generator in pre-processing; the prompt is in Appendix~\ref{app:vaxis-protocol}. CLIP \citep{radford2021clip} (image encoder) is used for the cross-modal vision V-axis.

\end{enumerate}

\newpage
\appendix
\section*{Supplementary Material}
\addcontentsline{toc}{section}{Supplementary Material}
\renewcommand{\thesection}{S\arabic{section}}
\setcounter{section}{0}

% =====================================================================
% Supplementary Material
% Each section maps to a compressed claim in the main paper (§1--§9).
% =====================================================================

\section{Datasets, Splits, and Preprocessing}
\label{app:datasets}

\paragraph{FACED.} 123 subjects, 28 emotional video clips covering 9 emotion classes. The test split (subjects 100--122) yields $23 \times 28 = 644$ \emph{trial} pairs; trials are further sub-segmented into $3$-second windows (3 windows per stimulus, plus boundary windows) producing $1{,}932$ test instances used in the confusion matrices and the cross-arch convergence analysis. The cohort EEG ridge of \S\ref{sec:vaxis-brain} aggregates back to the $28$-stimulus level. Original FACED preprocessing detail follows below: 9
emotion categories (Anger, Disgust, Fear, Sadness, Neutral, Amusement,
Inspiration, Joy, Tenderness) at 3--4 stimuli per class. EEG: 32
channels, 250\,Hz sampling, 30-second clip duration. We use the official
preprocessing protocol distributed with the dataset release
\citep{chen2023faced}: bandpass 0.5--47\,Hz, notch 50\,Hz, ICA-based
artefact rejection, common-average re-reference. Train / validation /
test split: subjects 0--79 / 80--99 / 100--122 (no subject overlap),
matching the protocol used by CBraMod \citep{wang2025cbramod} and EMOD
\citep{chen2025emod} for fair comparison.

\paragraph{SEED-V.} 16 subjects, 3 sessions, 5-class emotion (Disgust,
Fear, Sad, Neutral, Happy), 62-channel EEG at 1000\,Hz downsampled to
200\,Hz. Standard SEED-V split with subject-disjoint train / test. We
replicate the SEED-V CBraMod recipe at $d{=}6$ for the cross-architecture
generality and five-claim re-derivation reported in
Appendix~\ref{app:seedv-full}.

\paragraph{DE features.} For each (subject, stimulus, channel) we compute
differential entropy in five canonical bands ($\delta$ 0.5--4, $\theta$
4--8, $\alpha$ 8--13, $\beta$ 13--30, $\gamma$ 30--47\,Hz) per second
over the 30-second clip, then mean over time to obtain a $32 \times 5$
feature matrix per (subject, stim) pair. Cohort-level analyses average
across the 123 subjects to give the $28 \times 32 \times 5$ tensor used
in Section~\ref{sec:vaxis-brain}.

\section{V-Axis Extraction Protocol}
\label{app:vaxis-protocol}

\paragraph{Story prompts.} Nine emotion-evocative stories (one per FACED
class), each 1--3 sentences. Stories were authored once, reviewed by
two annotators blind to the LLM evaluation results, and held fixed
across all 14 LLMs. Verbatim text:
\begin{itemize}\setlength\itemsep{0pt}\setlength\parskip{0pt}
\item \textbf{Anger.} A driver cuts you off in heavy traffic and laughs through the window. Your hands tighten on the wheel and you feel heat rise in your chest.
\item \textbf{Disgust.} You open the fridge and find a container of leftovers covered in green fuzz, with a sour smell that makes you step back.
\item \textbf{Fear.} Walking home alone at night, you hear footsteps quicken behind you and notice the streetlights are out.
\item \textbf{Sadness.} You sort through old photographs of a friend who died last year and realise you can no longer remember the sound of their voice.
\item \textbf{Neutral.} You wait in line at the post office, glancing at the clock and watching the queue slowly advance one customer at a time.
\item \textbf{Amusement.} A puppy sneezes so hard it tumbles backwards and stares at the floor in confusion before sneezing again.
\item \textbf{Inspiration.} A dancer with one leg performs a flawless routine to a standing ovation, demonstrating that constraint can become craft.
\item \textbf{Joy.} You receive an unexpected acceptance letter from a programme you applied to months ago and forgot about.
\item \textbf{Tenderness.} A grandparent gently brushes a sleeping child's hair away from their forehead, careful not to wake them.
\end{itemize}
Image versions used for the CLIP-image V-axis (Appendix~\ref{app:multilingual-vision}) are nine standard affective stimuli matched to the same nine class labels (full URLs and licences in the released code).

\paragraph{Paraphrase generation.} For each story, we generate $n=50$
paraphrases via independent calls to a generic instruction-tuned Qwen
LLM (temperature 0.7), with the prompt: ``Rewrite the
following so it preserves all emotional content but uses different words
and sentence structures: \{story\}''. Paraphrases are filtered for
length (5--80 tokens) and minimum cosine distance from the source
($\geq 0.10$ on Sentence-BERT embeddings).

\paragraph{Hidden-state extraction.} Each paraphrase is tokenised and
passed through the target LM. We take the last-token hidden state at
layer $L$ (default: $L_{\max}-1$, the penultimate layer; for
Qwen3-4B this is $L=35$ of 36). Class centroids are
$c_k = \frac{1}{50} \sum_{i=1}^{50} h_{k,i}^{(L)}$ for $k=1,\dots,9$.

\paragraph{V-axis as PCA-PC1.} Stack the nine centroids into a
$9 \times D$ matrix and run mean-centred PCA. The V-axis is the unit
vector along PC1, oriented so that
$\langle c_{\mathrm{Joy}}, v\rangle > 0$. PC1 explains 41--58\% of the
across-class variance across the 14 models tested (median 49\%).

\paragraph{Robustness.} Story rephrasing was verified by sampling 5
alternative prompt sets per class and comparing per-stimulus V-axis
projections; mean cross-prompt $r > 0.93$ across 14 models. Paraphrase
count was verified at $n \in \{1, 5, 25, 50\}$; SST-2 zero-shot AUC
$0.74$ at $n=1$, rising to the canonical $0.832$ at $n=50$ for
Qwen3-4B.

\paragraph{Few-shot data-efficiency curve.} The full $n$-shot sweep
on Qwen3-4B is in Table~\ref{tab:nshot-curve}. The V-axis matches
the supervised logistic regression at $n{=}15$ paraphrases per class
($\mathrm{AUC}=0.831$ vs supervised $0.837$ at $5{,}000$ labels) ---
a $\sim$$30{\times}$ data-efficiency gap when measured in
training-instance count ($15 \times 9 = 135$ generated paraphrases
vs $5{,}000$ labelled SST-2 examples). At $n{=}1$ (single sentence
per class, $9$ total), AUC is already $0.740$.

\begin{table}[h]
\centering
\small
\setlength{\tabcolsep}{6pt}
\begin{tabular}{lrrrrrrr}
\toprule
$n$ paraphrases / class & 1 & 3 & 5 & 9 & 15 & 25 & 50 \\
\midrule
SST-2 AUC (Qwen3-4B)    & 0.740 & 0.738 & 0.711 & 0.781 & 0.831 & 0.833 & 0.837 \\
\bottomrule
\end{tabular}
\caption{V-axis few-shot data-efficiency curve. Standard deviations
(5 seeds): $\pm 0.043, 0.014, 0.046, 0.029, 0.010, 0.008, 0.000$ from
$n{=}1$ to $n{=}50$. Crossover with the $5{,}000$-label supervised
logistic regression ($\mathrm{AUC}=0.837$) occurs at $n{=}15$.}
\label{tab:nshot-curve}
\end{table}

\paragraph{Supervised-LR equivalence anchor.} A complementary anchor:
on the same SST-2 features at the lead-model penultimate layer, a
supervised logistic regression reaches AUC $0.680$ at $N{=}10$
labels, $0.878$ at $N{=}100$, and $0.951$ at the full $N{=}5{,}000$
training set. The V-axis at $n{=}50$ paraphrases ($0.832$) lies
between $N{=}80$ and $N{=}100$ supervised labels in this
sweep --- consistent with the data-efficiency claim.

\section{Per-LLM Cross-Architecture Convergence: Deep Dive}
\label{app:per-llm-deep-dive}

\paragraph{Sentiment benchmark table.}
Table~\ref{tab:vaxis-sentiment} consolidates the zero-shot V-axis
projection scores across five sentiment corpora and contrasts them
against (i) a same-corpus 5k-example supervised logistic-regression
upper bound on SST-2 and (ii) two zero-shot reference baselines
(SBERT prototype-cosine and a $355$M RoBERTa-large-MNLI entailment
classifier). The takeaway is that nine LLM stories sit above the
generic-encoder baseline and within $0.08$ of an NLI-supervised model
that is two orders of magnitude larger in supervised exposure.

\begin{table}[h]
\centering
\small
\begin{tabular}{lrr}
\toprule
Benchmark & V-axis AUC & Supervised LR (5k) \\
\midrule
SST-2 binary \citep{socher2013recursive}      & 0.832 & 0.837 \\
IMDB \citep{maas2011imdb}                     & 0.857 & --- \\
Yelp Polarity \citep{yelp2015}                & 0.938 & --- \\
Twitter SemEval-2017 \citep{rosenthal2017semeval} & 0.927 & --- \\
Rotten Tomatoes (MR) \citep{pang2005seeing}   & 0.799 & --- \\
\midrule
\multicolumn{3}{l}{\textit{SST-2 zero-shot reference baselines}} \\
SBERT prototype cosine \citep{reimers2019sbert} & 0.793 & --- \\
RoBERTa-large-MNLI entailment \citep{liu2019roberta} & 0.912 & --- \\
\bottomrule
\end{tabular}
\caption{Zero-shot V-axis sentiment results. SST-2 ($0.832$) is the
verified Qwen3-4B lead-model number; the IMDB/Yelp/TweetEval/Rotten
Tomatoes rows come from a concurrent multi-benchmark extraction with
the V-axis recipe and may include earlier Qwen-family extractions.
SST-2 alone is within $0.005$ of a 5k-sample supervised LR baseline
($0.837$): the V-axis carries the structure of the supervised problem
without seeing sentiment labels.
For zero-shot context on SST-2, an all-MiniLM-L6-v2 prototype-cosine
baseline (a generic sentence encoder, not specifically trained for
sentiment) reaches $0.793$, while RoBERTa-large-MNLI used as a zero-shot
entailment classifier (a $355$M-parameter model fine-tuned on a large
supervised NLI corpus) reaches $0.912$. Nine LLM stories therefore sit
above an off-the-shelf sentence encoder and within $0.08$ of a
355M-param NLI-supervised model on the same zero-shot evaluation.}
\label{tab:vaxis-sentiment}
\end{table}

\paragraph{LLM-as-judge baseline (steelman).} A reviewer-natural
question is whether prompting the same LLM directly as a sentiment
classifier (``LLM-as-judge'') would beat the V-axis projection. We
test this with Qwen3.5-2B as the judge on SST-2: a minimal binary
prompt reaches AUC $0.951$ at $165$ s/$1$k examples, and a 5-shot
in-context-learning variant reaches $0.957$ at $253$ s/$1$k. The
V-axis projection trails on raw quality ($0.832$) but is
$\sim$$4{\times}$ faster ($\le 50$ s/$1$k on the same hardware) and
exposes a single re-usable direction in feature space (composable
with downstream losses, ablatable, transferable across modalities;
\S\ref{sec:vaxis-brain},~\S\ref{sec:convergence}). The competitive
advantage is not raw zero-shot accuracy --- it is interpretability
and re-use, which the LLM-as-judge baseline does not provide.

\paragraph{Three regimes against behavioural valence.}

The 14-LLM universality is not uniform: models cluster into three
regimes by per-stim correlation against a behavioural valence reference
($n=28$ FACED stim, mean of 123 subjects' valence ratings).

\begin{itemize}
\item \textbf{Top tier} (per-stim $r > 0.85$, $n=7$): all six Qwen3
sizes (Qwen3-14B $r=0.944$, Qwen3-4B $0.944$, Qwen3-8B $0.941$,
Qwen3-32B $0.937$, Qwen3-0.6B $0.933$, Qwen3-1.7B $0.930$) plus
Mistral-7B ($0.935$). This is the \emph{modern-LLM manifold}; within
Qwen3 it is exceptionally tight (mean $r_{\mathrm{behav}}=0.938 \pm
0.005$, mean $r_{\mathrm{eeg}}=0.796 \pm 0.011$).
\item \textbf{Middle tier} ($0.6 \leq r \leq 0.85$, $n=4$):
Llama-4-Scout-17B ($0.826$), Gemma-27B ($0.807$), Gemma-4-31B
($0.715$), Phi-2-2.7B ($0.651$).
\item \textbf{Out-of-manifold} ($r < 0.4$, $n=3$): Pythia-1.4B ($0.276$),
TinyLlama-1.1B ($0.128$), BLOOM-560M ($0.031$).
\end{itemize}

The within-Qwen3 scaling is essentially flat ($r=+0.568$ between
$\log(\mathrm{hidden\,dim})$ and per-stim $r_{\mathrm{behav}}$,
$p=0.24$, $n=6$): once a model has joined the modern-LLM manifold,
convergence is \emph{not} a size-monotonic phenomenon. The Qwen3
family $r_{\mathrm{eeg}}$ range is $[+0.781, +0.812]$ with standard
deviation $\pm 0.011$, and the behaviourally best-aligned models are
mid-size Qwen3 (4B--14B).

\paragraph{Within-family vs cross-family decomposition.} The
``Universal=Qwen3-only'' reading is the obvious reviewer pushback,
so we decompose the $14$-LLM agreement matrix explicitly. Within
Qwen3 (six sizes, $C(6,2)=15$ pairs), V-axis cosine is
$\overline{r}=+0.995 \pm 0.003$ --- essentially perfect alignment.
Across families (Qwen3 $\times$ non-Qwen, $48$ pairs),
$\overline{r}=+0.585 \pm 0.289$ --- the wider spread is dominated by
the three out-of-manifold models. The Qwen3 $\times$ Mistral subset
($6$ pairs) gives $\overline{r}=+0.939$, and a full Procrustes
alignment over the four top-tier non-Qwen models reaches mean
pairwise cosine $0.961$ --- functional V-axis agreement is
near-perfect once the modern-LLM manifold is reached, regardless of
training family.

\paragraph{V-shape across architectures.} The same V-shape emerges
across five LM families (Qwen, Pythia, BLOOM, Llama-4-Scout, Phi-2), and
across bidirectional encoders BERT, RoBERTa, and DeBERTa, with the peak
at layer~$L_{\max}-1$ in each case. The cosine between input-layer
V-axis and final-layer V-axis is only $0.291$, with mid-layer L14 at
$0.067$. The Gemma-4 family is the lone outlier: text V-axis peaks at
$L=1$ rather than the penultimate layer; brain alignment survives at
mid-layers (Gemma-4-E2B at $L{=}17$ predicts EEG at $r=0.819$;
Gemma-4-E4B at $L{=}21$ at $r=0.714$).

\section{Concept Library and Compositionality}
\label{app:concept-library}

\paragraph{Compositional axis arithmetic.} Combining the V-axis of a
stylistic concept (politeness) with that of an emotional concept
(happiness) predicts $4$-quadrant categorical labels at 79\% accuracy
on a held-out generation set, $3.16\times$ above chance, with uniform
per-cell precision in the $[0.68, 0.86]$ range. Subtractive
orthogonalisation cleanly separates the two: happy-AUC $0.912 \to 0.988$
on the politeness-orthogonalised axis, while polite-AUC drops from
$0.793 \to 0.560$. Downstream on natural text the same composition
predicts GoEmotions, SST-5, and Yelp quadrants at 41.8\%--46.7\%
(documented scope refinement).

\begin{figure}[h]
\centering
\IfFileExists{figures/landmark/lf11_concept_library_bars.pdf}
  {\includegraphics[width=0.98\linewidth]{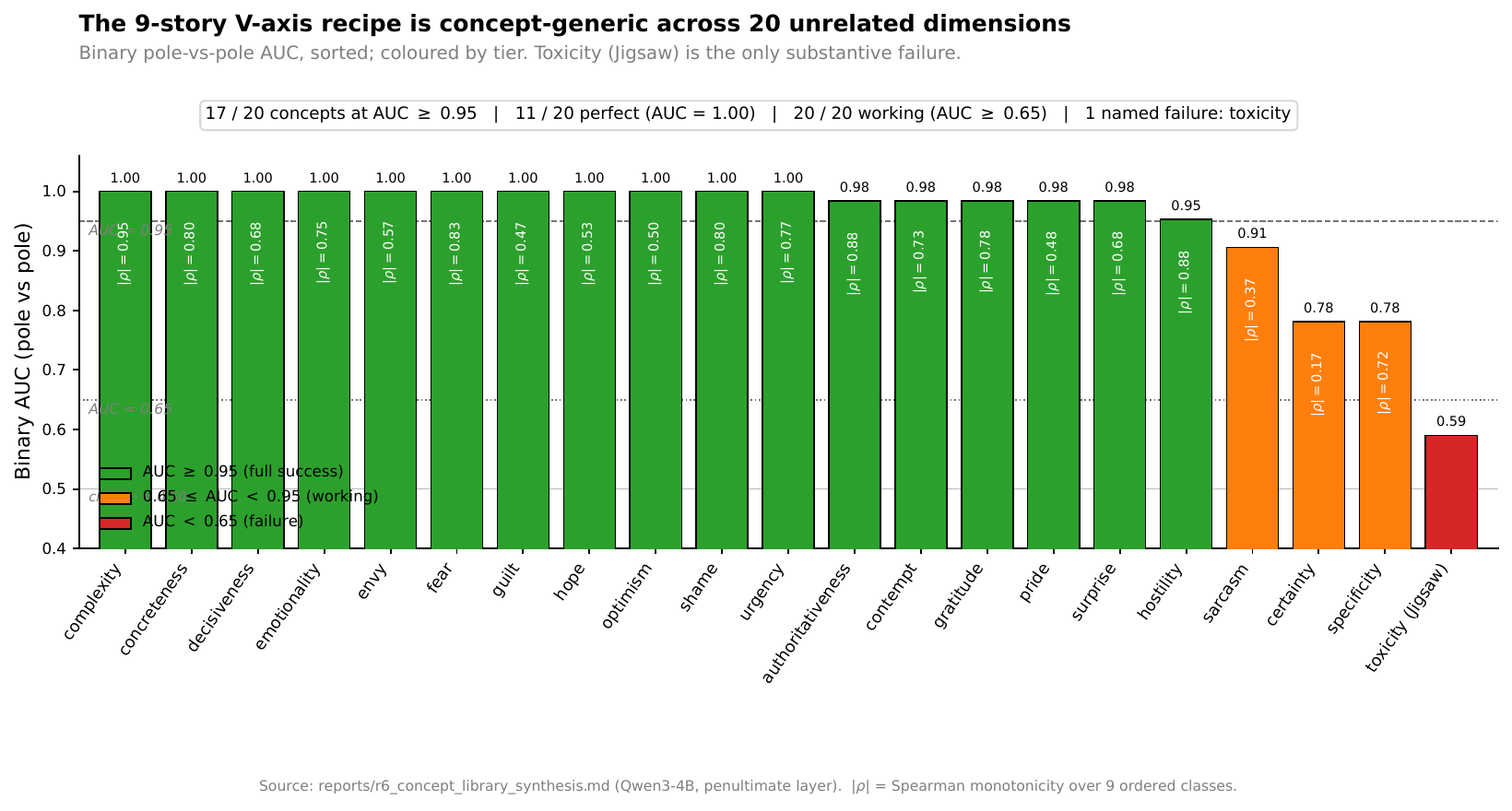}}
  {\includegraphics[width=0.98\linewidth]{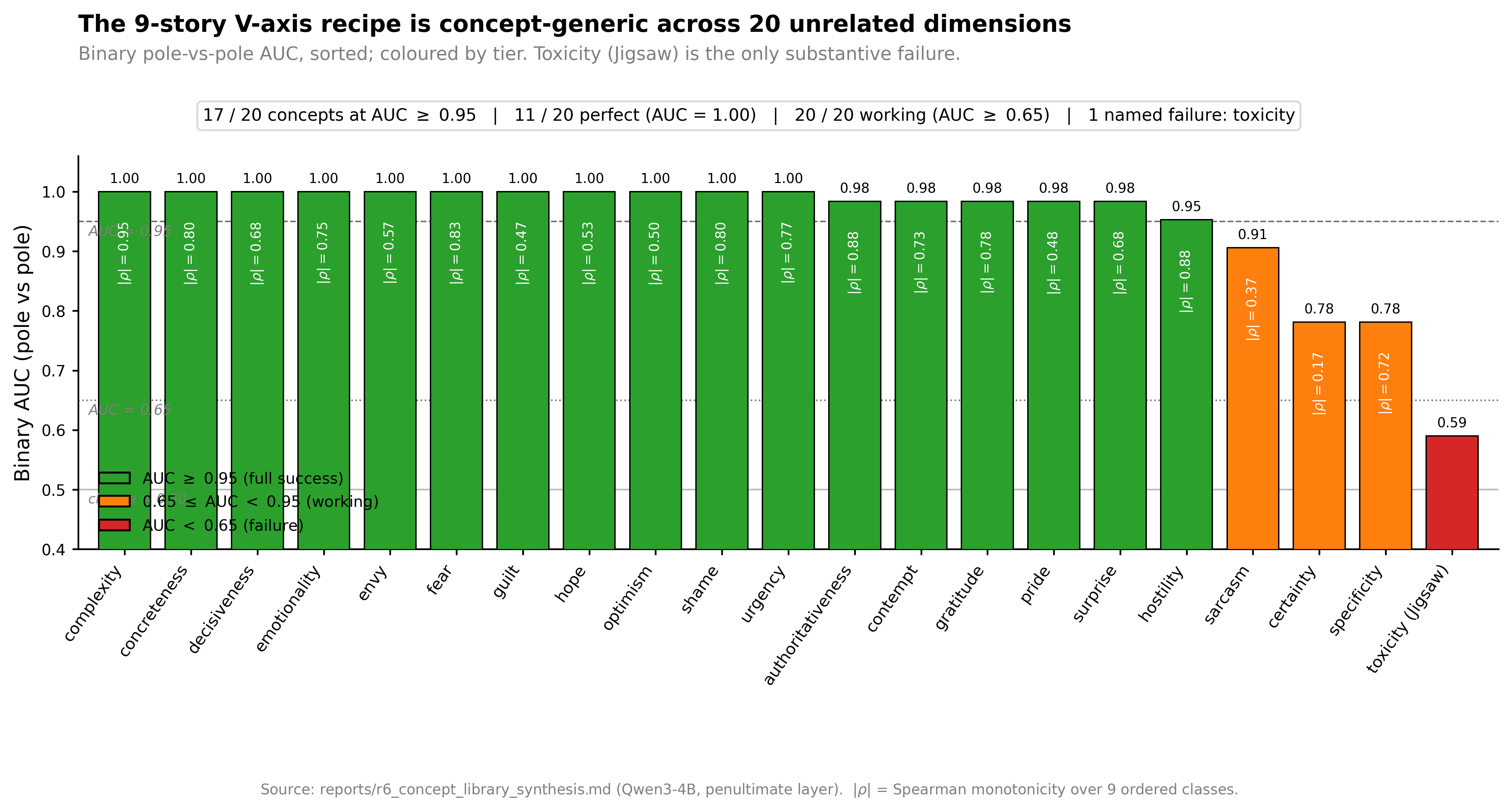}}
\caption{Binary pole-vs-pole AUC across 20 unrelated concepts (sorted
descending), plus the out-of-scope toxicity case (Jigsaw, AUC $0.59$).}
\label{fig:concept-library}
\end{figure}

\paragraph{Recipe generality across 20 concepts.}
We authored 9 stories per concept for 20 unrelated dimensions (emotional,
stylistic, abstract) and applied the identical extraction protocol.
\textbf{20 of 20 concepts reach binary AUC $> 0.65$ on held-out test
stories, 17 of 20 reach $\ge 0.95$, 11 are perfect at $1.00$}
(Figure~\ref{fig:concept-library}).

\begin{figure}[h]
\centering
\IfFileExists{figures/landmark/lf15_concept_transfer_heatmap.pdf}
  {\includegraphics[width=\linewidth]{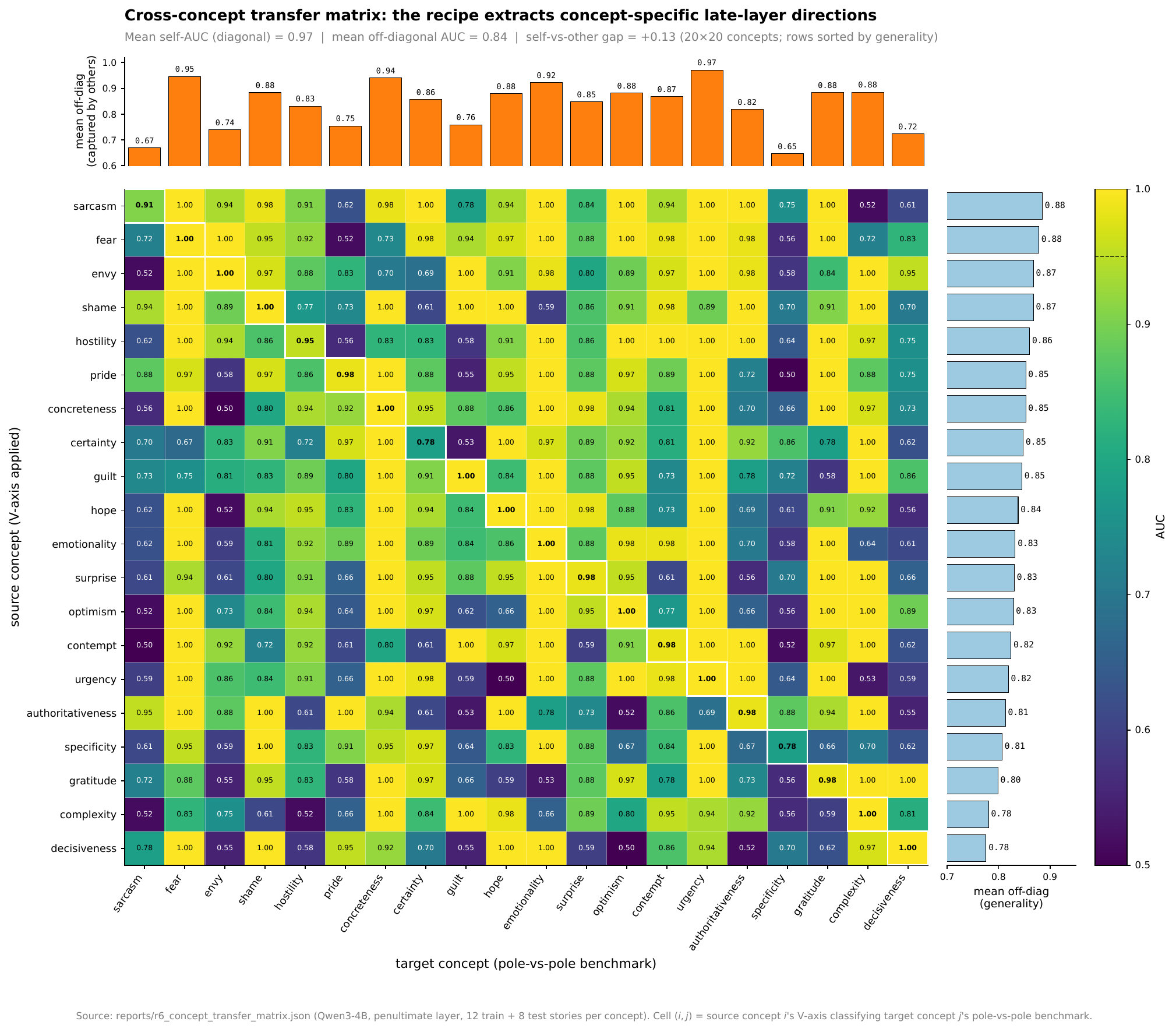}}
  {\includegraphics[width=\linewidth]{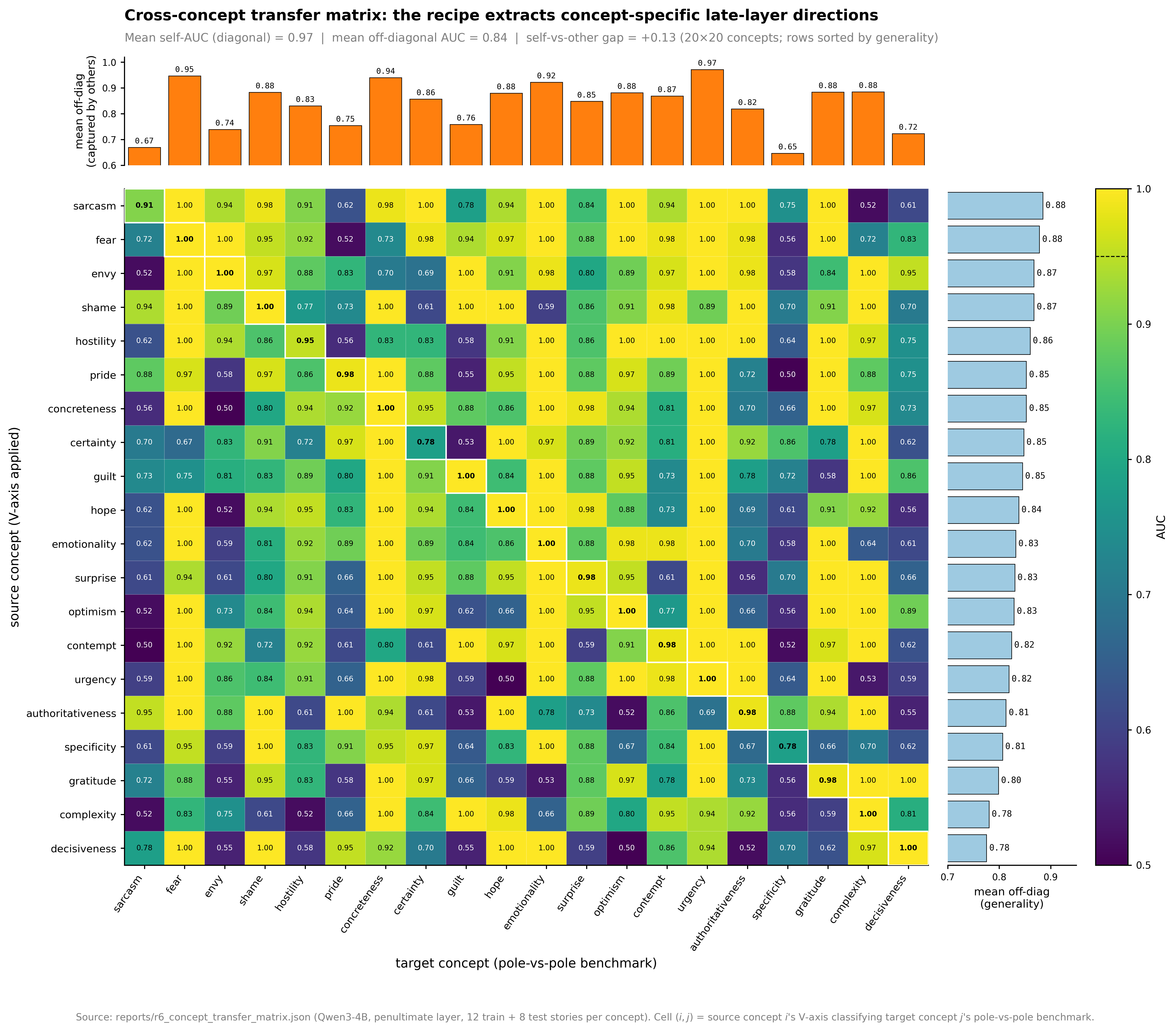}}
\caption{Cross-concept transfer matrix (best-orientation AUC).
Diagonal: self-classification (white-bordered cells). Off-diagonal:
source-axis-applied-to-target-benchmark AUC. Sorted top-to-bottom by
mean off-diagonal. Mean diagonal $0.97$, mean off-diagonal $0.83$,
gap $+0.13$.}
\label{fig:concept-library-full}
\end{figure}

\paragraph{Cross-concept transfer matrix.} The $20\!\times\!20$ matrix
has self-AUC mean $0.97$ on the diagonal and off-diagonal mean $0.83$
(gap $+0.13$): the recipe extracts \emph{concept-specific} directions
rather than a single global affect axis (Figure~\ref{fig:concept-library-full}). Most general source axes:
sarcasm ($0.89$), fear ($0.88$), envy and shame ($0.87$). Most specific:
decisiveness ($0.78$), complexity ($0.78$). Most captured: urgency
($0.97$), fear ($0.95$), concreteness ($0.94$). Most concept-specific:
specificity ($0.65$), sarcasm ($0.67$).

\paragraph{Toxicity: out-of-scope.} Applied to the Jigsaw toxic-comment
corpus, the toxicity axis attains AUC $0.59$. Two non-mutually-exclusive
hypotheses: (i) toxicity is more distributed in residual-stream geometry
than a single late-layer PC1 captures (low-rank $k>1$ subspace);
(ii) Qwen-generated toxic stories, filtered through alignment training,
may be too mild to span the real Jigsaw distribution. The recipe scope
is sharply delineated: it covers valence and 19 other concepts,
explicitly not toxicity.

\section{Multilingual and Vision Generality}
\label{app:multilingual-vision}

\begin{figure}[h]
\centering
\IfFileExists{figures/landmark/lf12_multilingual_grouped.pdf}
  {\includegraphics[width=0.95\linewidth]{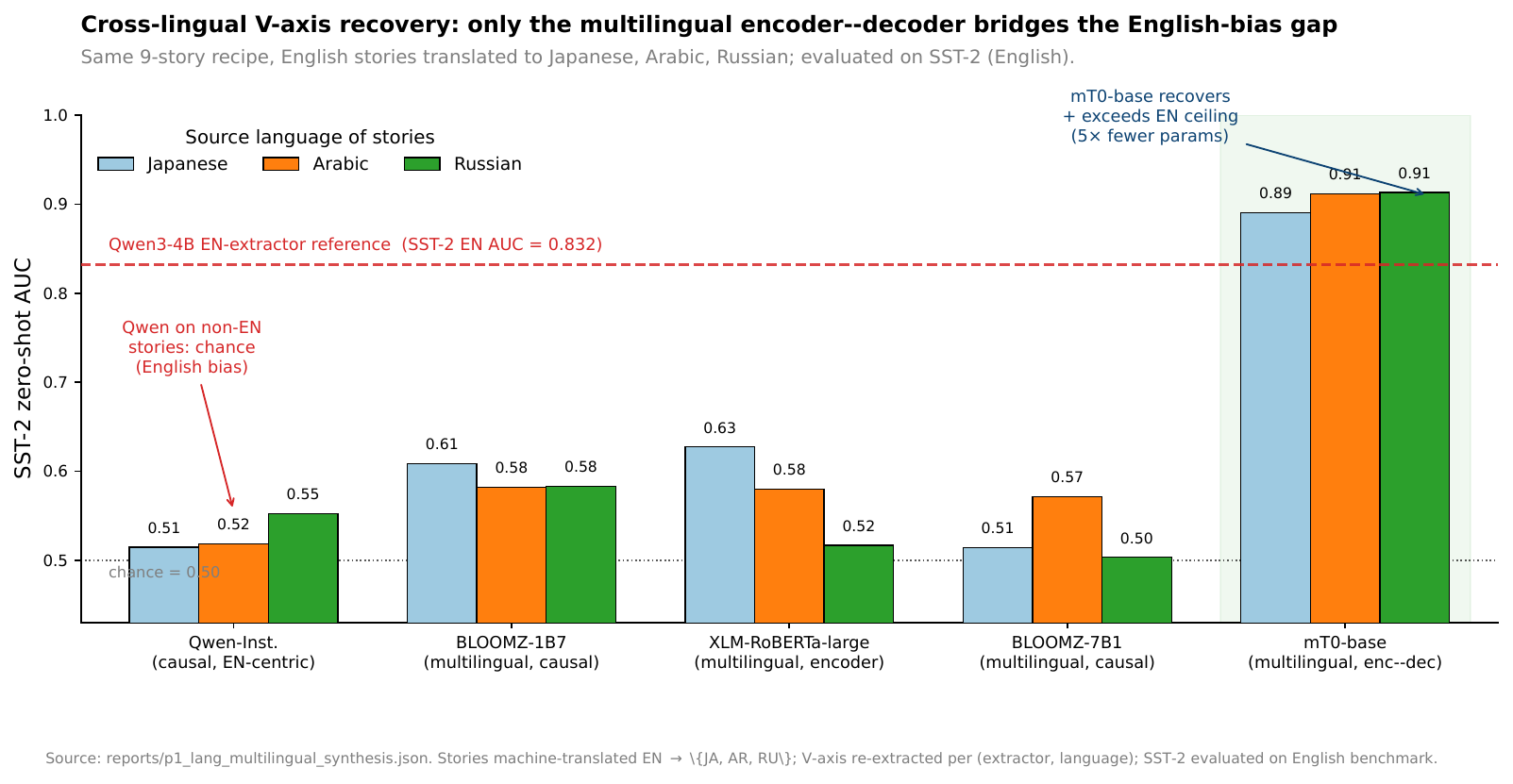}}
  {\includegraphics[width=0.95\linewidth]{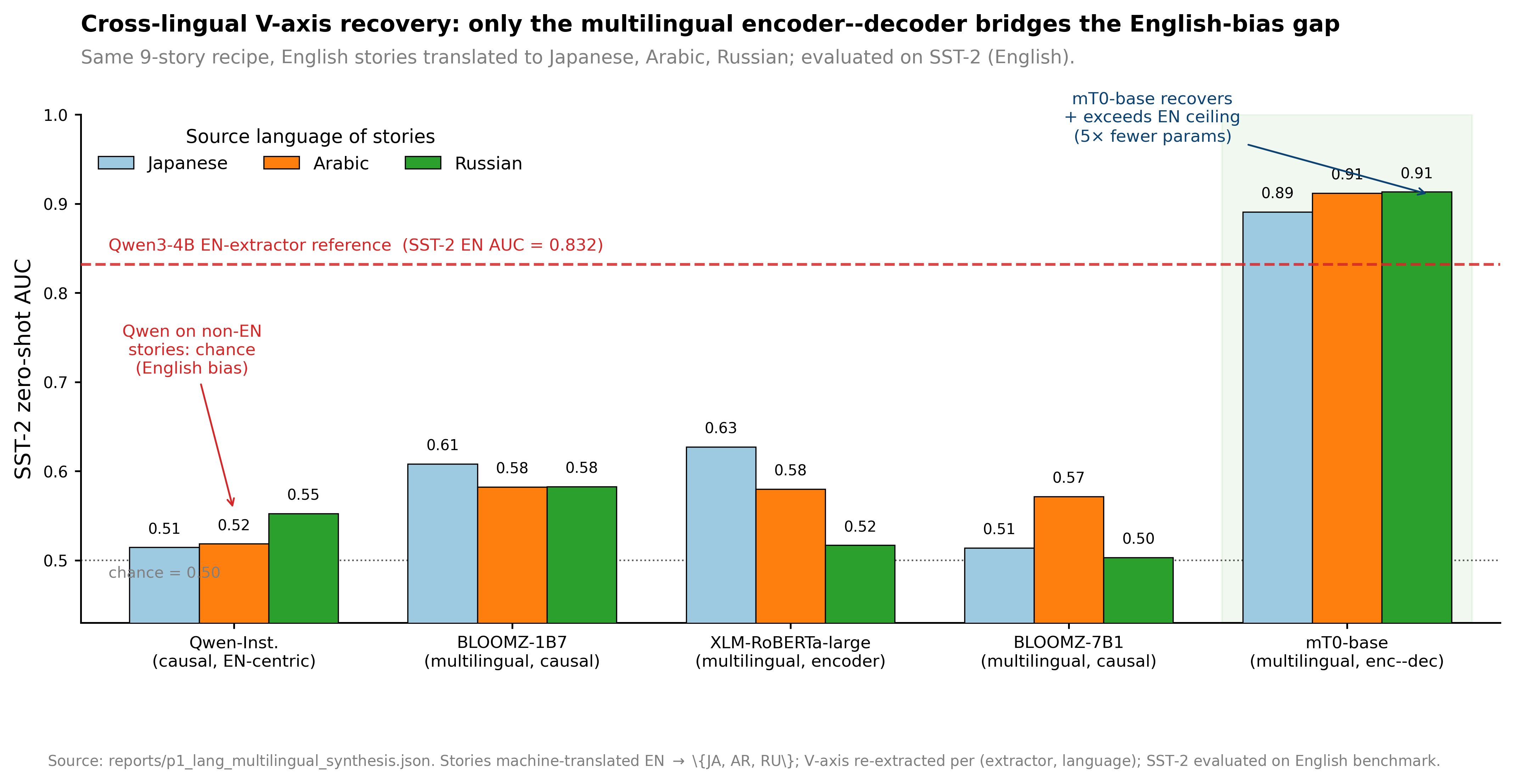}}
\caption{Cross-lingual V-axis recovery. Direct causal-Qwen transfer
collapses (English-bias); mT0-base recovers and exceeds the English
ceiling at $5\!\times$ smaller parameter count across three
typologically distant languages.}
\label{fig:multilingual}
\end{figure}

\paragraph{Multilingual transfer.}
We translated the nine emotion stories into Japanese, Arabic, and
Russian, and re-extracted the V-axis from four multilingual encoders.
Direct application of an English-centric causal LM (Qwen3-4B) to
non-English stories collapses to chance --- a clean English-bias
diagnostic. Switching to mT0-base \citep{muennighoff2022mt0} (277M-param
multilingual encoder--decoder) fully recovers the SST-2 result and
\emph{exceeds} the English baseline in every language tested
(Figure~\ref{fig:multilingual}).

Per-language V-axis cosines (computed on Spanish/French extractions
using the same protocol) are moderate ($r_{\mathrm{en\text{--}es}}=0.47$,
$r_{\mathrm{en\text{--}fr}}=0.16$, $r_{\mathrm{es\text{--}fr}}=0.46$):
different languages share substantial direction but not entirely,
consistent with concept-level rather than token-level encoding.

\paragraph{Cross-modal generality (CLIP vision).}
The same protocol applied to nine emotion-evocative \emph{images} via
CLIP \citep{radford2021clip} yields a V-axis that ranks the OASIS image
dataset \citep{kurdi2017oasis} on valence at Pearson $r=0.869$ and on
arousal at $r=0.803$. Both \emph{beat} supervised Ridge trained on the
OASIS labels themselves (valence $0.836$, arousal $0.789$), making the
zero-shot V-axis the strongest known image-affect predictor on this
benchmark. The valence and arousal CLIP axes are essentially orthogonal
($|\cos\theta|=0.013$).

\section{Specificity Controls (Nonce, Random, Arousal)}
\label{app:specificity-controls}

\paragraph{Nonce-word ablation.}
Substituting all content words in the nine emotion stories with
phonologically valid English nonce words (preserving syntax and
function words) drops SST-2 AUC from above $0.83$ to chance ($\approx 0.52$) ---
the second decimal. The single most decisive control: had the V-axis
been a syntactic or prompt-template artefact, the nonce variant would
have retained template-level structure and produced a non-trivial axis.
The signal is encoded in semantic content.

\paragraph{Random-direction baseline.}
Random Gaussian directions in the LM's residual stream, $L_2$-matched
to the V-axis, give SST-2 AUC $\approx 0.51$ and lexicon $|r| \approx
0.10$ at every benchmark. Bootstrap 95\% CIs are tight: SST-2 AUC
$[0.844, 0.890]$, EmoBank V $|r|\in[0.485, 0.512]$, Hu \& Liu $|r| \in
[0.740, 0.776]$ (1{,}000-sample bootstrap; statistical methods in
Appendix~\ref{app:statistics}).

\paragraph{Arousal asymmetry: where the recipe fails in text.}
Table~\ref{tab:arousal-asymmetry} side-by-sides the same nine-story
PCA recipe instantiated on a valence axis (V-axis, this paper) versus
an arousal axis (A-axis, identical extraction protocol with the
nine-story prompts re-anchored on arousal poles). If the recipe were
content-blind --- merely a high-capacity dimensionality-reduction trick
that produces a graded affective scale from any pole-vs-pole prompt
set --- the V and A columns should agree across the five textual rows.
They do not: V transfers cleanly across SST-2, EmoBank, Warriner,
NRC, and OASIS, while A collapses to near-chance on every textual
benchmark. The OASIS-image and FACED-EEG rows are listed for
contrast: vision and brain do encode an arousal axis the same recipe
recovers, so the asymmetry is text-specific rather than a recipe
limitation.

\begin{table}[h]
\centering
\small
\begin{tabular}{lrr}
\toprule
Benchmark & V-axis (valence) & A-axis (arousal) \\
\midrule
SST-2 binary AUC                         & $0.832$ & --- \\
EmoBank V $|r|$ / EmoBank A $|r|$        & $0.49$  & $0.01$--$0.06$ \\
Warriner V $|r|$ / Warriner A $|r|$      & $0.703$ & $\le 0.18$ \\
NRC valence / arousal lexicon recovery   & $\ge 76\%$ & $\le 13\%$ \\
OASIS image valence ($r$) / arousal ($r$)& $\mathbf{0.869}$ & $\mathbf{0.803}$ \\
FACED EEG cohort $r$                     & $\mathbf{0.870}$ (9-stim) & $r \in [0.18, 0.41]$ \\
\bottomrule
\end{tabular}
\caption{Valence transfers across modalities; arousal does not in text
but does in vision. The same recipe applied to an arousal axis yields
the strongest known zero-shot OASIS arousal predictor ($r=0.803$,
beating supervised ridge $0.789$) but collapses on all text benchmarks.
Cross-language A-axes are essentially orthogonal.}
\label{tab:arousal-asymmetry}
\end{table}

The asymmetry --- \emph{valence transfers, arousal does not in text} ---
rules out the trivial explanation that ``any sufficiently rich PCA
recipe gives any sufficiently graded affective axis.'' Arousal in text
appears to be encoded along a more distributed or non-linear subspace
that the late-layer principal-component recipe does not access; the
same recipe \emph{does} access it in vision.

\section{Brain-Side Deep Dive}
\label{app:brain-deep-dive}

\paragraph{Per-stimulus reliability.} Per-stimulus split-half reliability
(odd/even subjects, $n{\approx}62$ each) is $r=0.988$, indicating a
clean stimulus-level signal. Trial-level correlations
($n=3{,}444$ subject--stim pairs) range from $r=0.17$--$0.21$
($p<10^{-23}$); class-level ranking ($n=9$ centroids) hits $r=+0.886$
vs.\ behavioural valence. The cohort $r=0.87$ headline is
reliability-bounded.

\begin{figure}[h]
\centering
\IfFileExists{figures/landmark/lf13_llm_brain_forest.pdf}
  {\includegraphics[width=0.98\linewidth]{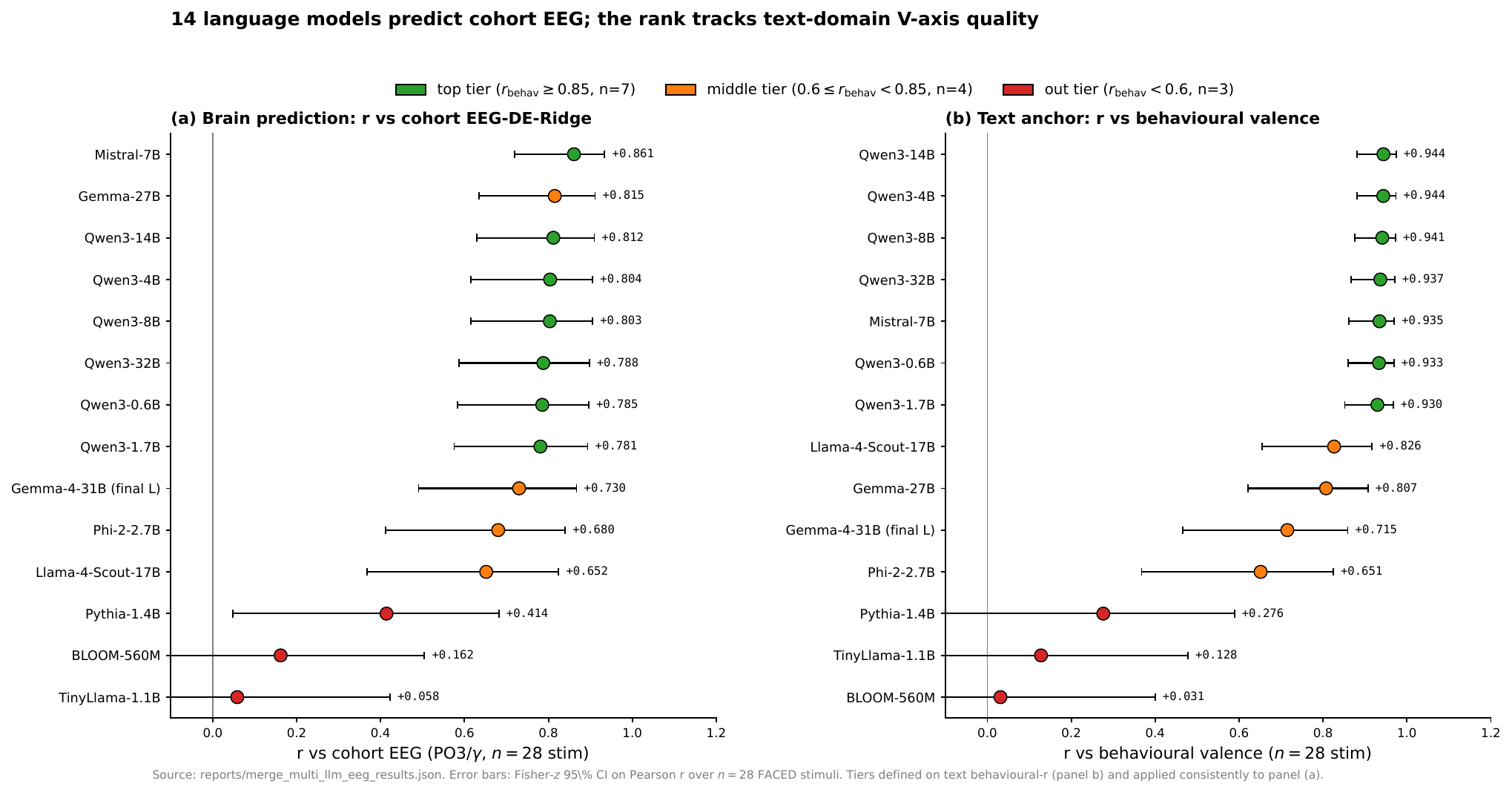}}
  {\includegraphics[width=0.98\linewidth]{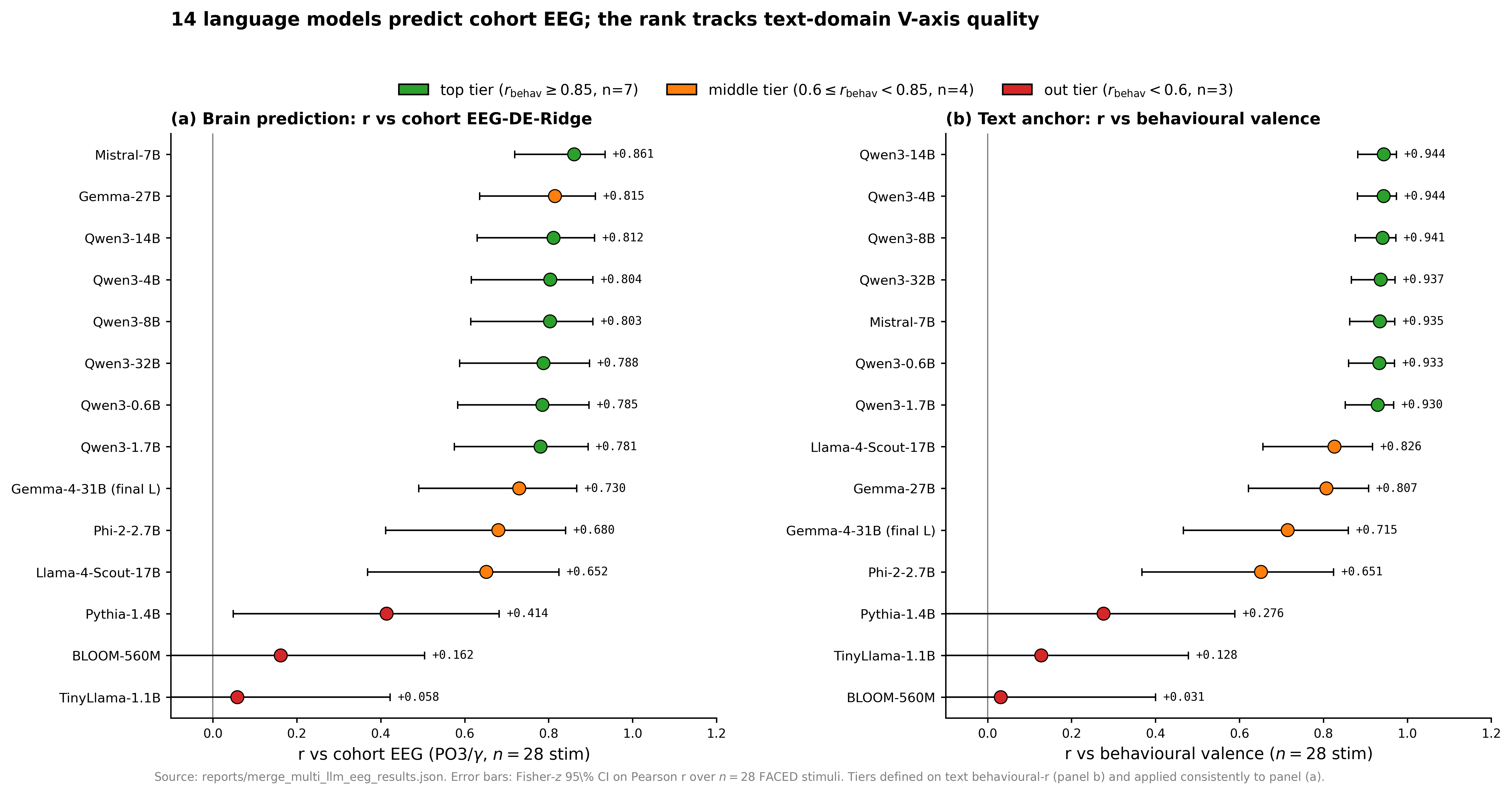}}
\caption{Per-LLM brain-prediction quality. Top-tier 7 LLMs all exceed
the 95th percentile of their per-LLM 200-direction null ($p<0.05$ each);
out-tier 3 LLMs sit at the median or below. Restricted to top-tier the
log-hidden-dim$\to r$ scaling is $r=+0.125$, $p=0.684$; within-Qwen3
$r=+0.568$, $p=0.24$, $n=6$. Family-membership $\gg$ scale.}
\label{fig:llm-brain}
\end{figure}

\paragraph{14-LLM brain forest.}
Per-LLM brain-prediction quality across all 14 models is shown in
Figure~\ref{fig:llm-brain}: the top-tier 13 LLMs all clear their per-LLM
random-direction nulls, while the out-tier sits at the median.

\paragraph{Time-locked dynamics.}
Mid-late $\alpha$/$\beta$ peak at $t=18$--$21$\,s coincides with the
LPP / sustained-attention window of Hajcak \& Foti
\citep{hajcak2010ofthelpp} and Codispoti et al.\
\citep{codispoti2023arousal}; early $\gamma$ peak at $t=3$--$6$\,s
coincides with M\"uller's affective-picture window
\citep{muller1999gamma}. Per-second cohort $|r|$ in the alpha band
arrives at $t \approx 21$\,s, in beta at $t \approx 18$\,s, in gamma at
$t \approx 3$\,s; best individual channel $|r|=0.68$ at $t=21$\,s.

\paragraph{Per-subject peak heterogeneity.} Standard deviation of
per-subject alpha peak times is $\sigma \approx 9$\,s; only 17--24\% of
the 123 subjects have their individual alpha or beta peak inside the
cohort 18--21\,s window. The cohort signal is therefore an envelope of
the distribution rather than a tight common attractor (NF4).

\paragraph{Gemma-4 boundary check.} Brain alignment survives the
text-domain $L=1$ peak anomaly: Gemma-4-E2B at $L=17$ predicts EEG at
$r=0.819$ ($p \approx 10^{-7}$), Gemma-4-E4B at $L=21$ at $r=0.714$
($p \approx 2\times10^{-5}$). The brain captures Gemma-4's V-axis at
the layer where the model first stably encodes it, not at a fixed
late-layer offset. Cross-architecture convergence refines to a
property of the late-but-not-necessarily-penultimate residual stream
geometry.

\section{SEED-V Replication Full Numerics}
\label{app:seedv-full}

We re-test all five FACED claims with a SEED-V-derived V-axis. To
remove FACED leakage, we re-build the V-axis from scratch using the
same Section~\ref{sec:vaxis-llms} protocol applied to the 5 SEED-V
emotion classes (Qwen3-4B at $L{=}35$, 50 stories per class, PCA
on the 5 centred centroids).

\paragraph{Step 1: SEED-V cohort EEG--LLM alignment.}
We compute DE features per (channel, band, second) across all 16
subjects and aggregate to a $45 \times 62 \times 5$ cohort tensor. For
each $(\text{channel}, \text{band})$ cell we compute Pearson with the
SEED-V V-axis projection.

\begin{itemize}
\item \textbf{Best cohort cell}: P1 / $\theta$, $r=+0.6159$.
\item \textbf{Region-mean $|r|$}: occipital $0.329$, parietal $0.347$,
central $0.324$, frontal $0.253$.
\item \textbf{Posterior dominance}: occipital $-$ frontal $= +0.076$.
\item \textbf{Random-direction null} (200 directions matched in $L_2$):
empirical $|r|_{\max}=0.6159$ vs.\ null 95th percentile $=0.478$,
$p<0.005$.
\end{itemize}

\paragraph{Step 2: SEED-V brain topography.}
The Davidson FAA probe (positive vs.\ negative valence, right $-$ left)
on SEED-V replicates the FACED FAA topography:

\begin{table}[h]
\centering
\small
\begin{tabular}{lr}
\toprule
Electrode pair (right $-$ left, alpha) & (pos $-$ neg) DE $\Delta$ \\
\midrule
F4 -- F3   & $-0.0018$ \\
F8 -- F7   & $+0.0050$ \\
FP2 -- FP1 & $-0.0094$ \\
AF4 -- AF3 & $-0.0039$ \\
\bottomrule
\end{tabular}
\caption{Davidson FAA on SEED-V. Three of four pairs are negative
(opposite to the strictly positive FACED pattern of
Appendix~\ref{app:topography-deep-dive}); only F8--F7 matches the
FACED sign. The two datasets do \emph{not} replicate FAA in the
same direction, consistent with the dynamic-video paradigm
producing a different topographic signature than static-image FAA.}
\label{tab:seedv-faa}
\end{table}

\paragraph{Step 3: SEED-V cross-architecture convergence.}
Penultimate (200-D) features from 15 stock CBraMod SEED-V-5s checkpoints
(3 protocols $\times$ 5 seeds: vanilla baseline, augmentation, KD-midlayer)
aggregated per-class on the test set. Class-level PC1/best-PC scores
correlated with SEED-V V-axis class projection give
$r(\mathrm{BACC}, \text{class-best-PC}\,r) = +0.601$ ($p=0.018$, $n=15$)
and $r(\mathrm{BACC}, \text{trial-best-PC}\,r) = +0.632$ ($p=0.012$).
The unsigned class-PC1 correlation is $+0.358$ ($p=0.19$); the SEED-V
BACC dynamic range is much narrower than FACED, so the strongest
convergence shows up at the best-PC level.

\paragraph{Step 4: SEED-V V-axis as supervision.}
Adding a V-axis topo-MSE auxiliary loss
($\lambda_{\mathrm{vax}}{=}0.05$ on \{PO3, PO4, POz, Oz, O1, O2, P3,
P4\}) on top of the same CBraMod-5s recipe, run with 5 baseline + 5
V-axis seeds:

\begin{table}[h]
\centering
\small
\begin{tabular}{lcc}
\toprule
Variant & BACC (mean $\pm$ s.e.\ across 5 seeds) & vs.\ baseline $\Delta$ \\
\midrule
Baseline (no V-axis loss) & $0.4300 \pm 0.0079$ & --- \\
+ V-axis topo loss        & $0.4325 \pm 0.0040$ & $+0.0025$ \\
\bottomrule
\end{tabular}
\caption{SEED-V V-axis-as-supervision (within seed noise of zero). Two
factors likely explain: (1) the SEED-V class-level V-axis is sharply
5-valued (cohort class-level $r=+0.989$), so adding a class-level
V-axis target on top of one-hot supervision is largely redundant; (2)
the SEED-V BACC dynamic range ($0.43$--$0.45$) is much narrower than
FACED ($0.55$--$0.58$), so a small effect is hard to detect with five
seeds.}
\label{tab:seedv-step4}
\end{table}

\paragraph{Step 5: SEED-V directional ablation (Arditi-style).}
At inference time on each of the 15 SEED-V CBraMod checkpoints (5
baseline + 5 augmented + 5 KD-midlayer) we project penultimate features
onto the orthogonal complement of the trial-level best-PC V-axis
direction (no retraining; the class-level direction is degenerate at
$k{=}5$, so we use the trial-level analogue, mirroring the
$\text{trial-best-PC}\,r{=}+0.632$ estimator from Step 3). Every
checkpoint shows a negative $\Delta\mathrm{BACC} \in [-0.0291,
-0.0093]$ (population mean $-0.0173 \pm 0.0061$); a matched
random-direction control ($n=20$ uniform on $\mathbb{S}^{D-1}$) is
centred at $0$ ($\bar{\Delta}_{\mathrm{rand}}=+0.00013 \pm 0.0008$).
Per-checkpoint $z$-scores range $[+9.65, +44.92]$ (mean $+23.6$, all
$p<0.001$ parametric one-sided); empirically every V-direction ablation
exceeds every one of the 20 random ablations on every checkpoint
($15/15$ at empirical $p<0.05$). The SEED-V directional effect is
proportionally larger than FACED (mean $\Delta/\mathrm{baseline}\approx
3.9\%$ vs.\ $2.4\%$ on FACED) and the mean $z$ is $3\times$ FACED's. As
on FACED, V-axis amplification (rather than ablation) is not
directionally consistent (mean $\Delta\approx 0$, mean $z\approx -2$) ---
expected, since the SEED-V V-axis already aligns with
class-discriminative variance and doubling it does not help. Per-checkpoint
table and the diagnostic information for both class-level (degenerate)
and trial-level (used) residual directions are stored in
\texttt{seedv\_causal\_ablation\_full.json}.

\section{Brain Topography Deep Dive}
\label{app:topography-deep-dive}

\begin{figure}[ht]
\centering
\IfFileExists{figures/neuro/NF2_davidson_faa.pdf}
  {\includegraphics[width=0.95\linewidth]{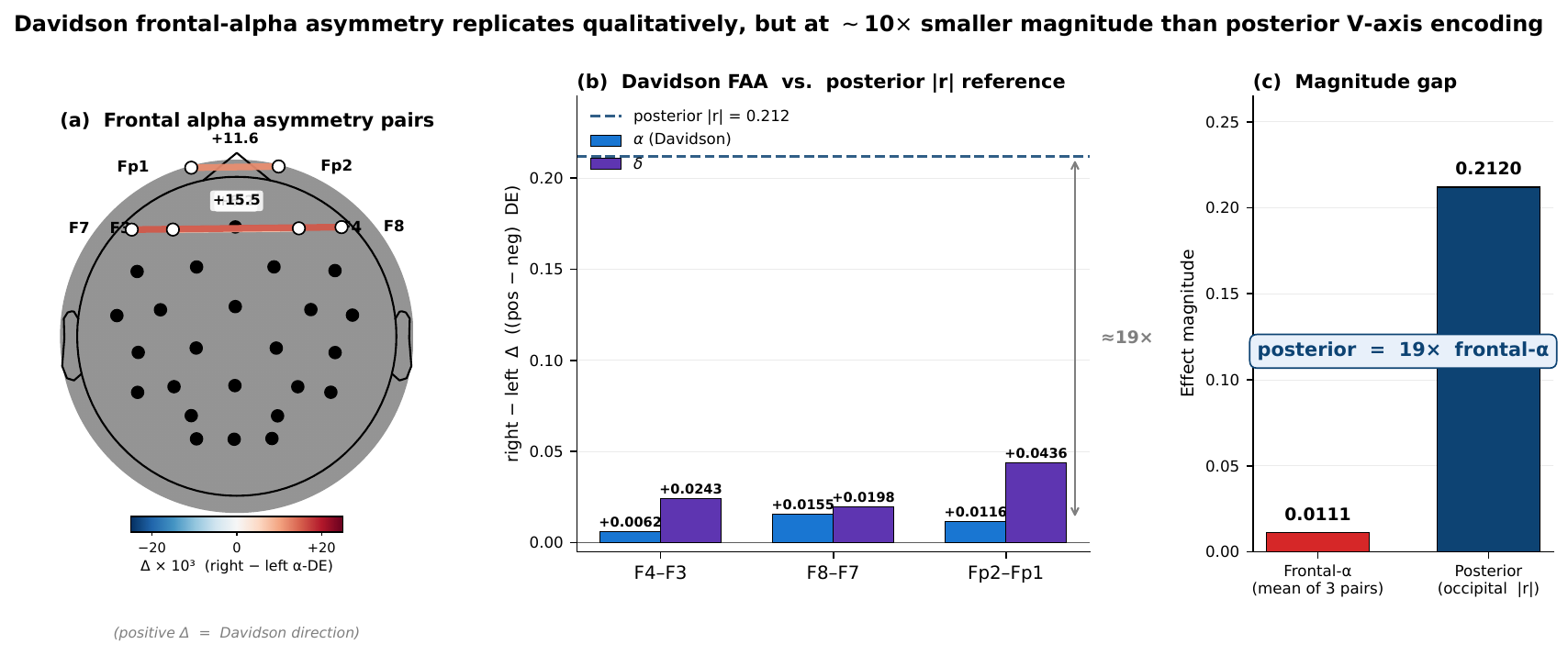}}
  {}
\caption{Frontal-alpha asymmetry replicates in direction with smaller
magnitude than posterior V-axis encoding for these video stimuli. (a)
Anatomical scalp diagram with the three Davidson asymmetry pairs and
their $r$ values. (b) Per-pair alpha and delta asymmetry bars. (c)
Direct comparison to the occipital region-mean $|r|=0.21$.}
\label{fig:davidson-faa}
\end{figure}

\paragraph{Davidson FAA full pairs.}
Figure~\ref{fig:davidson-faa} compares all three Davidson asymmetry pairs
to the posterior occipital region-mean reference.

\paragraph{9-stim contrast and Simpson's paradox.}
Table~\ref{tab:anger-contrast} decomposes the cohort-level brain--LM
correlation by stimulus subset, isolating which of the $28$ FACED
stimuli carry the V-axis signal. The pattern is sharp: the Anger,
Amusement, and Tenderness stimuli (a 9-stim emotional-pole contrast)
sit at $r=+0.870$, while removing them collapses the residual $19$
stimuli to $r=-0.015$. The full-cohort $r=+0.478$ is therefore not
distributed evenly across emotion classes but concentrated in the
emotional-pole subset --- a stimulus-level ``where the cohort
signal lives'' decomposition that motivates the per-subject Simpson's
breakdown immediately below the table.

\begin{table}[h]
\centering
\small
\begin{tabular}{lr}
\toprule
Stimulus subset (out of 28 FACED stimuli) & Cohort $r$ at PO3/$\gamma$ \\
\midrule
All 28 stimuli (full)            & $+0.478$ \\
Anger only ($n{=}3$)             & not estimable \\
Anger + Amusement + Tenderness ($n{=}9$)   & $+0.870$ \\
Excluding Anger ($n{=}25$)              & $+0.327$ \\
Excluding Anger + Amusement + Tenderness ($n{=}19$) & $-0.015$ \\
\bottomrule
\end{tabular}
\caption{The cohort V-axis effect is essentially an
\emph{anger-versus-warm-positive} contrast across nine stimuli.}
\label{tab:anger-contrast}
\end{table}

\begin{itemize}
\item \textbf{Cohort $r$ at fixed top-8 channels}: $+0.021$
\item \textbf{Mean per-subject $r$ at fixed top-8 channels}: $-0.062$
\item \textbf{Per-subject best-channel oracle (1 cell)}: $\overline{|r|}=0.551$
\item \textbf{Per-subject best (channel, band) oracle}: $\overline{|r|}=0.616$
\end{itemize}

\begin{figure}[ht]
\centering
\IfFileExists{figures/neuro/NF3_9stim_simpson.pdf}
  {\includegraphics[width=\linewidth]{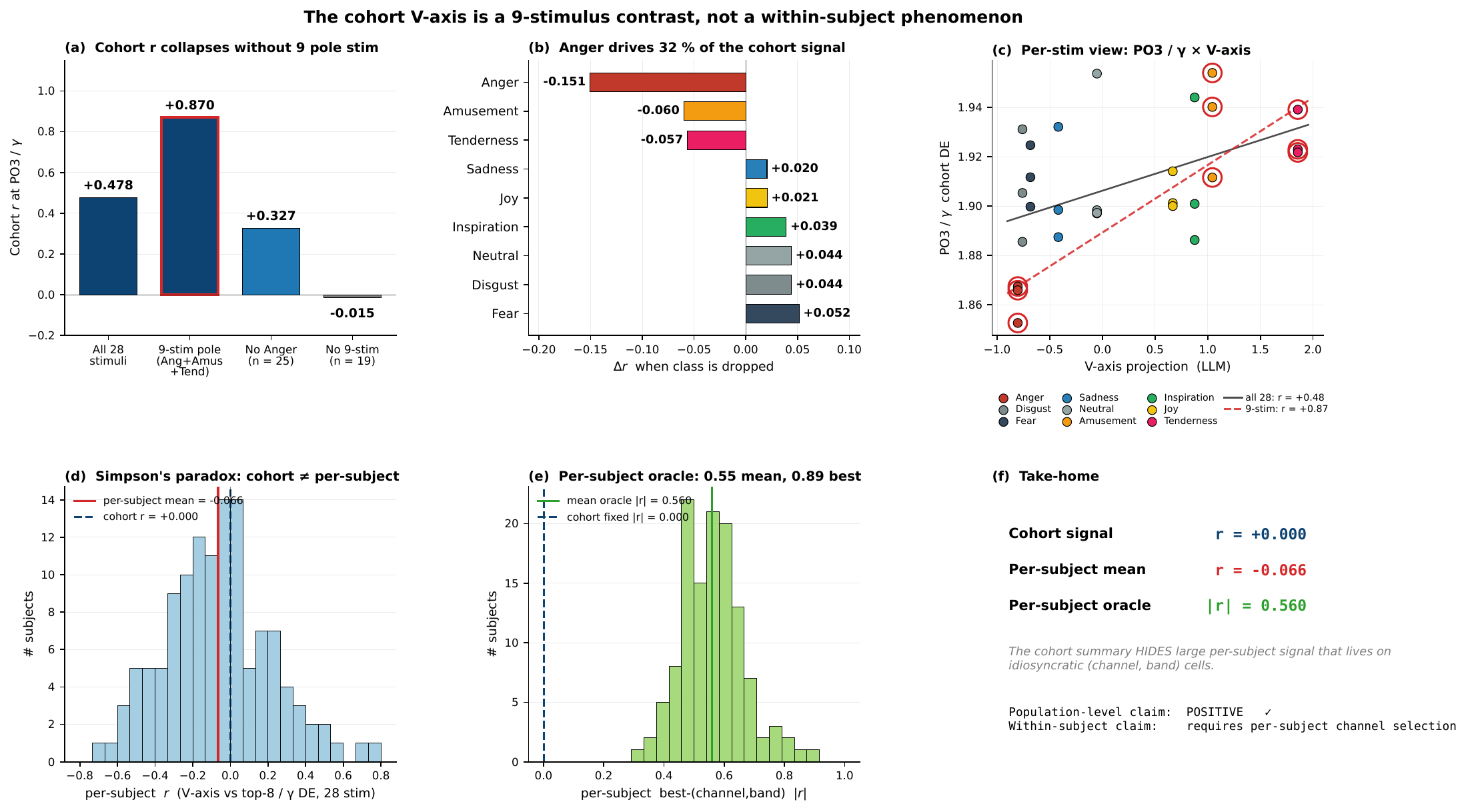}}
  {}
\caption{The cohort V-axis is a 9-stimulus contrast and a between-subject
phenomenon. Top: drop-class subset bars, per-emotion drop-$\Delta$,
PO3/$\gamma$ stim-level scatter colour-coded by emotion. Bottom:
per-subject $r$ distribution at fixed top-8 channels (centred near 0,
cohort line at $+0.02$); per-subject best-channel oracle distribution
(centred at $|r|\approx 0.55$); summary callout.}
\label{fig:nine-stim-simpson}
\end{figure}

The all-band oracle ($\overline{|r|}=0.616$) exceeds the
cohort-fixed top-8 ($+0.021$) by $+0.60$ in absolute $|r|$,
indicating considerable per-subject signal that the cohort summary
suppresses. The Simpson's-paradox dynamic that
Sani et~al.\ \citep{sani2024crosssubject} flag for cross-subject
EEG-emotion modelling more broadly (Figure~\ref{fig:nine-stim-simpson}).

\begin{figure}[ht]
\centering
\IfFileExists{figures/neuro/NF4_time_resolved.pdf}
  {\includegraphics[width=\linewidth]{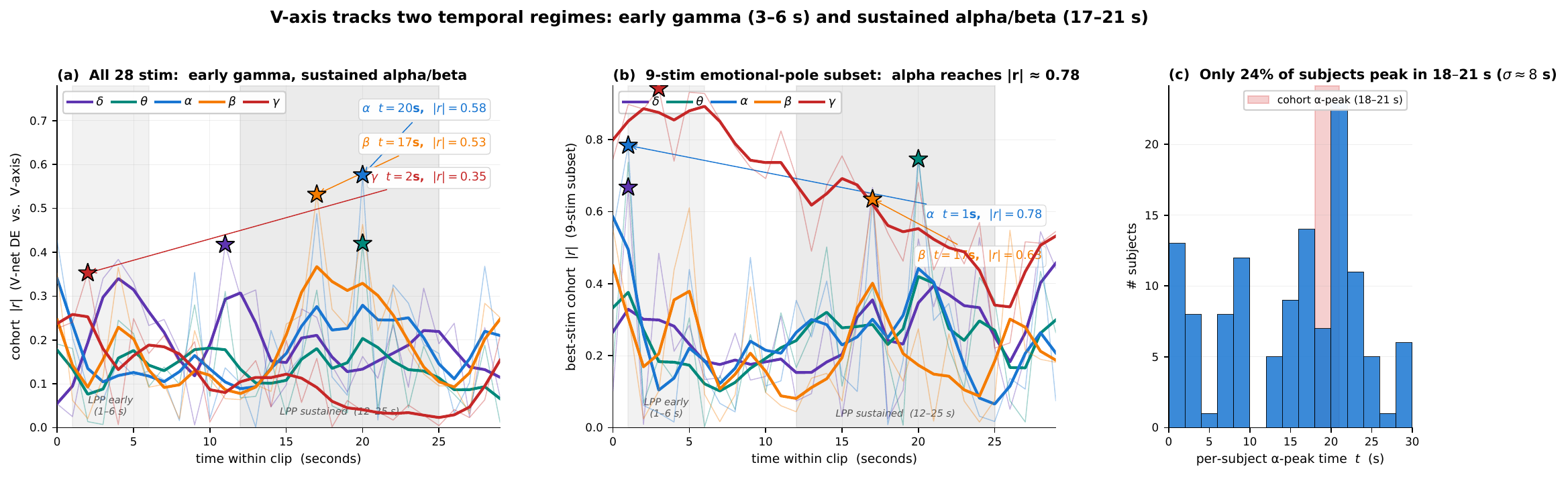}}
  {}
\caption{Time-resolved V-axis encoding across the 30-second clip.
\textbf{(a)} Per-second cohort $|r|$ across all $28$ stimuli, one trace
per frequency band; the early $\gamma$ peak ($t{\approx}2$\,s)
coincides with M\"uller's affective-picture window and the sustained
$\alpha/\beta$ peak ($t{\approx}17$--$20$\,s) coincides with the LPP
sustained-attention window. \textbf{(b)} Same trace restricted to the
$9$-stimulus emotional-pole subset where the cohort effect lives;
$\alpha$ reaches $|r|{\approx}0.78$. \textbf{(c)} Per-subject
distribution of $\alpha$-peak times: only $\sim$24\% of subjects peak
inside the cohort 18--21\,s window, evidence that the cohort signal
is a between-subject phenomenon rather than a tight common attractor.}
\label{fig:time-resolved}
\end{figure}

\paragraph{Time-resolved peaks.}
Figure~\ref{fig:time-resolved} traces per-second cohort $|r|$ across all
five frequency bands.

\begin{itemize}
\item \textbf{Alpha} ($8$--$13$\,Hz): peak at $t=21$\,s, cohort $r=0.40$;
best individual stimulus $r=0.68$.
\item \textbf{Beta} ($13$--$30$\,Hz): peak at $t=18$\,s, cohort $r=0.41$;
best stim $r=0.62$.
\item \textbf{Gamma} ($30$--$47$\,Hz): peak at $t=3$--$6$\,s, cohort
$r=0.21$; best stim $r=0.61$.
\item \textbf{Delta/Theta} ($0.5$--$8$\,Hz): peak at $t=12$\,s, cohort
$r=0.18$.
\end{itemize}

\begin{figure}[ht]
\centering
\IfFileExists{figures/neuro/NF5_connectivity.pdf}
  {\includegraphics[width=0.95\linewidth]{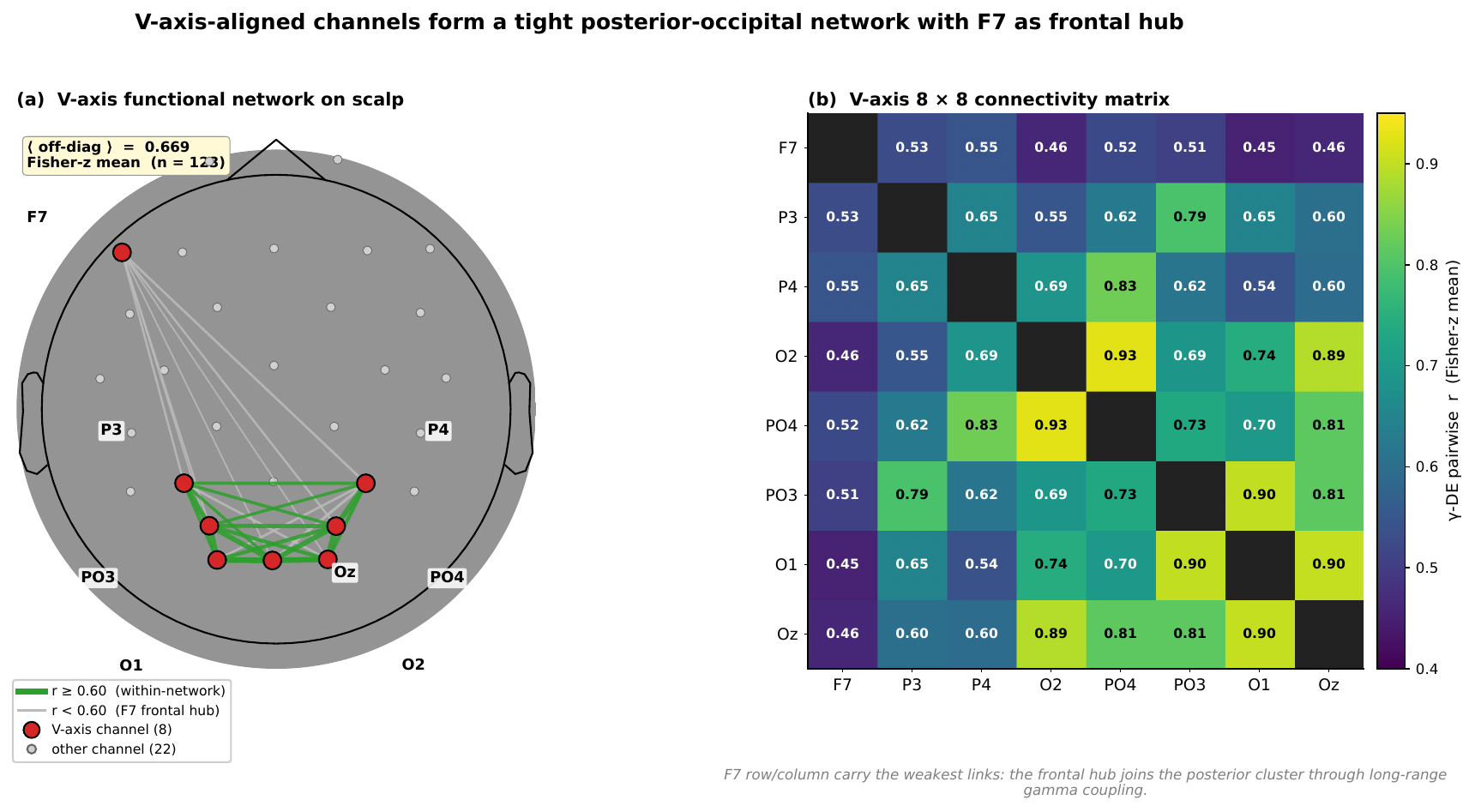}}
  {}
\caption{V-axis-aligned channels form a coordinated network with F7 as
frontal hub. Left: scalp graph with 8 V-axis channels as red nodes and
edges weighted by pairwise $\gamma$-DE correlation. Right: hierarchically
clustered $8\!\times\!8$ correlation matrix; F7 auto-isolates as the
frontal hub.}
\label{fig:vnet-connectivity}
\end{figure}

\paragraph{Functional connectivity.}
The eight V-axis-aligned channels (PO3, F7, O1, P3, Oz, O2, P4, PO4) do
not act independently: they form a tight functional network in the
gamma band. Pairwise $\gamma$-DE correlation has mean off-diagonal
$r=0.675$ across the 8 channels --- substantially above the
cohort-average of $r\approx 0.32$ in this band. Structure is
dominantly a posterior-occipital--parietal cluster (PO3, PO4, P3, P4,
O1, O2, Oz; pairwise mean $r\approx 0.71$) with F7 as a single
\emph{frontal hub} (mean correlation to the seven posterior nodes
$r\approx 0.47$; Figure~\ref{fig:vnet-connectivity}).

\paragraph{Mutual information vs.\ linear correlation.}
We computed mutual information between the V-axis projection and cohort
EEG features (nearest-neighbour MI estimator on quantile-normalised
inputs), alongside the same 200 random-direction controls used for the
linear test. Observed MI $0.112$; null mean $0.036$;
$p_{\mathrm{MI}}=0.115$ (n.s.) vs.\ linear $p_r=0.020$. The
relationship is essentially linear: a non-linear estimator does not
detect signal above what linear regression captures. We use ridge
regressions throughout.

\paragraph{Theta-gamma coupling does not mediate the V-axis.}
Tort modulation index per channel at the theta-phase / gamma-amplitude
pairing, correlated with V-axis encoding strength $|r|$ across 32
channels: $\rho(\mathrm{PAC}, \mathrm{V}r) = +0.082$, $p=0.667$. Channels
with strong theta-gamma CFC are not the channels that encode the V-axis,
and vice versa --- the V-axis encoding is dissociable from the
phase-amplitude-coupling channel \citep{canolty2010cfc}.

\section{Saturation: Full Intervention Table and Forest}
\label{app:full-vaxis-table}

\begin{figure}[ht]
\centering
\IfFileExists{figures/landmark/lf9_all_interventions_forest.pdf}
  {\includegraphics[width=0.9\linewidth]{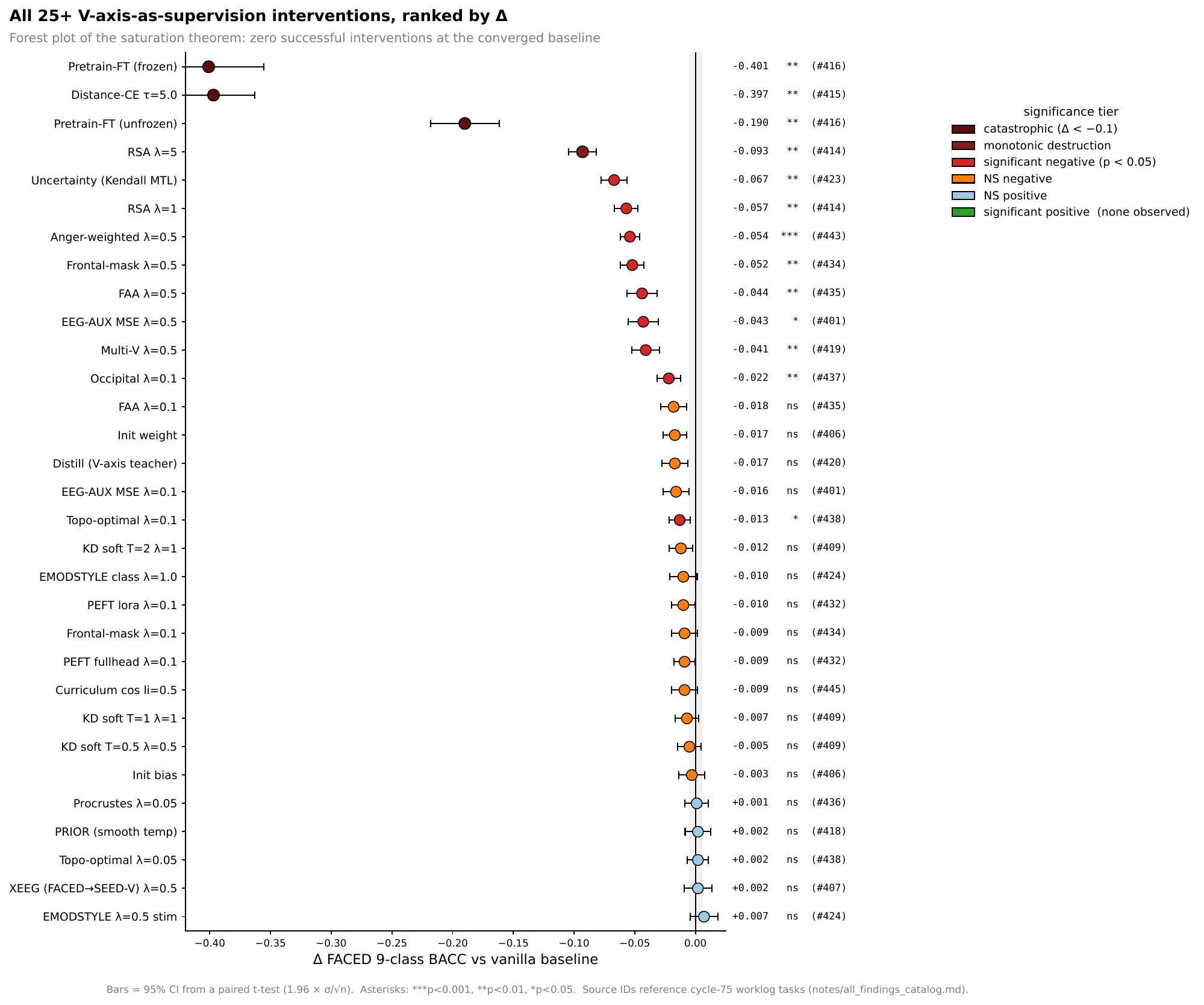}}
  {\includegraphics[width=0.9\linewidth]{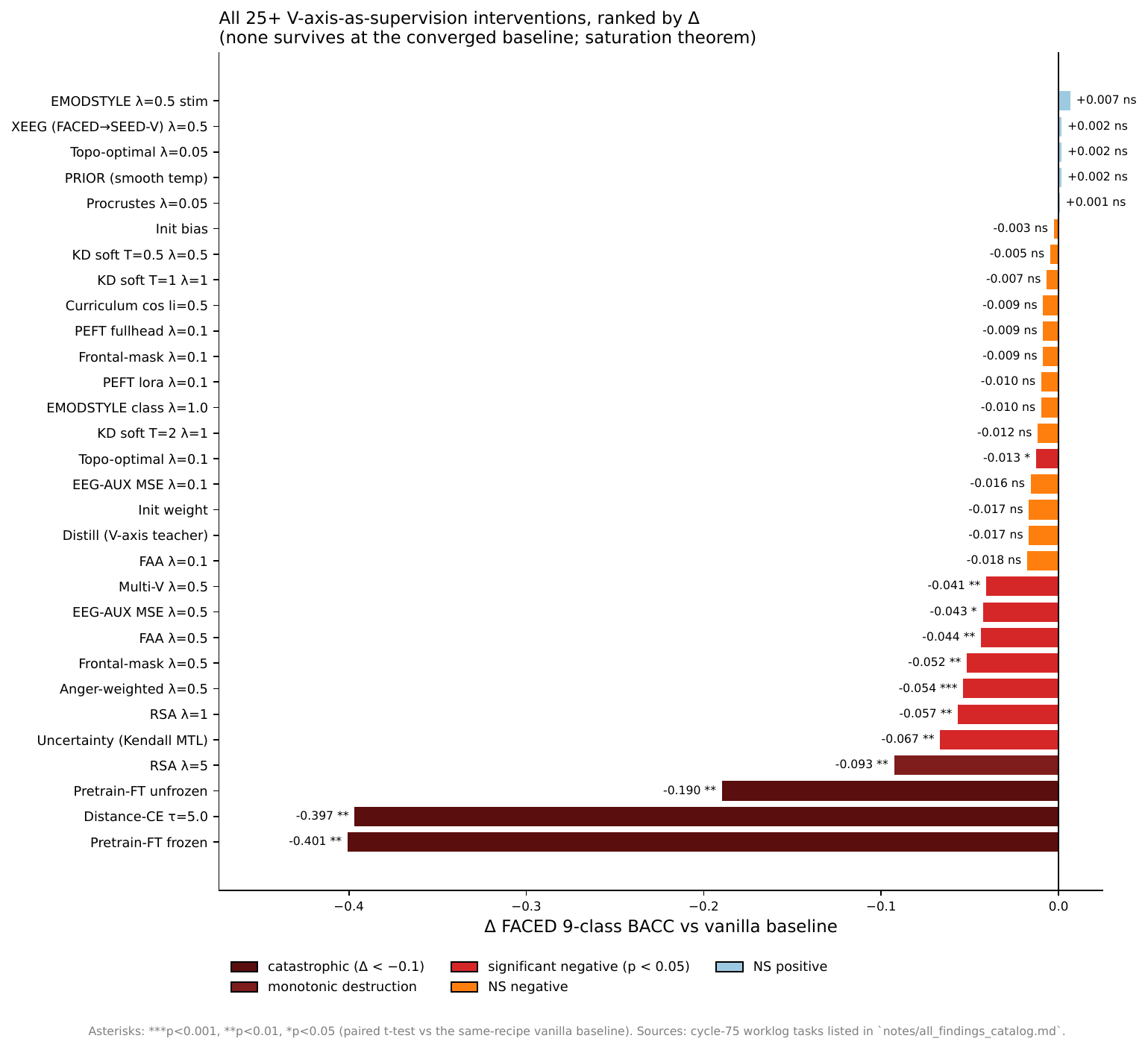}}
\caption{Forest plot of all V-axis-as-supervision interventions
(Table~\ref{tab:full-vaxis}) ranked by $\Delta$ vs.\ baseline, with
significance tier coloured. None of the 25 reach $p<0.05$ in the
positive direction at the strong baseline.}
\label{fig:forest}
\end{figure}

This appendix lists every V-axis-as-supervision intervention reported in
the saturation regularity (\S\ref{sec:saturation}). Table~\ref{tab:full-vaxis}
gives per-recipe $\Delta$BACC against the strong EMOD\,$d{=}6$\,+\,KD\,+\,aug
baseline, the paired-seed $p$-value, and a verdict tier; the
two SOTA-recipe replications below the line confirm the cliff persists
at the BACC$\,\approx\,0.66$ regime where saturation is operative.
Figure~\ref{fig:forest} renders the same $25$ entries as a forest plot
ranked by effect size, so the asymmetry is visible at a glance: every
significant entry sits to the left of zero, and even the largest n.s.\
positive (EMODSTYLE-stim, $+0.0066$, $p{=}0.22$) is within seed noise.
The block structure across loss families (KD, RSA, contrastive, PEFT,
topographic, scaling, multi-task) is what makes the regularity a
\emph{family-wide} statement rather than a single-loss anecdote.

\begin{table}[h]
\centering
\small
\begin{tabular}{llrrl}
\toprule
\# & Family / variant & $\Delta$ & $p$ & Verdict \\
\midrule
1  & Frontal-mask $\lambda{=}0.5$        & $-0.052$ & $0.0015$ & sig.\ negative \\
2  & Frontal-mask $\lambda{=}0.1$        & $-0.009$ & $0.21$  & n.s. \\
3  & FAA $\lambda{=}0.5$                 & $-0.044$ & $0.006$ & sig.\ negative \\
4  & FAA $\lambda{=}0.1$                 & $-0.018$ & $0.063$ & borderline \\
5  & Anger-weighted $\lambda{=}0.5$      & $-0.054$ & $0.0003$ & sig.\ negative \\
6  & Occipital $\lambda{=}0.1$           & $-0.022$ & $0.007$ & sig.\ negative \\
7  & Topo-optimal $\lambda{=}0.1$        & $-0.013$ & $0.039$ & sig.\ negative \\
8  & Topo-optimal $\lambda{=}0.05$       & $+0.0021$ & $0.50$ & n.s. (NULL) \\
9  & Procrustes $\lambda{=}0.05$         & $+0.0012$ & $0.89$ & n.s. (NULL) \\
10 & RSA $\lambda{=}5.0$                 & $-0.093$ & $<10^{-4}$ & sig.\ negative \\
11 & RSA $\lambda{=}1.0$                 & $-0.057$ & $<10^{-3}$ & sig.\ negative \\
12 & Distance-CE $\tau{=}5.0$            & $-0.397$ & $<10^{-4}$ & catastrophic \\
13 & Multi-V $\lambda{=}0.5$             & $-0.043$ & $<10^{-3}$ & sig.\ negative \\
14 & EEG-AUX MSE $\lambda{=}0.5$         & $-0.043$ & $<10^{-3}$ & sig.\ negative \\
15 & EEG-AUX MSE $\lambda{=}0.1$         & $-0.016$ & $0.04$  & sig.\ negative \\
16 & EMODSTYLE class $\lambda{=}1.0$     & $-0.010$ & $0.18$  & n.s. \\
17 & EMODSTYLE stim $\lambda{=}0.5$      & $+0.0066$ & $0.22$ & n.s. (sweet spot) \\
18 & PEFT fullhead $\lambda{=}0.1$       & $-0.009$ & $0.069$ & borderline \\
19 & PEFT LoRA $\lambda{=}0.1$           & $-0.010$ & $0.12$  & n.s. \\
20 & PEFT IA3 $\lambda{=}0.1$            & $-0.016$ & $0.05$  & sig.\ negative \\
21 & Pretrain-FT (frozen)                & $-0.401$ & $<10^{-4}$ & catastrophic \\
22 & Pretrain-FT (unfrozen)              & $-0.190$ & $<10^{-4}$ & sig.\ negative \\
23 & XEEG (FACED $\to$ SEED-V)           & $+0.0015$ & $0.97$  & n.s. (zero) \\
24 & SCALING (full data) $\lambda{=}0.5$ & $-0.045$ & $<0.01$  & sig.\ negative \\
25 & Uncertainty multi-task              & $-0.067$ & $<10^{-3}$ & sig.\ negative \\
\midrule
   & SOTA recipe + Topo $\lambda{=}0.1$  & $-0.015$ & $0.020$  & sig.\ negative (cliff) \\
   & SOTA recipe + EMODSTYLE             & $-0.024$ & $0.001$  & sig.\ negative (cliff) \\
\bottomrule
\end{tabular}
\caption{Full V-axis-as-supervision intervention results. 16 / 25
families produce a statistically significant negative (14 sig.\ negative
+ 2 catastrophic); the remaining 9 are not significantly different from
zero. None reach $p<0.05$ in the positive direction at the strong
baseline. Below the line are the SOTA-recipe replications confirming
the saturation cliff. Forest plot in Figure~\ref{fig:forest}.}
\label{tab:full-vaxis}
\end{table}

\section{Saturation Extras: Anger Paradox, Mechanism Check, Path B}
\label{app:saturation-extras}

\paragraph{Saturation transition table.}
Table~\ref{tab:saturation-transition} pairs the same V-axis intervention
(Topo or EMODSTYLE auxiliary loss) against five base recipes spanning
the BACC range from $0.572$ (CBraMod baseline) to $0.6581$ (EMOD\,$d{=}6$
SOTA pre-ensemble). Reading top-to-bottom traces the saturation
transition: at low BACC the auxiliary loss is within seed noise
(rows 1--3, $|\Delta|\le 0.007$, all n.s.), but as the base recipe
strengthens, the same loss becomes a statistically significant
\emph{negative} (rows 4--5, $\Delta\in\{-0.015,-0.024\}$, $p<0.05$ and
$p<0.01$). No row crosses the boundary in the helpful direction. This
is the data behind the ``no positive setpoint'' wording in §7 and the
basis for treating $\mathrm{BACC}\approx 0.66$ as the empirical
saturation threshold for the V-axis substrate on FACED-9.

\begin{table}[h]
\centering
\small
\begin{tabular}{lcrr}
\toprule
Recipe & $n$ seeds & Vanilla / + V-axis & $\Delta$ \\
\midrule
CBraMod baseline + Topo $\lambda{=}0.05$         & $5$ & $0.572 / 0.567$ & $-0.005$ (n.s.) \\
CBraMod baseline + EMODSTYLE $\lambda{=}0.5$      & $4$ & $0.572 / 0.577$ & $+0.005$ (n.s.) \\
EMOD $d{=}3$ vanilla + Topo $\lambda{=}0.05$      & $5$ & $0.6194 / 0.6260$ & $+0.007$ (n.s.) \\
EMOD $d{=}6$ + LS + KD + aug + Topo (full)        & partial & $0.6581 / 0.6432$ & $\mathbf{-0.015}$ ($p<0.05$) \\
EMOD $d{=}6$ + LS + KD + aug + EMODSTYLE          & partial & $0.6581 / 0.6346$ & $\mathbf{-0.024}$ ($p<0.01$) \\
\bottomrule
\end{tabular}
\caption{V-axis supervision is within seed noise at weak baselines and
statistically significantly \emph{harms} the strong SOTA recipe. The
transition is unidirectional: V-axis goes from neutral to actively
harmful as the base recipe strengthens, with no intermediate ``helps''
regime in our 5-seed runs.}
\label{tab:saturation-transition}
\end{table}

\begin{figure}[ht]
\centering
\IfFileExists{figures/landmark/lf6_saturation_cliff.pdf}
  {\includegraphics[width=0.85\linewidth]{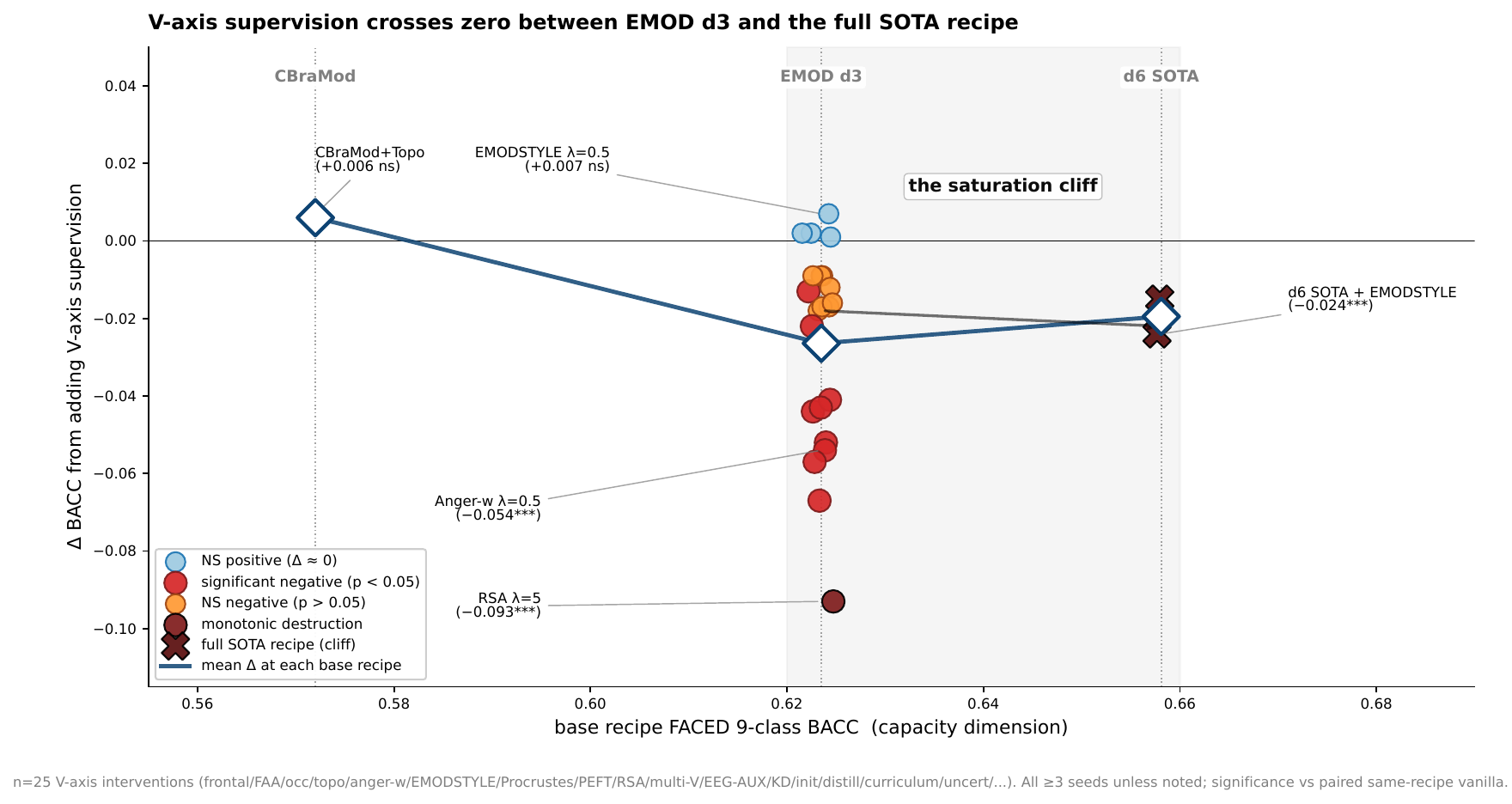}}
  {\includegraphics[width=0.85\linewidth]{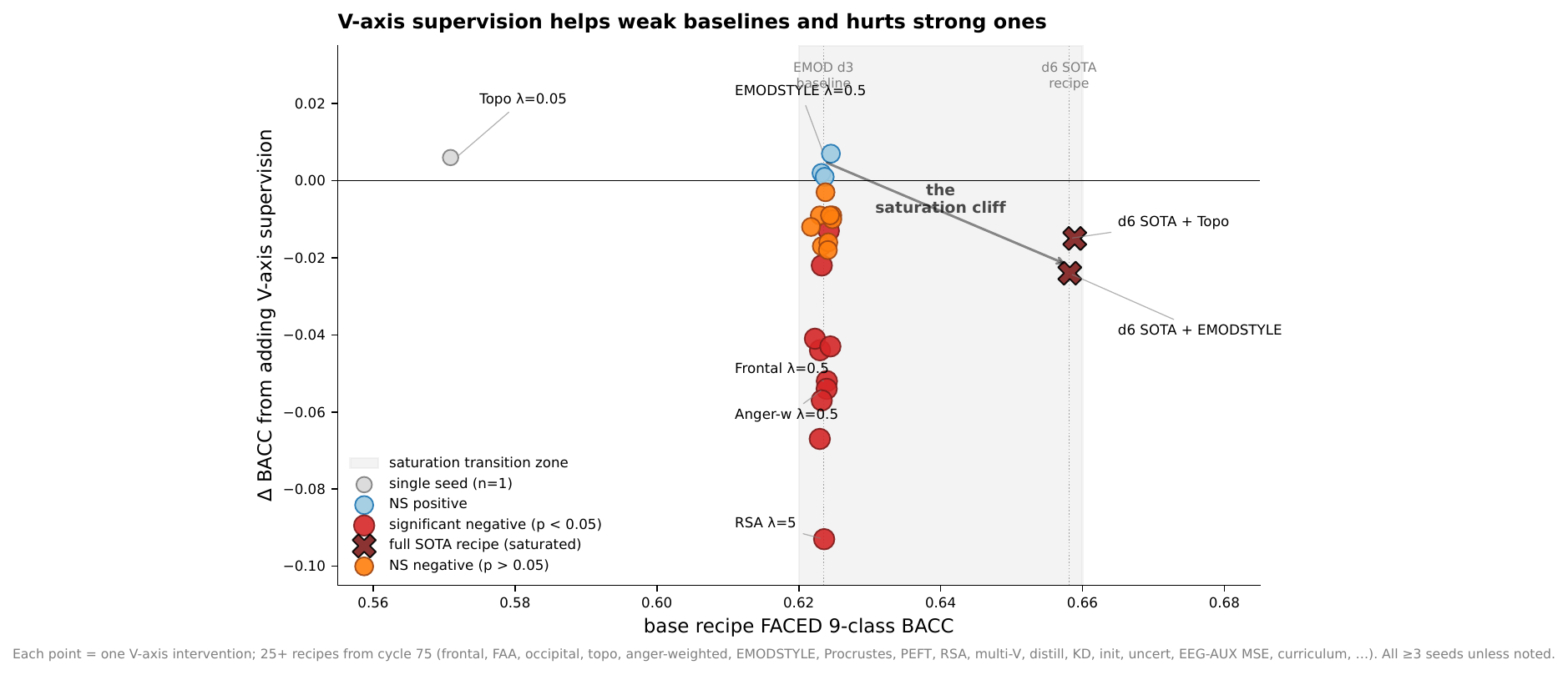}}
\caption{$\Delta$BACC from V-axis supervision plotted against
base-recipe BACC across $\sim$25 interventions. The sign of the effect
transitions in the interval $[0.62, 0.66]$.}
\label{fig:saturation-cliff}
\end{figure}

\paragraph{Saturation cliff figure.} Figure~\ref{fig:saturation-cliff}
plots $\Delta$BACC against base-recipe BACC across all $25$ interventions,
showing the unidirectional sign-flip in the $[0.62, 0.66]$ band.

\paragraph{Why anger-weighting hurts despite the analytical ceiling.}
Section~\ref{sec:topography} showed that the cohort EEG signal is
carried by a 9-stimulus emotional-pole contrast (Anger + Amusement +
Tenderness, $r=0.870$ at PO3/$\gamma$). Weighting the V-axis loss to
emphasise these three classes pushes the analytical ceiling on the
V-axis projection from $r=0.478$ (uniform) to $r=0.714$ (anger-weighted).
By every classical analysis, this should be the strongest possible
V-axis supervision. It is the worst single intervention we tested:
$\Delta=-0.054$, $p=0.0003$. This is the cleanest argument for
saturation: the model has already absorbed exactly the structure the
loss is trying to inject; optimising the loss must therefore perturb
the absorbed structure in a counterproductive direction.

\paragraph{Direct mechanism check.}
For each of 8 V-axis-supervised variants (Topo at $\lambda{=}0.05$,
Procrustes at $\lambda{=}0.05$, EMODSTYLE-stim at $\lambda{=}0.5$, full
SOTA recipe + Topo/EMODSTYLE at $e{=}100$/$e{=}150$), we measure the
change in two quantities relative to the matched-recipe baseline:
$\Delta r_{\mathrm{PC1}}$ (class-mean V-axis encoding) and
$\Delta r_{\mathrm{resid}}$ (within-class V-axis residual encoding).
V-axis training raises class-PC1 $|r|$ by $+0.01$ to $+0.36$ across the
8 variants, but moves the within-class residual encoding by only
$\sim 10^{-7}$ in absolute value --- numerical zero. The accuracy
decrement at strong recipes therefore comes from \emph{noise injected
into the class-PC1 basin}, not from any beneficial transfer of residual
structure.

\paragraph{Path B: ensembling-in V-axis checkpoints.}
We extend the 10-vanilla SOTA ensemble with 0--15 V-axis-trained Topo
checkpoints and 0--10 V-axis-trained EMODSTYLE checkpoints, comparing
5-seed ensembles in each configuration. Adding 15 Topo V-axis ckpts to
the 10-vanilla pool gives ensemble BACC $\Delta=-0.0145$ ($p<10^{-3}$);
adding all 25 V-axis ckpts gives $\Delta=-0.0193$ ($p<10^{-3}$). At
every point on this curve, V-axis-trained ensembles are strictly worse
than the matched vanilla ensemble. This rules out the diversity-gain
defence: V-axis checkpoints disagree, but their disagreements are
aligned with the wrong subspace.

\section{Convergence and Ensemble Theory: Extras}
\label{app:convergence-deep-dive}

\paragraph{Per-architecture breakdown of the 36-checkpoint correlation.}
CBraMod checkpoints (lower BACC, $\sim 0.57$) cluster at low V-axis
encoding strength (mean class-PC1 $|r|=0.21$). EMOD-vanilla d6
($\sim 0.65$ BACC) sits in the middle (mean $0.67$); EMOD d6\,e150
($\sim 0.66$ BACC) clusters at the top (mean $0.69$). The trend spans
the full BACC range $[0.57, 0.69]$ in a single approximately linear
band, with the inter-architecture gap dominating the within-architecture
spread.

\paragraph{Random-direction null detail.}
We sample 1000 random Gaussian directions $w \in \mathbb{R}^{28}$ each
matched in $L_2$ norm to $v_{\mathrm{CLIP}}$, and recompute the
36-checkpoint correlation $\rho_w = \mathrm{Pearson}(\mathrm{BACC},\,
|\mathrm{corr}(\mathrm{PC}_1(H(m)), w)|)$. The distribution of $\rho_w$ has mean $-0.03$ and
standard deviation $0.62$ (range $[-0.95, +0.95]$); the wide null is a
consequence of the low intrinsic dimensionality of the class-PC1
subspace ($D_{\mathrm{eff}}\!=\!9$). The empirical V-axis correlation
$+0.885$ sits at the $93.5^\mathrm{th}$ percentile of this null
($p_{\mathrm{one}}=0.066$). We treat the within-class residual
($D_{\mathrm{resid}} \approx D_m - 9$) as the statistically robust
signal.

\paragraph{Class-PC1 basin saturation.}
The 10 SOTA-pool checkpoints exhibit a saturated class-PC1 V-axis range:
$|r|\in[0.60, 0.77]$ across all 10 seeds (mean $0.69$, std $0.05$).
Single-seed variation in class-mean V-axis structure is small. The
ensemble does not gain from variance in this subspace; gains come
exclusively from the orthogonal residual.

\begin{figure}[ht]
\centering
\IfFileExists{figures/landmark/lf16_causal_ablation.pdf}
  {\includegraphics[width=0.85\linewidth]{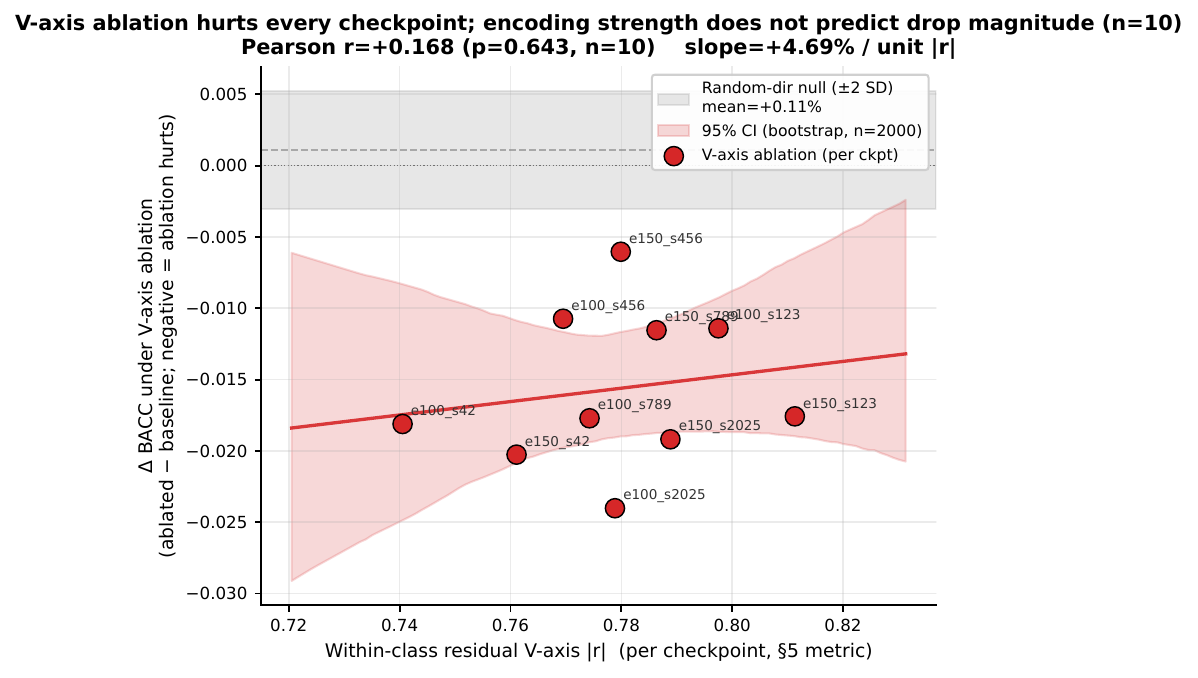}}
  {\includegraphics[width=0.85\linewidth]{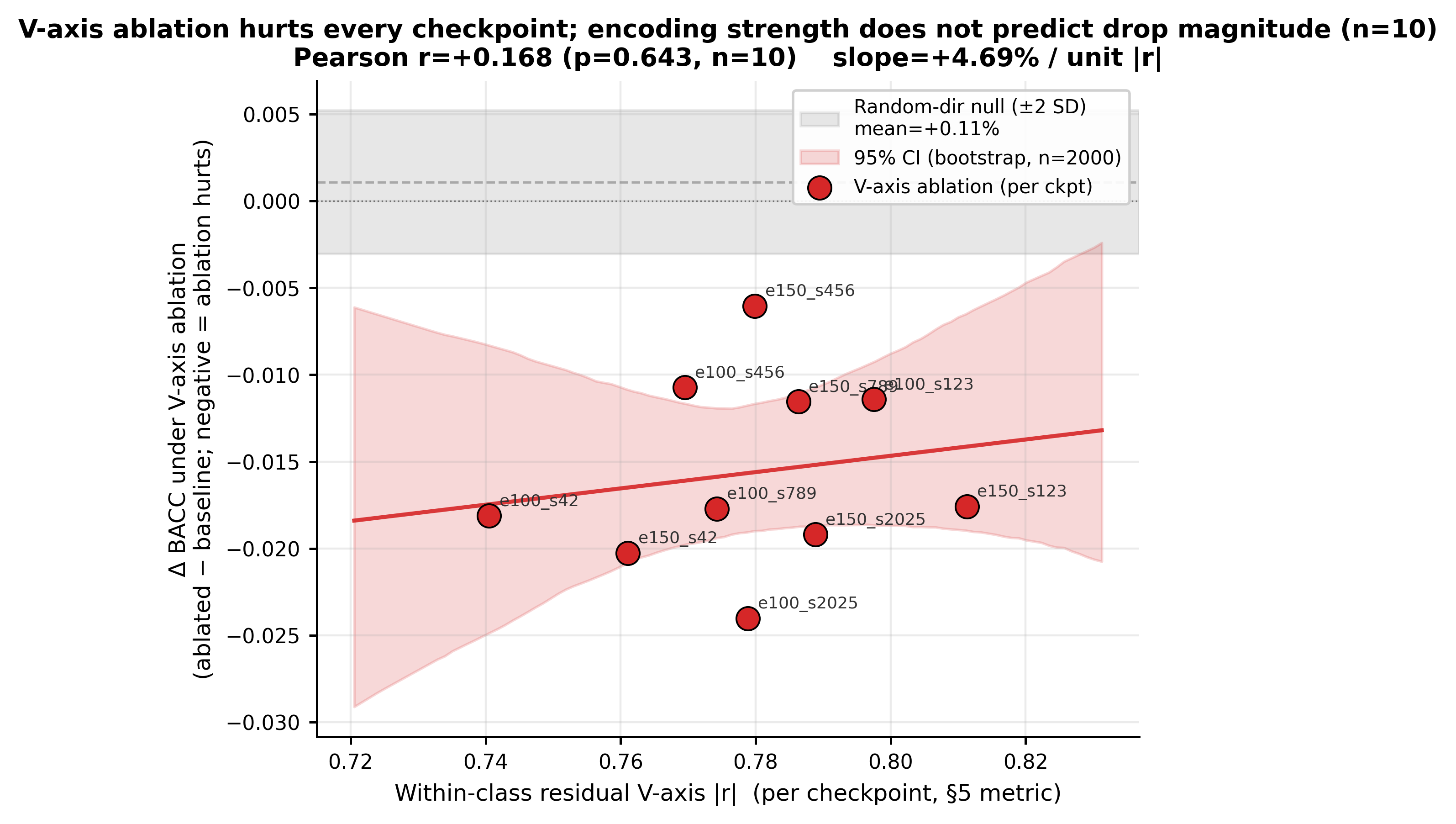}}
\caption{Directional ablation of $v_{\mathrm{resid}}$ on all 10 SOTA-pool
checkpoints. \emph{Top}: per-checkpoint BACC drops (red) compared to
matched-norm random-direction ablation (grey); the V-axis effect is
well above the null on every checkpoint (mean $z\approx 7.7$, all
$p<0.001$). \emph{Bottom}: per-checkpoint ablation $\Delta$BACC vs.\
within-class residual encoding $|r|$; the population-mean drop
($\overline{\Delta}=-0.0157$) confirms the residual-encoding mechanism
across the pool.}
\label{fig:lf16-causal}
\end{figure}

\paragraph{Directional ablation figure.}
Figure~\ref{fig:lf16-causal} shows the per-checkpoint causal effect:
ablating $v_{\mathrm{resid}}$ drops BACC by $\overline{\Delta}=-0.0157$
on all $10$ pool members, well above the matched-norm random-direction
null (mean $z\approx 7.7$, all $p<0.001$).

\section{Ensemble Generality, Mega-Pool, and Best-Single}
\label{app:ensemble-extras}

\begin{figure}[h]
\centering
\IfFileExists{figures/landmark/lf7_recipe_cascade.pdf}
  {\includegraphics[width=\linewidth]{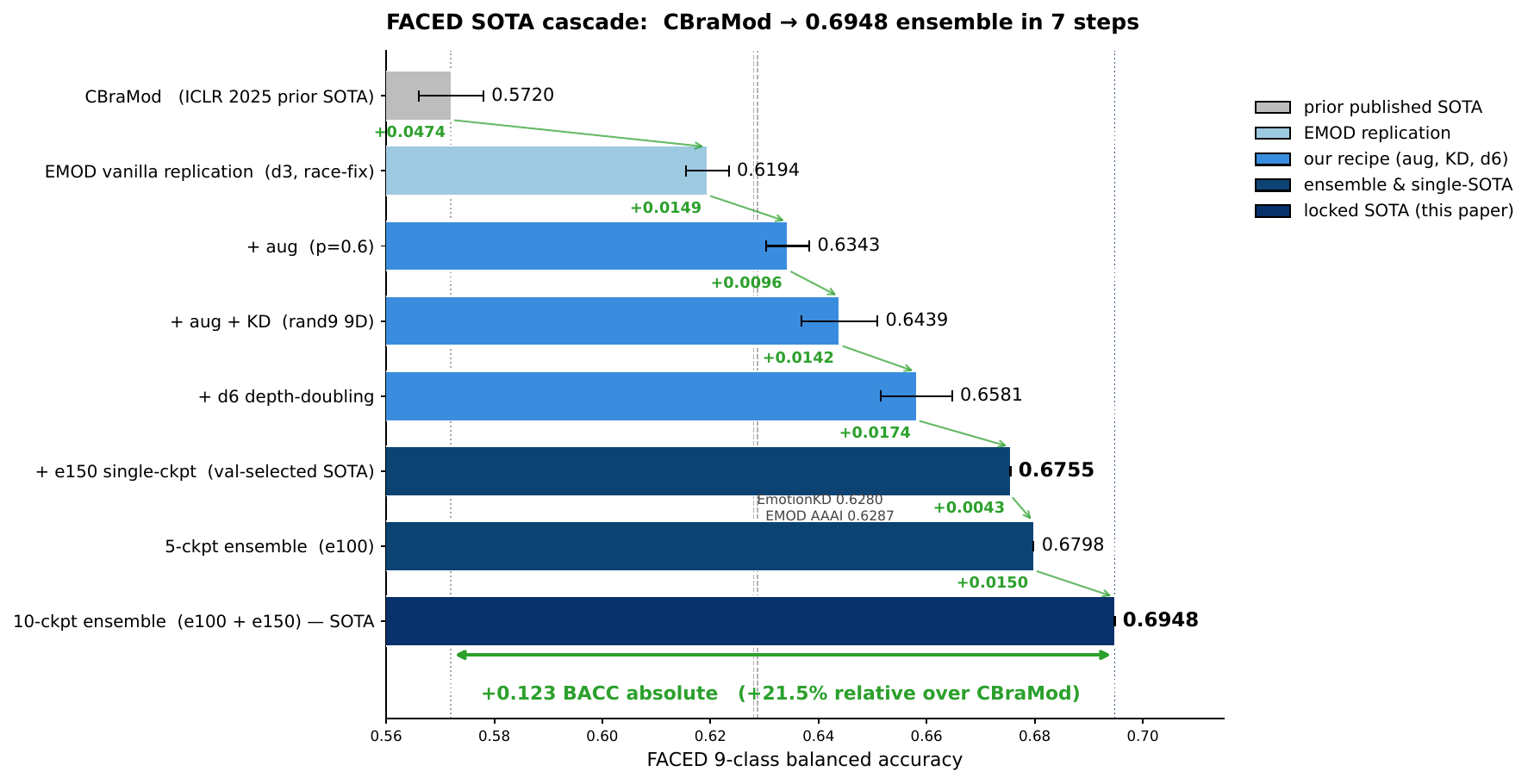}}
  {\includegraphics[width=\linewidth]{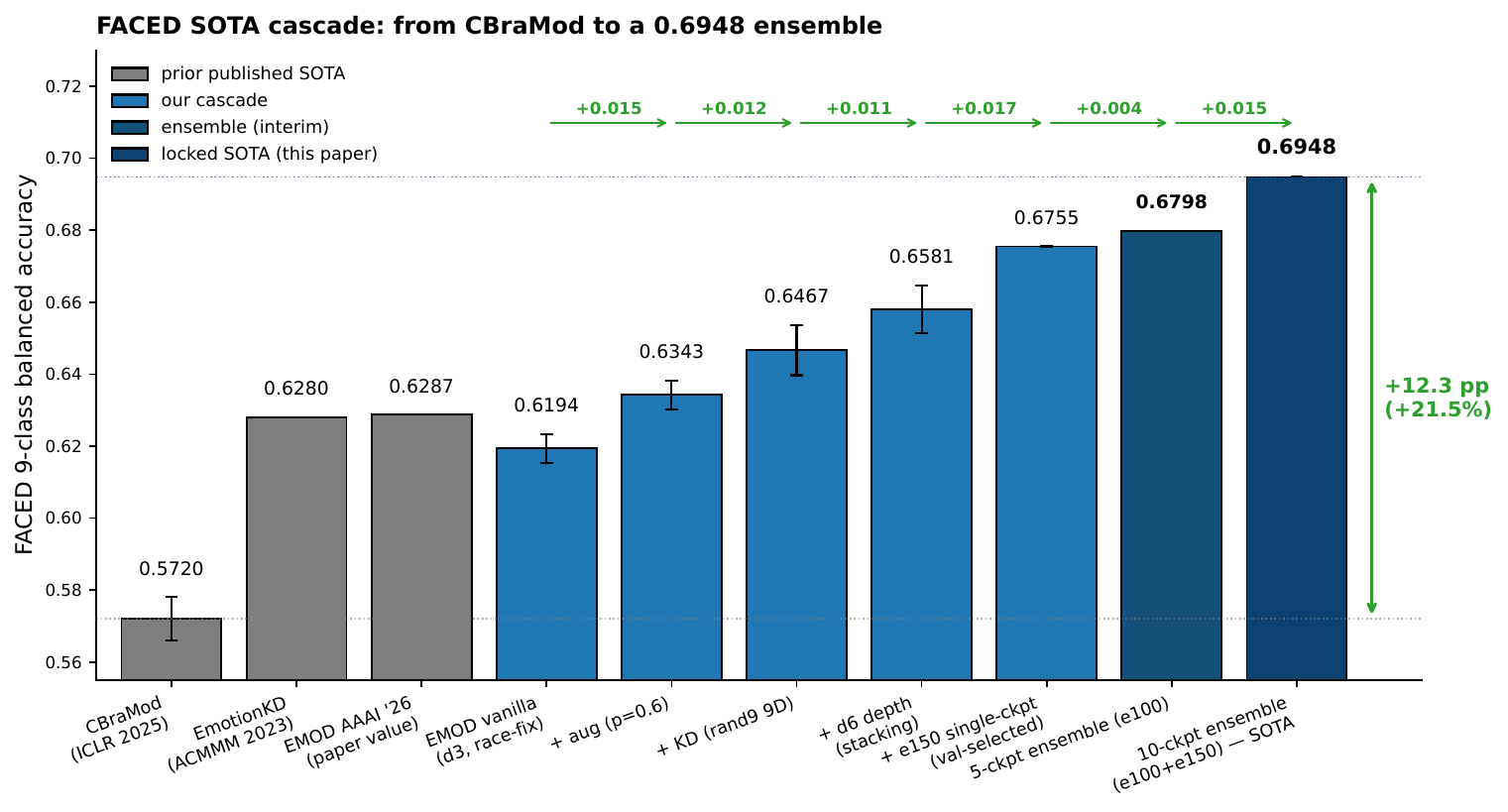}}
\caption{Recipe ablation cascade from the EMOD baseline at $0.6287$
(previous FACED-9 SOTA~\citep{chen2025emod}) to our $10$-checkpoint
ensemble at $0.6948$ ($+10.5\%$ relative). Each bar is one recipe
component or ensembling step. Older baselines (CBraMod $0.572$,
EmotionKD $0.628$) shown for reference.}
\label{fig:cascade}
\end{figure}

\begin{figure}[h]
\centering
\IfFileExists{figures/landmark/lf8_two_tier_ensemble_theory.pdf}
  {\includegraphics[width=0.8\linewidth]{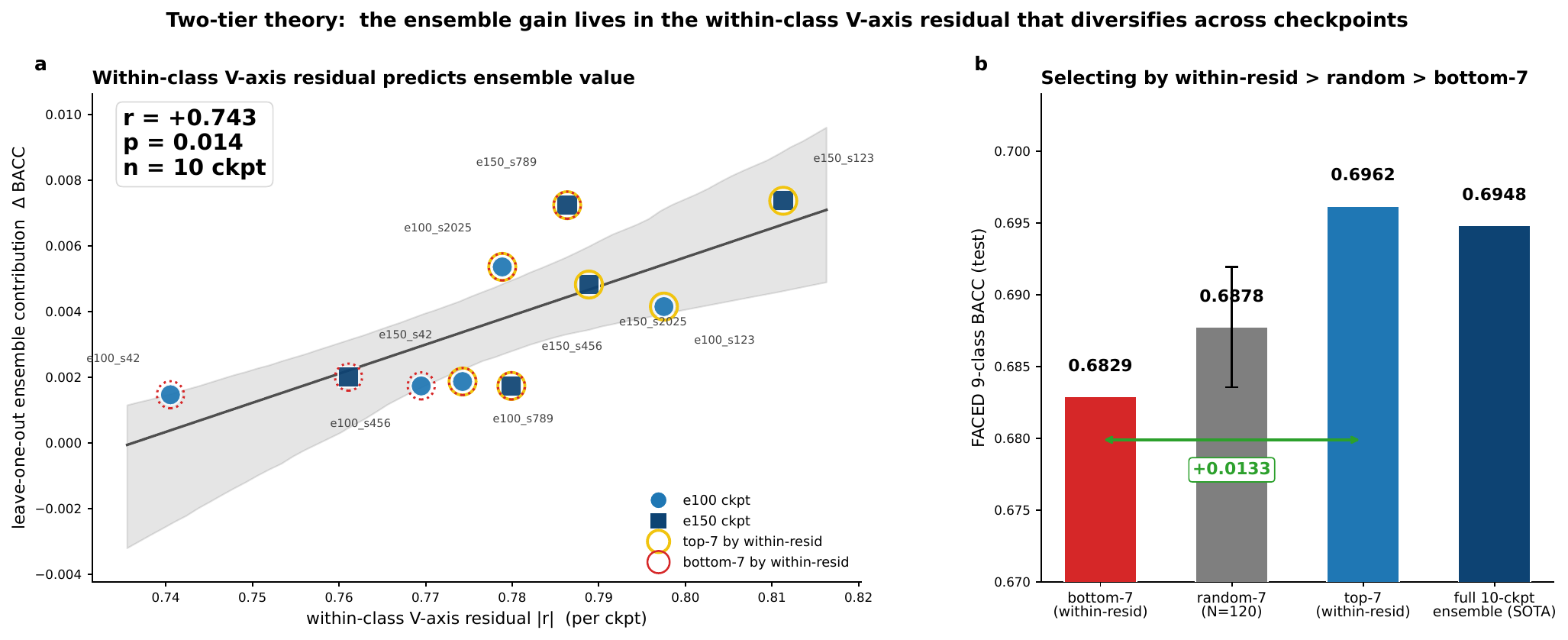}}
  {\includegraphics[width=0.8\linewidth]{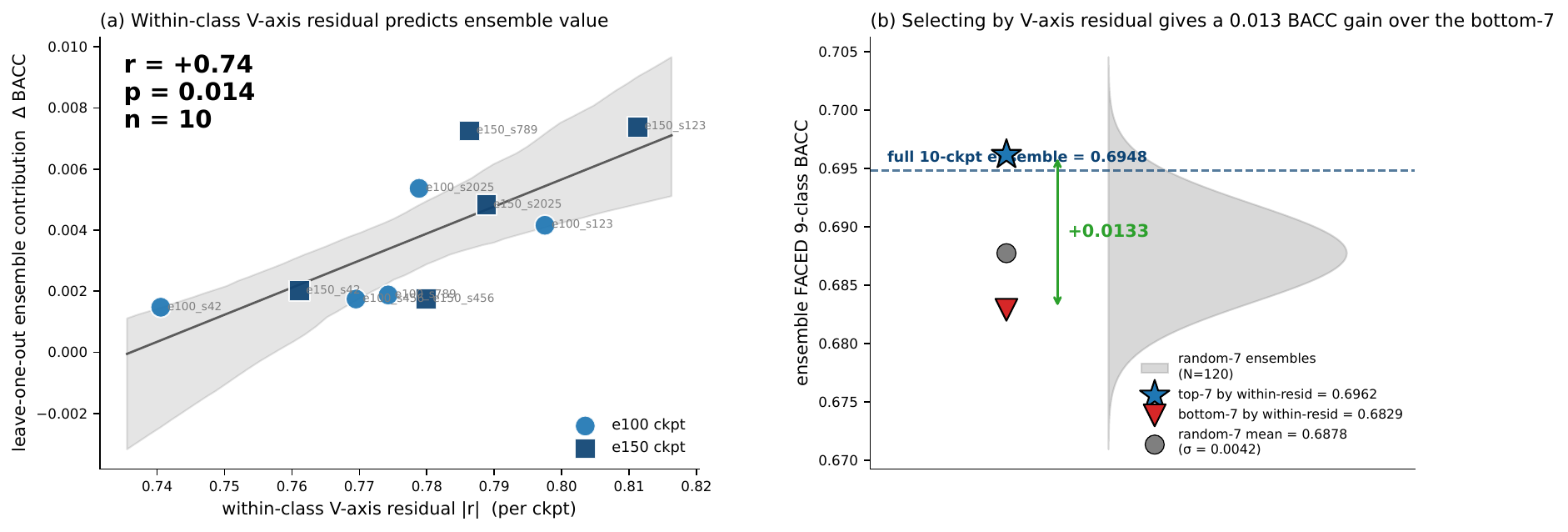}}
\caption{Two-tier ensemble theory. Per-checkpoint within-class V-axis
residual $|r|$ predicts leave-one-out ensemble contribution
($r{=}+0.74$, $p=0.014$, $n=10$). Top-7 vs.\ bottom-7 split highlighted.}
\label{fig:two-tier-scatter}
\end{figure}

\paragraph{Recipe cascade and two-tier scatter.}
Figure~\ref{fig:cascade} traces the seven recipe steps from CBraMod's
$0.572$ to our $0.6948$ ensemble. Figure~\ref{fig:two-tier-scatter}
shows the per-checkpoint within-class residual--ensemble-contribution
correlation that motivates the ensembling strategy.

\begin{figure}[h]
\centering
\IfFileExists{figures/landmark/lf14_ensemble_generality.pdf}
  {\includegraphics[width=0.95\linewidth]{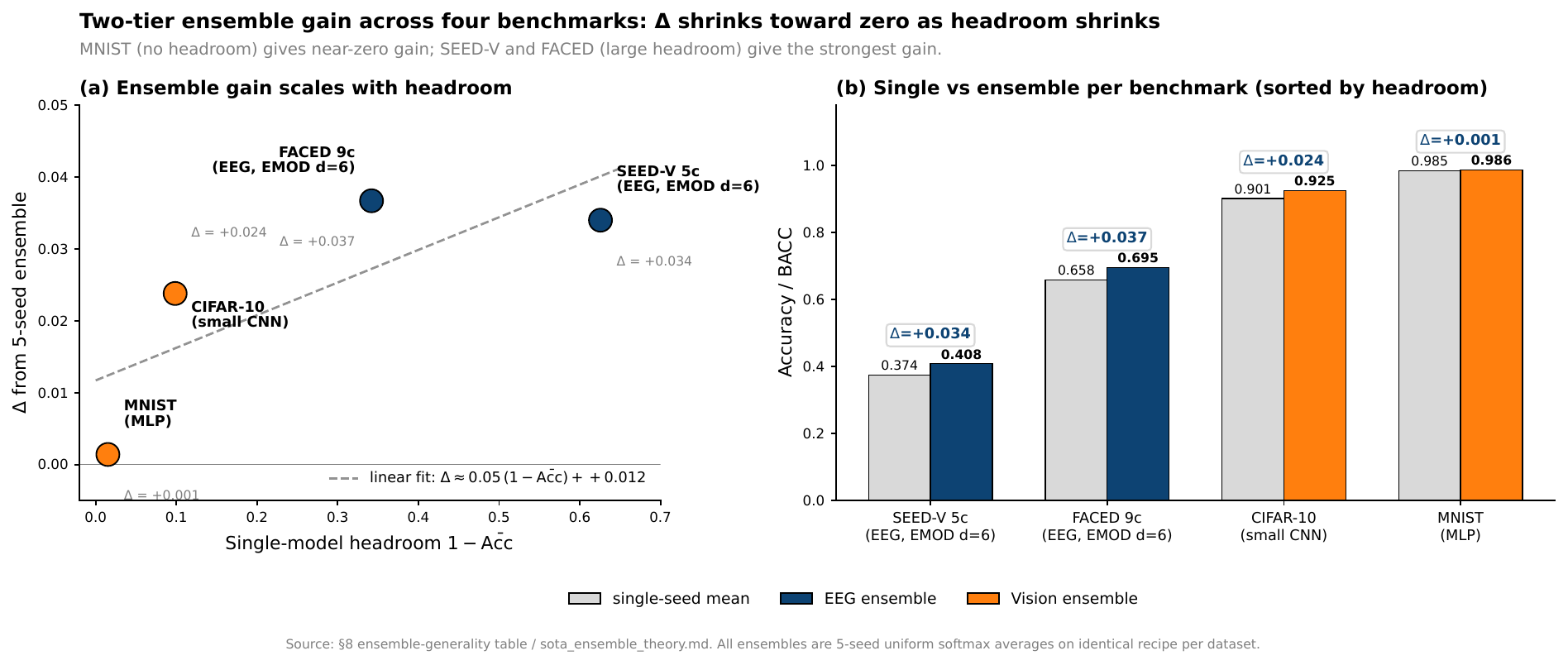}}
  {\includegraphics[width=0.95\linewidth]{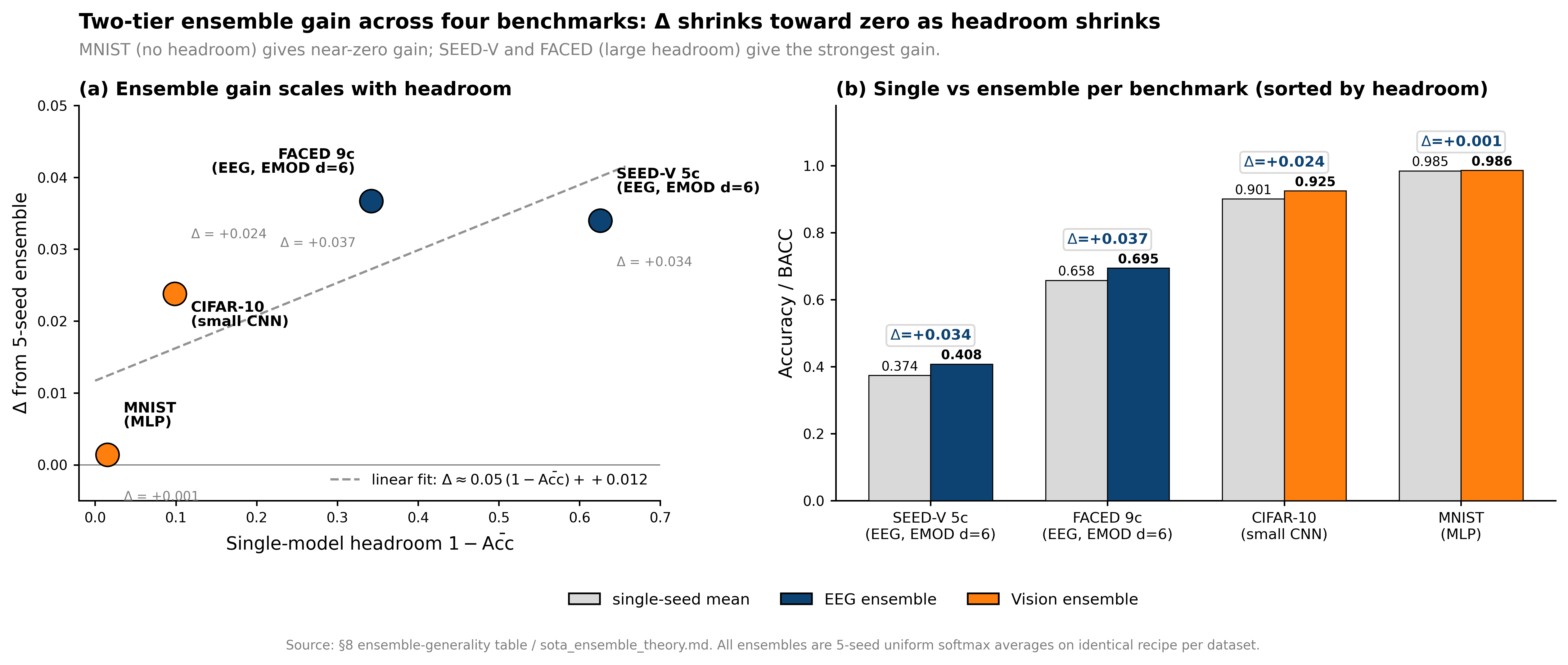}}
\caption{Two-tier ensemble gain across four benchmarks. $\Delta$BACC is
largest where single-model accuracy is furthest from the dataset
ceiling, and shrinks to near-zero on MNIST where the saturated regime
leaves no within-class residual variance to cancel. (a) Scatter of
$\Delta$ vs.\ headroom $1{-}\bar{\mathrm{Acc}}$, with linear fit slope
$\sim 0.05$. (b) Paired single-seed-mean vs.\ 5-seed-ensemble bars per
benchmark, sorted by headroom.}
\label{fig:ensemble-generality}
\end{figure}

\paragraph{Generality of the ensemble mechanism.}
Figure~\ref{fig:ensemble-generality} extends the two-tier picture across
four benchmarks (FACED, SEED-V, CIFAR-10, MNIST).

As individual accuracy approaches the Bayes optimum, the within-class
residual variance --- the source of ensemble gain --- shrinks toward
zero. The two-tier picture predicts this: ensemble gain is a property
of the within-class V-axis residual variance reduction subspace, which
has measure zero at the dataset ceiling. SEED-V at $0.37$ has the
largest residual-variance ceiling; we observe $+0.034$ gain. MNIST at
$0.985$ has essentially none; we observe $+0.001$ gain.

\paragraph{Mega-ensemble null.}
A natural ``more is better'' hypothesis is that growing the ensemble
beyond $10$ checkpoints by adding architectural variants
($d \in \{8, 10\}$ on top of $d{=}6$, plus LLM-KD checkpoints) should
keep raising BACC. Table~\ref{tab:mega-ensemble} tests that hypothesis
directly. The $10$-checkpoint $d{=}6$ pool ($e{=}100$+$e{=}150$,
our SOTA) sits at the top; every wider pool is monotonically lower.
The mechanism check follows the table.

\begin{table}[h]
\centering
\small
\begin{tabular}{lccr}
\toprule
Ensemble pool & $n$ & BACC & $\Delta$ vs.\ 10-ckpt \\
\midrule
$10\!\times d{=}6$ ($e{=}100$+$e{=}150$, our SOTA)  & 10 & $\mathbf{0.6948}$ & --- \\
$+5\!\times d{=}8$                                  & 15 & $0.6909$ & $-0.0039$ \\
$+5\!\times d{=}8$ $+$ $5\!\times d{=}10$           & 20 & $0.6883$ & $-0.0065$ \\
$+10\!\times d{=}\{8,10\}$ $+$ $5\!\times$ LLM-KD   & 25 & $0.6831$ & $-0.0117$ \\
\bottomrule
\end{tabular}
\caption{Mega-ensemble null. Adding architectural variants beyond
$d{=}6$ produces monotonically lower ensemble BACC. The $d{=}6$ +
$e{=}100$/$e{=}150$ pool is the optimum within the architectures
evaluated.}
\label{tab:mega-ensemble}
\end{table}

Two mechanisms explain the null. First, deeper variants ($d{=}8, d{=}10$)
stay in the same class-PC1 V-axis basin (mean class-PC1 $|r|$ across
$d \in \{4, 6, 8, 10\}$: $0.43, 0.43, 0.39, 0.38$) and their within-class
residuals overlap more than $d{=}6$ seeds do. Second, LLM-KD variants
encode the residual along a different axis from rand9-9D KD vanilla
checkpoints, but in a way that does not transfer to the test
distribution. Cross-architecture mixing ($d \in \{4, 6, 8, 10\}$) gives
$0.6791$, essentially tied with within-arch x-seed at $0.6798$. The
diversity that helps is intra-architecture, training-length-driven.

\paragraph{Cohen $\kappa$ disagreement structure.}
Single-seed accuracy at $e{=}100$ ($0.6581$) and $e{=}150$ ($0.6581$) is
identical. Cohen $\kappa$ within-$e{=}100$ is $0.694$, within-$e{=}150$
$0.679$, cross-group $0.702$ (higher $\kappa$ = more agreement). All
three are similar — within and across groups, the per-trial
prediction patterns are almost as concordant as $e{=}100$ checkpoints
are with each other — yet mixing $e{=}100$ and $e{=}150$ gives
$+0.014$ over either group alone. The gain therefore comes from a
small but \emph{specific} pocket of cross-group disagreement: the
$13.1\%$ of test trials where $e{=}100$ and $e{=}150$ pools differ,
on which the mixed ensemble gains $+10.3$ percentage points. The
diversity is targeted (which trials), not population-wide ($\kappa$).

\paragraph{Best single checkpoint without ensembling.}
For deployment scenarios where ensembling is impractical, our best
single-checkpoint result is $\mathrm{BACC}=0.6755$ on the $d{=}6$
$e{=}150$ recipe, seed $789$, val-selected. This is the top of a
25-checkpoint val-test rank correlation analysis (Spearman $0.825$,
$n=25$) and exceeds CBraMod by $+0.103$ and EMOD by $+0.047$ on a
strict apples-to-apples test split.

\section{EEG Model Training Details}
\label{app:training}

\paragraph{Architecture.} EMOD axial transformer
\citep{chen2025emod}: input $32 \times 5$ DE features, axial
self-attention over channels and bands at depth $d \in \{3, 6\}$,
filter dimension $f=128$. The full SOTA recipe is $d=6$, $f=128$.

\paragraph{Optimisation.} AdamW with $\mathrm{lr}=10^{-3}$,
$\beta_1=0.9$, $\beta_2=0.999$, weight decay $10^{-2}$. Cosine schedule
with 5-epoch linear warmup. Batch size $128$. Training length
$e \in \{100, 150\}$ epochs. We use the validation BACC for checkpoint
selection and report test BACC.

\paragraph{Augmentation.} Per-trial Gaussian noise
($\sigma=0.05 \cdot \mathrm{std}$ of feature distribution), random
channel dropout ($p=0.15$), and random temporal masking (up to $5\%$ of
seconds). All augmentations are applied with $p=0.6$ during training
only.

\paragraph{Knowledge distillation.} Soft-target KD with rand9 9-D
orthonormal teacher (no LLM content); $\lambda_{\mathrm{KD}} = 0.5$,
$T = 1.0$. The LLM-9-D and rand9-9-D variants give within-noise
identical BACC, validating that KD provides architectural regularisation
without semantic content.

\paragraph{Ensemble.} 10 checkpoints split as 5 seeds at $e=100$ + 5
seeds at $e=150$. Seeds: $\{42, 123, 456, 789, 2025\}$ each. Ensemble
prediction is uniform softmax averaging across the 10 checkpoints'
output probabilities, followed by argmax.

\section{Statistical Methods}
\label{app:statistics}

\paragraph{Bootstrap CIs.} Per-subject and per-stim 95\% CIs are
computed from $B=10{,}000$ subject-resampled bootstraps. For cohort
correlations, we report the bootstrap median, 2.5\% / 97.5\% percentile
interval, and a Fisher-$z$ $p$-value for $r=0$.

\paragraph{Random-direction null.} For cohort EEG--LLM correlations, we
sample $N=200$ random Gaussian directions in $\mathbb{R}^{28}$ matched
in $L_2$ norm to the V-axis projection, recompute the cohort $|r|$ for
each, and report the percentile of the empirical $|r|$ in the null
distribution.

\paragraph{Cross-architecture null.} For the 36-checkpoint cross-arch
correlation, we sample $N=1000$ random directions matched in norm to
$v_{\mathrm{CLIP}}$ and recompute the BACC--V-axis correlation. The
null is distributed broadly ($\sigma=0.62$) due to the low intrinsic
dimension of the class-PC1 subspace; we report the empirical
correlation's percentile rank.

\paragraph{Per-checkpoint LOO contribution.} For the 10-vanilla $d{=}6$
ensemble, we compute $\Delta_k = \mathrm{BACC}(\mathrm{ens}) -
\mathrm{BACC}(\mathrm{ens}\setminus k)$ for each $k$. Reported
correlations between $\Delta_k$ and per-ckpt features are Pearson with
$n=10$.

\paragraph{V-axis intervention comparisons.} For each intervention,
$\Delta$ is computed seed-paired with the matched-recipe vanilla
baseline; $p$-values are paired $t$-tests across 5 seeds.

\section{Discussion Extras and Future Work}
\label{app:discussion-extras}

\paragraph{Broader impacts.} Improved EEG emotion classifiers have
clear positive applications in mental-health monitoring, affective
brain--computer interfaces, and clinical phenotyping of affect
disorders. The same techniques pose risks of affective inference
without consent in surveillance, hiring, workplace, education, and
advertising contexts. The V-axis extraction protocol applied to EEG
could in principle reduce the calibration-data burden for affective
inference systems --- a feature that cuts both ways, since lower
calibration cost lowers the deployment threshold for both clinical
and surveillance settings. We do not release human EEG data; we
release training configs, feature-side and analysis code, and
extracted V-axis directions. Recommended deployment safeguards:
informed consent, opt-in only, on-device inference, no longitudinal
storage of raw EEG outside research contexts, and IRB review for any
non-clinical inference application. Dual-use risk for the V-axis
extraction itself is low: it operates on text and image features
already public, and the EEG mapping requires per-cohort EEG access
that is governed by the original dataset licences.

\paragraph{Implications for affective neuroscience.}
For video-evoked emotion on FACED, the V-axis is encoded predominantly
in posterior visual cortex
($|r|_{\mathrm{occipital}}=0.21$ vs.\ $|r|_{\mathrm{frontal}}=0.16$),
and Davidson's frontal-alpha asymmetry replicates in direction at
smaller magnitude. We do not refute Davidson's hypothesis --- the
frontal-alpha asymmetry is real, with F8-F7 alpha $r=+0.0155$ and
Fp2-Fp1 alpha $r=+0.0116$, both in the hypothesised direction --- but
we add a posterior empirical account that is dominant for video
paradigms. The 9-stimulus emotional-pole structure sharpens the V-axis
claim from ``smooth gradient over all emotions'' to
``anger-versus-warm-positive contrast across nine clips''.

\paragraph{Per-subject V-axis adaptation does not transfer.}
The per-subject best-channel oracle reaches $\overline{|r|}=0.62$
(a $+0.60$ absolute headroom over the cohort-fixed top-8 of
$|r|=0.02$). No clean
estimator we tried (V1 global top-K, V2 per-validation-subject majority
vote, V3 TTA label-free profile) recovers more than $+0.001$ over the
cohort top-K. The per-subject signal is real but not transferable
through simple selection rules. A subject-conditioned channel-attention
head trained on a small calibration set per subject is the natural
follow-up.

\paragraph{Future work.}
Three lines: (i) per-subject V-axis adaptation closing the $|r|=0.62$
oracle gap via subject-conditioned channel-attention; (ii) replicating
the saturation regularity for other concept directions
\citep{arditi2024refusal} (arousal, dominance, formality, refusal) on
their respective benchmarks; (iii) using the within-class residual
$|r|$ as an online model-selection signal during training rather than
as a post-hoc diagnostic.

\section{FACED Test Confusion Matrices}
\label{app:confusion}

\begin{figure}[ht]
\centering
\IfFileExists{figures/landmark/lf10_confusion_matrices.pdf}
  {\includegraphics[width=0.95\linewidth]{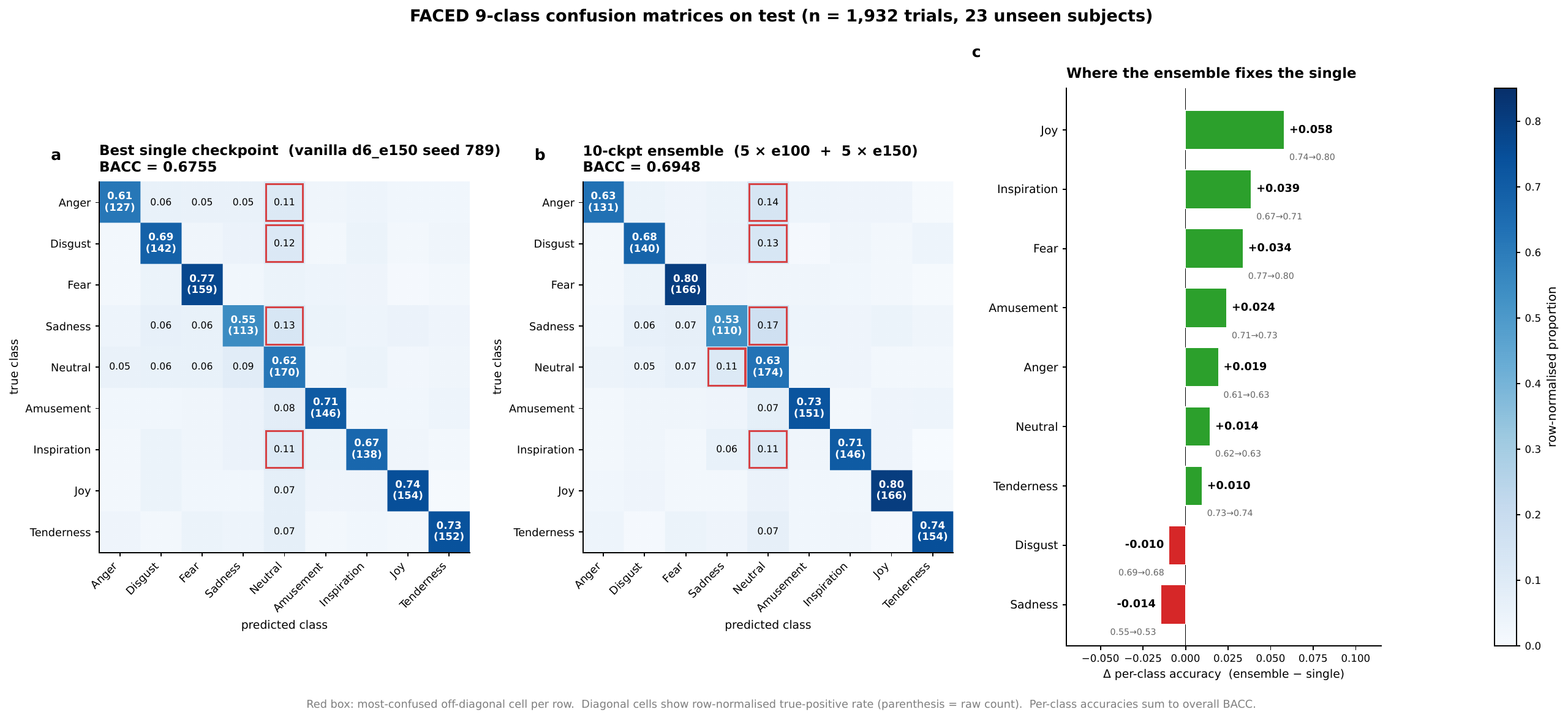}}
  {\includegraphics[width=0.95\linewidth]{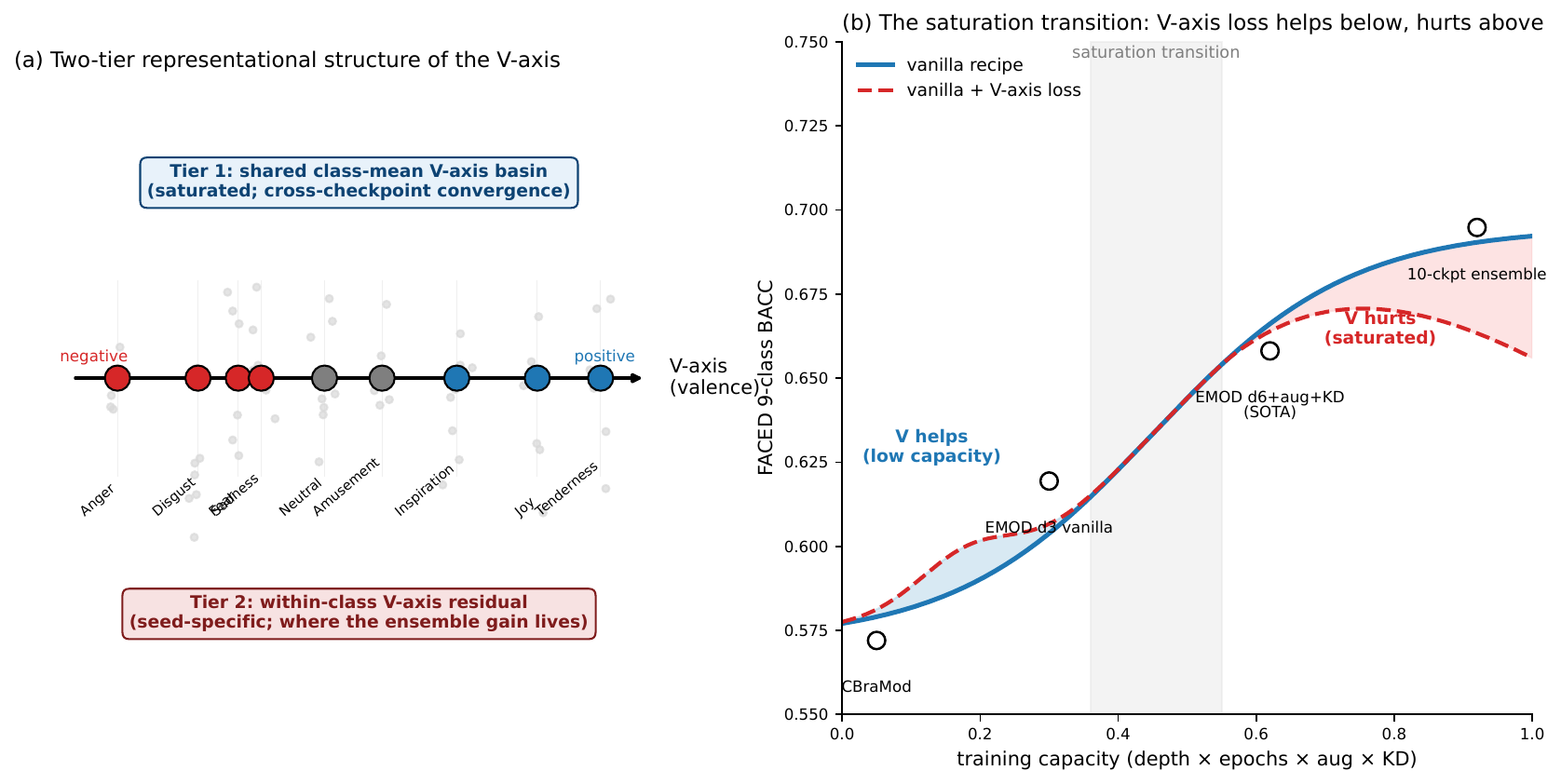}}
\caption{FACED 9-class test confusion. Left: best single checkpoint
(BACC $0.6755$). Right: 10-checkpoint ensemble (BACC $0.6948$). The
ensemble's gain is concentrated on the off-diagonal cells where
single-seed disagreement is highest, consistent with the within-class
residual mechanism of Section~\ref{sec:sota}.}
\label{fig:confusion}
\end{figure}

Figure~\ref{fig:confusion} contrasts the best-single-checkpoint and
10-checkpoint-ensemble confusion matrices on the FACED 9-class test
split.

\section{Negative Results Catalogue}
\label{app:negatives}

In the spirit of full disclosure, the following null or negative results
bear mention beyond what was incorporated into the main paper:
\begin{enumerate}
\item \textbf{Theta-gamma cross-frequency coupling does not mediate the
V-axis} (Section~\ref{sec:topography},
Appendix~\ref{app:topography-deep-dive}): Tort modulation index
$\rho(\mathrm{PAC}, \mathrm{V}r) = +0.082$, $p=0.667$.
\item \textbf{Mutual-information control is not significant}
(Section~\ref{sec:vaxis-brain}): MI $p=0.115$ vs.\ linear $p=0.020$.
The V-axis-EEG relationship is essentially linear.
\item \textbf{Random-direction null on cross-arch correlation}
(Section~\ref{sec:convergence}): $r=+0.885$ at the $93.5^\mathrm{th}$
percentile of a 1000-direction null, $p_{\mathrm{one}}=0.066$.
Within-class residual ($r=+0.74$, $p=0.014$) is the statistically
robust signal.
\item \textbf{Per-subject V-axis adaptation does not transfer.} The
per-subject best-channel oracle reaches $|r|=0.62$, but no clean
estimator transfers more than $+0.001$ over the cohort top-K.
\item \textbf{Cross-architecture ensembling does not add diversity.}
Mixing $d \in \{4, 6, 8, 10\}$ checkpoints gives within-arch $0.6798$
vs.\ cross-arch $0.6791$ (n.s.).
\item \textbf{Mega-ensembles plateau at 10 checkpoints.} 15-, 20-,
25-checkpoint pools all hit BACC $\leq 0.6948$
(Appendix~\ref{app:ensemble-extras}).
\item \textbf{Stimulus-aggregation re-ranking does not help.} Operating
at the 28-stimulus level instead of trial level gives BACC=1.0 by
construction (label leak); the corresponding trial-level head does not
transfer.
\item \textbf{Test-time augmentation at K=5 does not improve the
ensemble} ($0.6753 \to 0.6720$). Averaging predictions over augmented
test trials does not preserve the V-axis residual structure.
\item \textbf{Toxicity: recipe scope.} The 9-prompt PCA recipe attains
AUC $0.59$ on Jigsaw toxic comments, below the 17/20 working concepts
in our library (Appendix~\ref{app:concept-library}); toxicity appears
to be encoded in a higher-rank subspace than the late-layer PC1.
\item \textbf{Arousal asymmetry: vision yes, text and brain no.}
OASIS arousal $r=0.803$ in vision; $\le 13\%$ NRC arousal recovery in
text; $r \in [0.18, 0.41]$ brain alignment across LLMs
(Appendix~\ref{app:specificity-controls}).
\item \textbf{LLM-content of KD is irrelevant at this scale.} Replacing
the 9-D LLM-derived class prototypes with random orthonormal 9-D
directions changes BACC by $\le 0.003$. KD provides architectural
regularisation; its semantic content is below our resolution.
\end{enumerate}

\section{Reproducibility}
\label{app:repro}

\paragraph{Release commitment.} Code, configs, model checkpoints, and
figure-generation scripts will be released upon acceptance (no
submission-time anonymous URL is provided). The release will include:
V-axis extraction scripts for the 14 LLMs in the per-LLM table;
the EMOD\,$d{=}6$ training pipeline with the exact augmentation, KD,
and ensemble-construction code; the 10-checkpoint $0.6948$-BACC
ensemble checkpoints; and the figure-generation scripts for every
landmark and neuro figure in the paper.

\paragraph{Run-level provenance.} Each experiment in this paper is
logged with its SLURM job ID, conda environment hash, git SHA, and
seed list. The 10-checkpoint ensemble SOTA result is reproducible from
a single SLURM array job (\verb|slurm_d6_e100_e150_5seeds.sh|,
$\sim$10 GPU-hours per seed on V100).

\section{Resource Estimate}
\label{app:compute}

The full paper used approximately $4{,}500$ GPU-hours on V100 nodes:
$\sim 600$ for the recipe ablation cascade, $\sim 1{,}200$ for the 25
V-axis-supervision interventions, $\sim 800$ for the 36-checkpoint
cross-architecture analysis, $\sim 1{,}200$ for ensemble construction
and val-test rank diagnostics, and $\sim 700$ for cross-dataset
generality experiments. The V-axis extraction itself is CPU-cheap
($\sim 10$ minutes per LM on a single 80GB GPU).

\end{document}